\documentclass[preprint]{acmart}
\renewcommand\footnotetextcopyrightpermission[1]{} % removes footnote with conference information in first column
\usepackage{booktabs} % For formal tables

\usepackage[ruled]{algorithm2e} % For algorithms

\usepackage{booktabs} % For formal tables
\usepackage{url}
\usepackage{color}
\usepackage{multirow}
\usepackage{enumitem}
\usepackage{array}
\usepackage{siunitx}
\usepackage[labelfont={bf}]{caption}
\usepackage{balance}
\usepackage{array} 
\usepackage{graphicx}
\usepackage{todonotes}
\usepackage{subfloat}
\usepackage{subfig}
\usepackage[export]{adjustbox}
\usepackage{natbib}
\usepackage[nolist,nohyperlinks]{acronym}
\usepackage{footnote}
\usepackage{color,soul}

\makesavenoteenv{tabular}

\SetAlFnt{\small}
\SetAlCapFnt{\small}
\SetAlCapNameFnt{\small}
\SetAlCapHSkip{0pt}
\IncMargin{-\parindent}

\acrodef{GI}{Gastrointestinal}
\acrodef{CRC}{Colorectal Cancer}
\acrodef{EU}{European Union}
\acrodef{CAD}{Computer-Aided Diagnosis}
\acrodef{CNN}{Convolutional Neural Network}
\acrodef{DNN}{Deep Neural Network}
\acrodef{GF}{Global Feature}
\acrodef{SVM}{Support Vector Machine}
\acrodef{DL}{deep learning}
\acrodef{HC}{handcrafted}
\acrodef{LMT}{Logistic Model Tree}
\acrodef{SL}{SimpleLogistic}
\acrodef{MCC}{Matthews Correlation Coefficient}
\acrodef{MLP}{Multi-Layer Perceptron}
\acrodef{ML}{machine learning}
\acrodef{TN}{True Negative}
\acrodef{TP}{True Positive}
\acrodef{SGD}{Stochastic Gradient Descent}
\acrodef{SSD}{Single Shot MultiBox Detector}
\acrodef{MCC}{Matthews Correlation Coefficient}
\acrodef{DNN}{Deep Neural Network}
\acrodef{ROC}{Receiver Operating Characteristic}
\acrodef{PRC}{Precision-Recall Curve}
\acrodef{REC}{Recall}
\acrodef{PREC}{Precision}
\acrodef{SPEC}{Specificity}
\acrodef{ACC}{Accuracy}
\acrodef{REC}{Recall}
\acrodef{F1}{F-score}
\acrodef{CM}{Confusion Matrix}
\acrodefplural{CM}{Confusion Matrices}
\acrodef{CMs}{Confusion Matrices}
\newcommand{\simulamet}{SimulaMet}  
\newcommand{\oslomet}{Oslo Metropolitan University}

\newcolumntype{P}[1]{>{\centering\arraybackslash}p{#1}}

\begin{document}
%\floatsetup[figure]{style=plain,subcapbesideposition=top}
%\captionsetup[subfigure]{position=top, labelfont=bf,textfont=normalfont,singlelinecheck=off,justification=raggedright}

% Title portion. Note the short title for running heads

\title[GI Tract Abnormaly Classification with Cross-Dataset Evaluation]{[Preprint] \\ An Extensive Study on Cross-Dataset Bias and Evaluation Metrics Interpretation for Machine Learning applied to Gastrointestinal Tract Abnormality Classification
}

\author{Vajira Thambawita}
\affiliation{%
  \institution{\simulamet}
%  \streetaddress{P.O. Box 134}
  \city{Oslo}
%  \state{Lysaker}
%  \postcode{1325}
  \country{Norway}}
  \affiliation{%
  \institution{\oslomet}
  \city{Oslo}
  \country{Norway}
}
\email{vajira@simula.no}

\author{Debesh Jha}
\affiliation{%
  \institution{SimulaMet}
  \city{Oslo}
  \country{Norway}}
\affiliation{%
  \institution{UiT The Arctic University of Norway}
  \city{Troms{\o}}
%  \country{Norway}
}  
\email{debesh@simula.no}

\author{Hugo Lewi Hammer}
\affiliation{%
  \institution{\oslomet}
%  \city{Oslo}
  \country{Norway}}
  \affiliation{%
  \institution{\simulamet}
  \city{Oslo}
  \country{Norway}
}
\email{hugoh@oslomet.no}

\author{H{\aa}vard D. Johansen}
\affiliation{%
  \institution{UiT The Arctic University of Norway}
  \city{Troms{\o}}
%  \country{Norway}
}
\email{havard.johansen@uit.no}

\author{Dag Johansen}
\affiliation{%
  \institution{UiT The Arctic University of Norway}
  \city{Troms{\o}}
%  \country{Norway}
}
%\affiliation{%
 % \institution{Institution 7}
 % \country{}}
\email{dag.johansen@uit.no}

%%%%%%%%%%%%%%%%%%%%%
\author{P{\aa}l Halvorsen}
\affiliation{%
  \institution{\simulamet}
  \city{Oslo}
  \country{Norway}
}
\affiliation{%
  \institution{\oslomet}
  \city{Oslo}
  \country{Norway}
}
%\affiliation{%
%  \institution{University of Oslo, Norway }
%  \department{}
%  \city{Oslo}
%  \state{}
%  \postcode{}
%  \country{Norway}
%}  
\email{paalh@simula.no}
%%%%%%%%%%%%%%%%%%%%%%%%%%%
\author{Michael A. Riegler}
\affiliation{%
  \institution{\simulamet}
%  \streetaddress{P.O. Box 134}
  \city{Oslo}
%  \state{Lysaker}
%  \postcode{1325}
  \country{Norway}
  }
%\affiliation{%
%  \institution{\krist}
%  \department{}
  \city{Oslo}
%  \state{}
%  \postcode{}
  %\country{Norway}
%}  
\email{michael@simula.no}

\begin{abstract}

Precise and efficient automated identification of \ac{GI} tract diseases can help doctors treat more patients and improve the rate of disease detection and identification. Currently, automatic analysis of diseases in the GI tract is a hot topic in both computer science and medical-related journals. Nevertheless, the evaluation of such an automatic analysis is often incomplete or simply wrong. Algorithms are often only tested on small and biased datasets, and cross-dataset evaluations are rarely performed. A clear understanding of evaluation metrics and \acl{ML} models with cross datasets is crucial to bring research in the field to a new quality level.
Towards this goal, we present comprehensive evaluations of five distinct \acl{ML} models using \aclp{GF} and \aclp{DNN} that can classify 16 different key types of GI tract conditions, including pathological findings, anatomical landmarks, polyp removal conditions, and normal findings from images captured by common \ac{GI} tract examination instruments. In our evaluation, we introduce performance hexagons using six performance metrics such as recall, precision, specificity, accuracy, F1-score, and \acl{MCC} to demonstrate how to determine the real capabilities of models rather than evaluating them shallowly. 
Furthermore, we perform cross-dataset evaluations using different datasets for training and testing. With these cross-dataset evaluations, we demonstrate the challenge of actually building a generalizable model that could be used across different hospitals. Our experiments clearly show that more sophisticated performance metrics and evaluation methods need to be applied to get reliable models rather than depending on evaluations of the splits of the same dataset, i.e., the performance metrics should always be interpreted together rather than relying on a single metric.

%%%%%%%%%%%%%%%%%%%%%%%%%%%%%%%%%%%%%%%%%%%%%%%%

\end{abstract}




\maketitle
 
\renewcommand{\shortauthors}{Thambawita et al. [preprint]}

%=== Main boady ===========
%\input{mainbody}

% This file is the main file containing all other sections

%=============================
\section{Introduction}
\label{sec:intro}

Cancer is one of the leading causes of death worldwide and a significant barrier to life expectancy~\cite{bray2018global}. In particular, the \acf{GI} tract can be affected by a variety of diseases and abnormalities~\cite{pogorelov2017efficient}. Using data from the Global Cancer Observatory,\footnote{\url{https://gco.iarc.fr}} \Citet{bray2018global} estimated that, for 2018, there would be around 5~million new luminal \ac{GI} cancer incidences and about 3.6~million deaths due to \ac{GI} cancer.\footnote{We have considered the statistic of esophagus, stomach, colon, rectum, anus, gallbladder, and pancreas.} The most frequently diagnosed \ac{GI} cancers in 2018 for new cases are \ac{CRC} (6.1\%), stomach cancer (5.7\%), liver cancer (4.7\%), rectum cancer (3.9\%), and esophageal cancer (3.2\%) out of 36 types of cancers~\cite{bray2018global}.

Gastroscopy and colonoscopy are the most successful medical procedures for \ac{GI} endoscopy examinations. Among both, colonoscopy has been proven to be an effective preventative method by improving declination in the occurrence of \ac{CRC} by 30\%~\cite{lieberman2005quality}. During a colonoscopic procedure, endoscopists insert a colonoscope carefully through the anus to examine the rectum and colon. A tiny wide-angle video camera mounted at the end of the colonoscope captures a live video signal of the internal mucosa of the patient's colon. The endoscopist uses the video signal for real-time diagnosis of the patient, where one of the primary goals is to identify and remove abnormalities such as polyps~\cite{wang2015polyp}. 

The current \acs{EU} guidelines~\cite{von2012european} recommend \ac{GI} tract screening for all people above 50~years. Such regular screenings can be of great significance for early detection and prevention of cancer inside the GI tract, but they are challenging due to many factors. Moreover, a colonoscopy examination is entirely an operator-dependent screening procedure~\cite{shin2017comparison}. The detection rate of \ac{GI} tract lesions mostly relies on the clinical experience of the gastroenterologist. The shortage of experienced gastroenterologists, and the clinicians' tiredness and lack of concentration during the colonoscopic examination, can lead to missing polyps that otherwise would be detected~\cite{tajbakhsh2016automated}. The estimated miss rate for the subject undergoing a colonoscopy examination is 25\%~\cite{leufkens2012factors}.

Although considerable work has been done to develop and improve systems for automatic polyp detection, the performance of existing solutions are still behind that of an expert endoscopist~\cite{de2018methodology,mori2018detecting,wang2018development,wang2014part,bernal2012towards}. Most of the published papers in the field use non-public datasets or develop models from too small training, validation, and test sets~\cite{wang2014part,bernal2012towards,wang2018development}. The performance metrics used to measure the performance of methods are also not sufficient (for example, see the first part of Table~\ref{tab:Overview_of_the_related_work}). Thus, it is difficult for researchers to compare and reproduce some of the present related works. Moreover, the state of the art research in this field does not present the generalizability of their solutions using cross-dataset evaluations.  As a result, it makes a distrust for applying these \ac{ML} solutions in practice.

An automatic and efficient \emph{\ac{CAD} system} in a clinic could assist medical experts during the endoscopic and colonoscopy procedure to improve the detection rate by finding unrecognized lesions and act as a second observer by providing better insights to the gastroenterologist concerning the presence and types of lesions. With this inspiration, we conducted five experiments to classify $16$ classes of \ac{GI} tract conditions for the Medico Multimedia Task at MediaEval 2018~\cite{pogorelov2018medico}. One example for each of the 16 classes is depicted in Figure~\ref{fig:images_16}.

In this work, we focus on identifying the limitations of generalizing \ac{ML} models across different datasets and how to interpret the evaluation metrics in that context.  For this, we are using \acf{GF}-based and \acf{DL}-based methods that performed well at the 2018 Medico Task~\cite{pogorelov2018medico} where one specific dataset was used. In addition, we here explore the different performance metrics of both methods (\ac{GF} and \ac{DL}) to identifying the limitations of each.  We show that combined complex \ac{DNN} models outperform other methods. Finally, we explore how multiclass models perform on polyp and non-polyps detection with and without retraining the model for the two specific classes. The effects of retraining for classifying the sub-categories of the same dataset and using them in other datasets are analyzed in detail to identify the cross-dataset generalization capabilities of our models.  We emphasize that a large number of performance measures do not show the real performance of ML models. We also highlight the necessity of having cross-dataset evaluations to determine the real capabilities of \ac{ML} models before using them in clinical settings.

%%%%%%%%%%%%%%%%%%%%%%%%%%%%%%%%%%%%%%%%%%%%%%%%%%%%%%%%%%%%%%%%%%%%%%%%%%%%%%%%%%%%%%%%%%%
% Images of 16 classes
%%%%%%%%%%%%%%%%%%%%%%%%%%%%%%%%%%%%%%%%%%%%%%%%%

\newcommand{\figsz}{1.5cm}
\newcommand{\figtxt}{\footnotesize}

\begin{figure}[t]
    \centering
    \subfloat[][\centering\figtxt Blurry nothing]{
        \includegraphics[width=\figsz, height=\figsz]{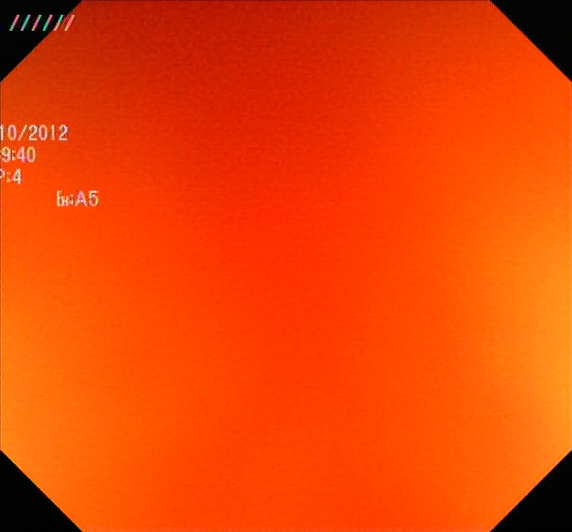}
    }
    \subfloat[][\centering\figtxt Colon clear]{
        \includegraphics[width=\figsz, height=\figsz]{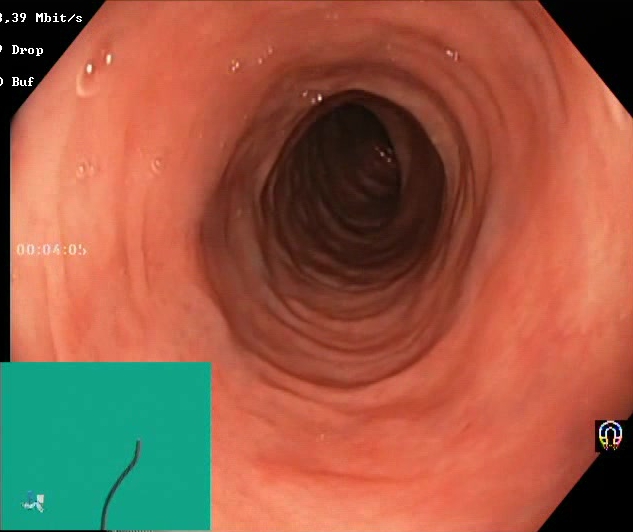}
        \label{fig:colon_clear}
    }
    \subfloat[][\centering\figtxt Dyed lifted polyps]{
        \includegraphics[width=\figsz, height=\figsz]{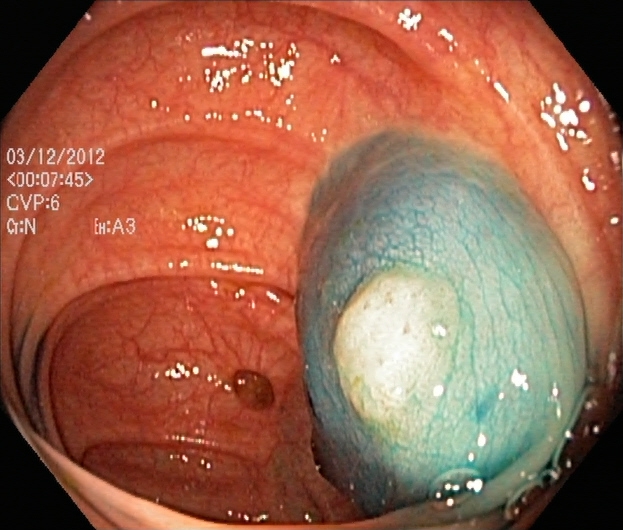}
        \label{fig:dyed_lifted}
    }
     \subfloat[][\centering\figtxt Dyed resection margins]{
        \includegraphics[width=\figsz, height=\figsz]{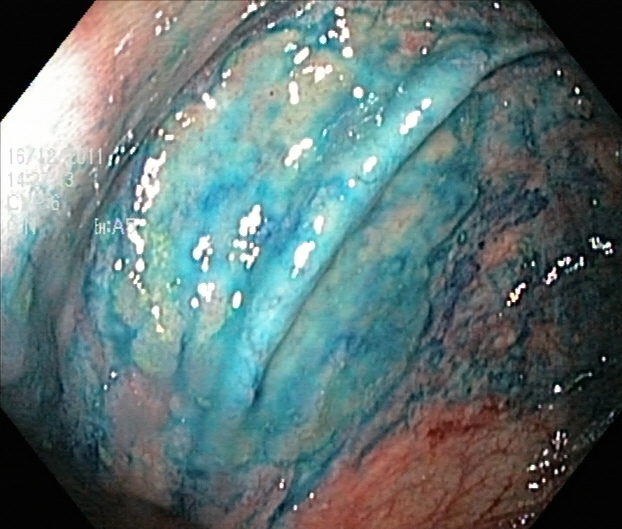}
        \label{fig:dyed_resection}
    }
    \hspace{0mm}
     \subfloat[][\centering\figtxt Esophagitis]{
        \includegraphics[width=\figsz, height=\figsz]{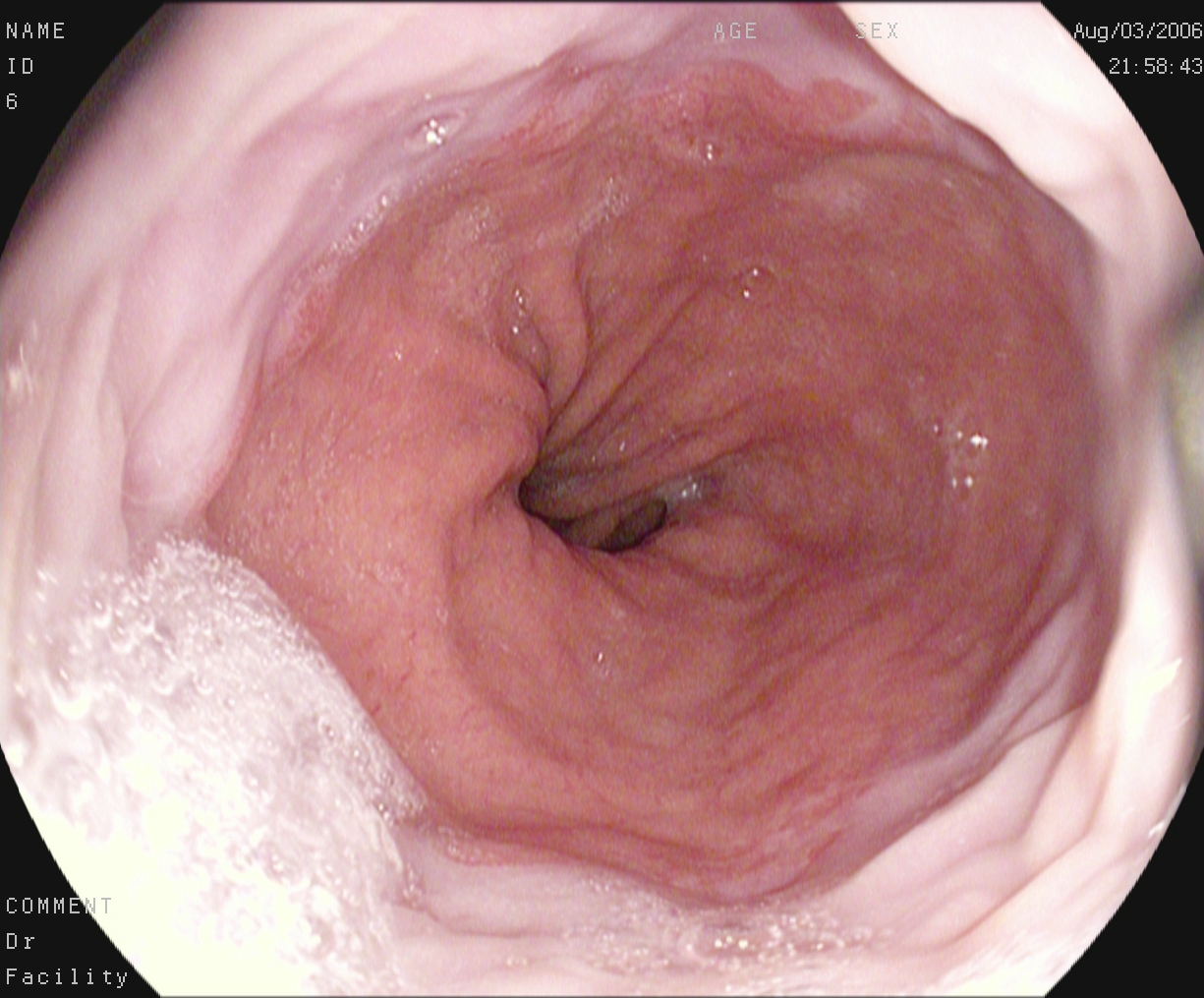}
        \label{fig:esophagitis}
    }
     \subfloat[][\centering\figtxt Instruments]{
        \includegraphics[width=\figsz, height=\figsz]{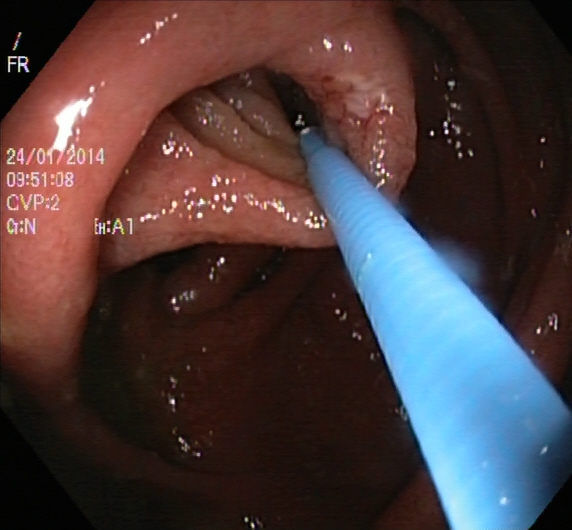}
   }
     \subfloat[][\centering Normal cecum]{
        \includegraphics[width=\figsz, height=\figsz]{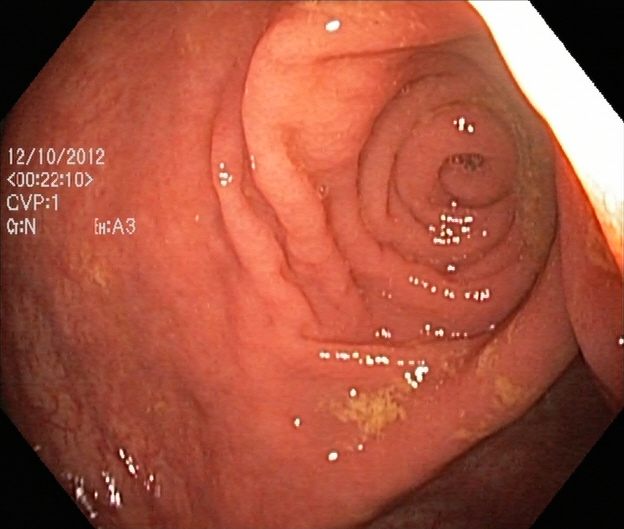}
    }
     \subfloat[][\centering\figtxt Normal pylorus]{
        \includegraphics[width=\figsz, height=\figsz]{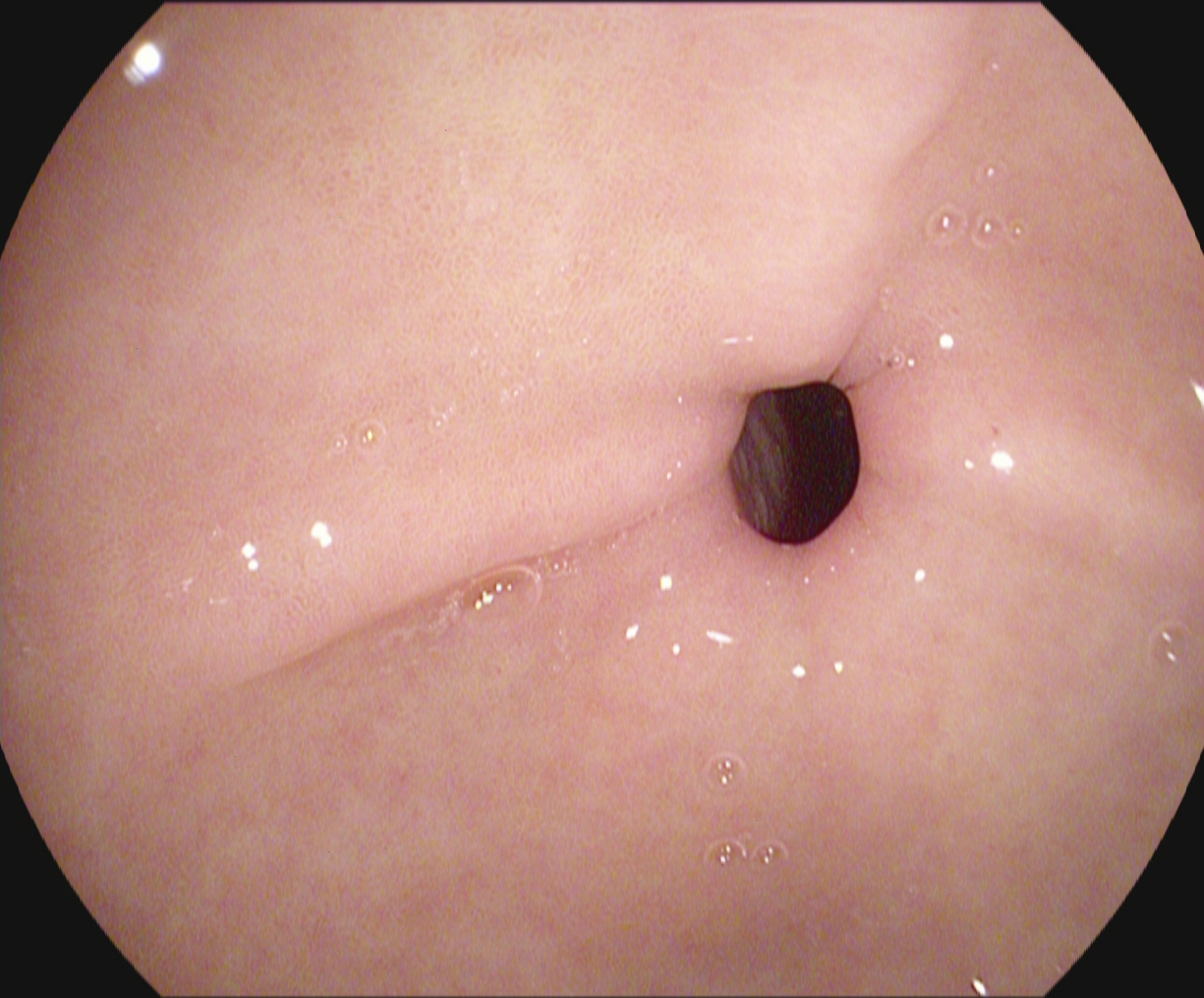}
    }
    \hspace{0mm}
     \subfloat[][\centering\figtxt Normal z-line]{
        \includegraphics[width=\figsz, height=\figsz]{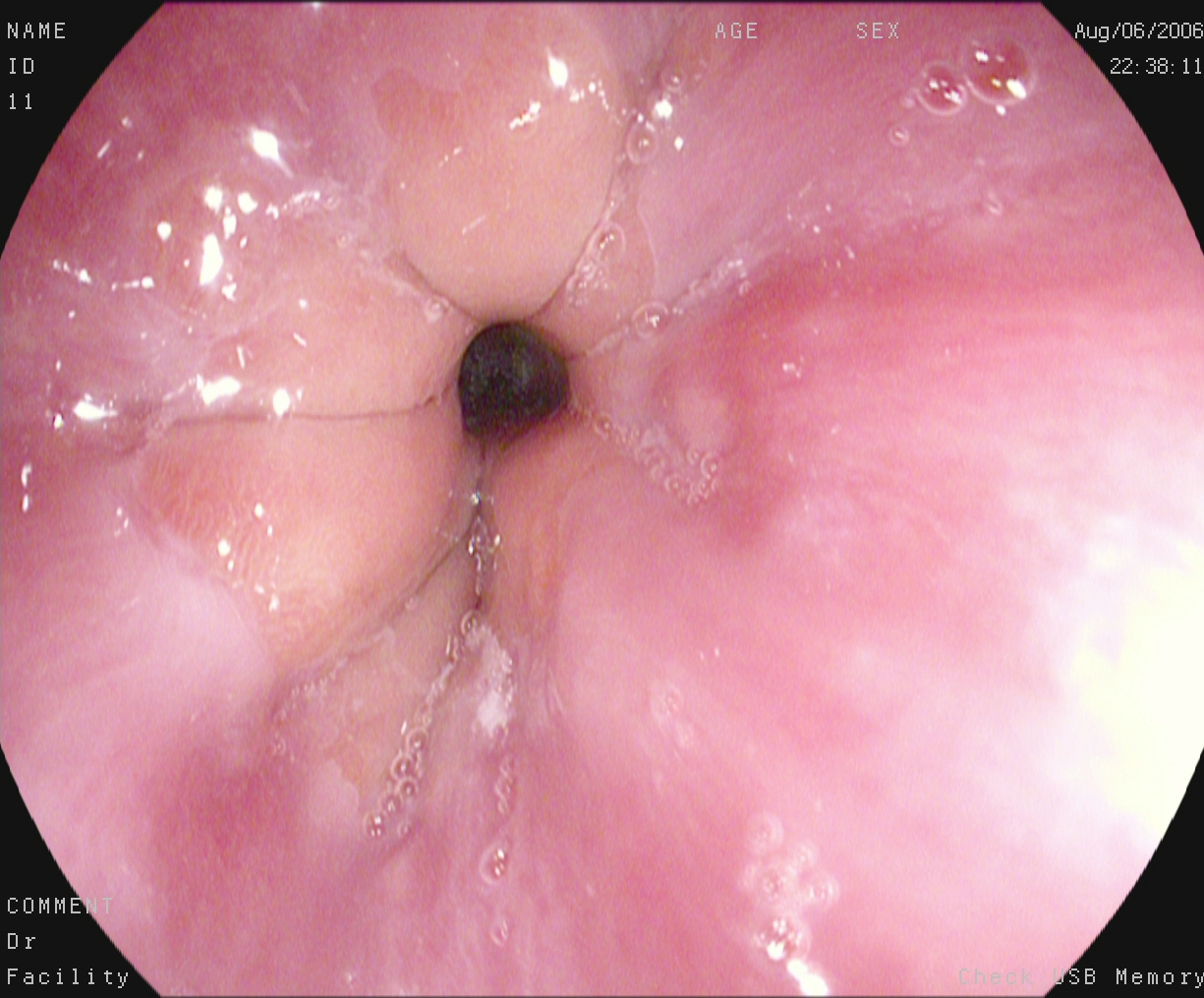}
        \label{fig:normal_z_line}
    }
     \subfloat[][\centering\figtxt Out of patient]{
        \includegraphics[width=\figsz, height=\figsz]{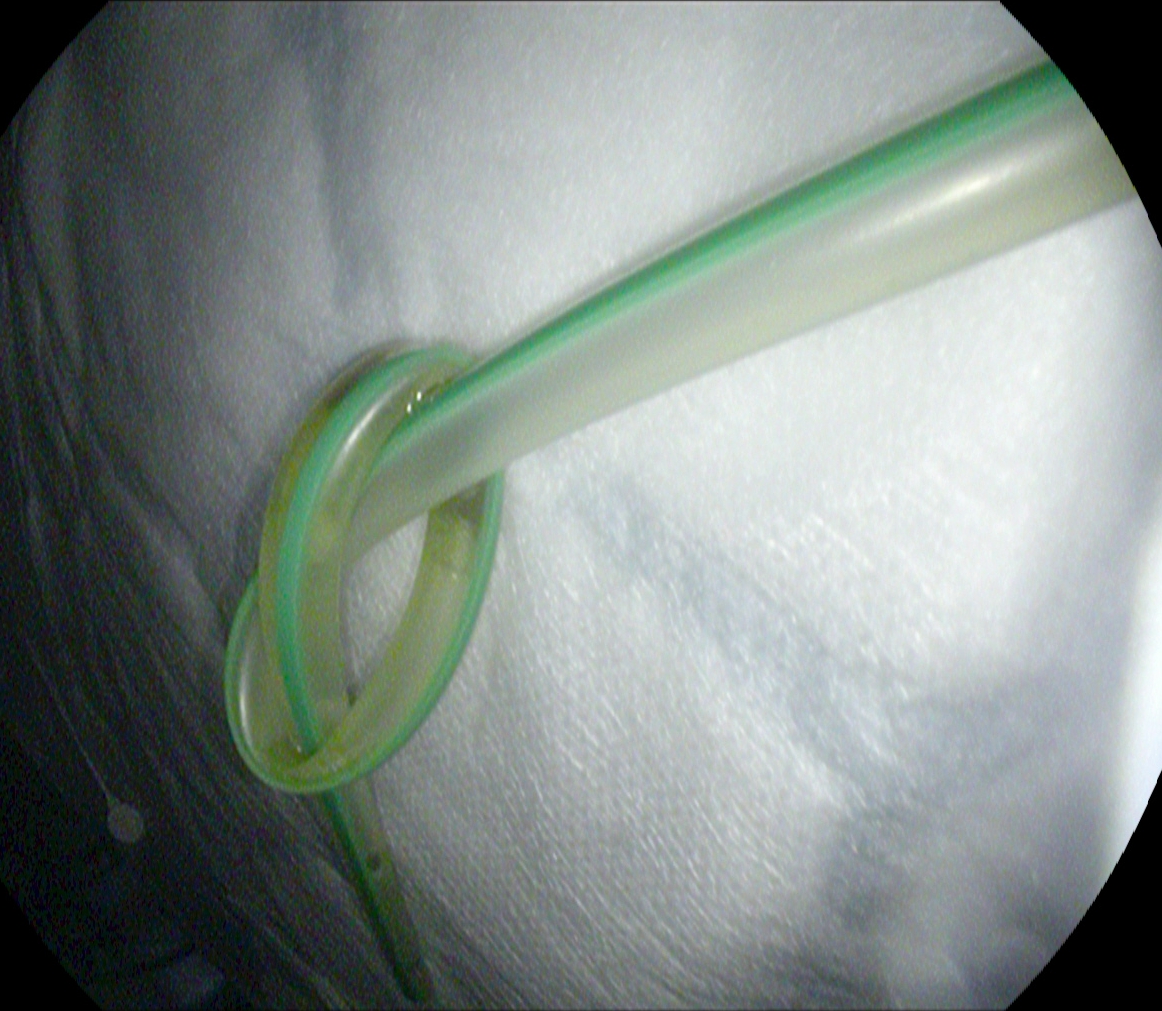}
   }
     \subfloat[][\centering Polyps]{
        \includegraphics[width=\figsz, height=\figsz]{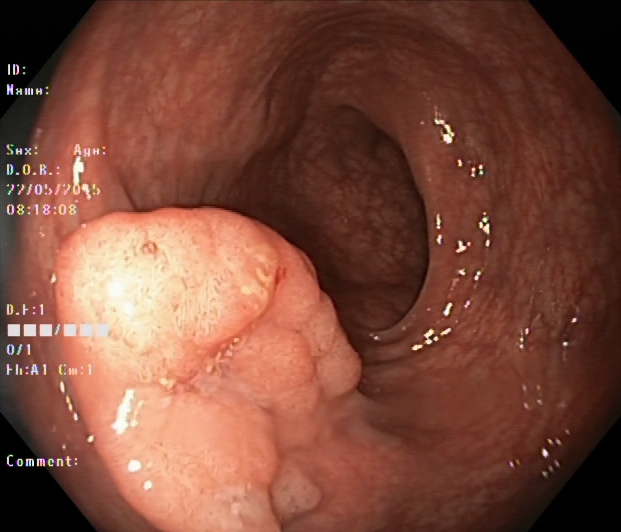}
    }
     \subfloat[][\centering\figtxt Retroflex rectum]{
        \includegraphics[width=\figsz, height=\figsz]{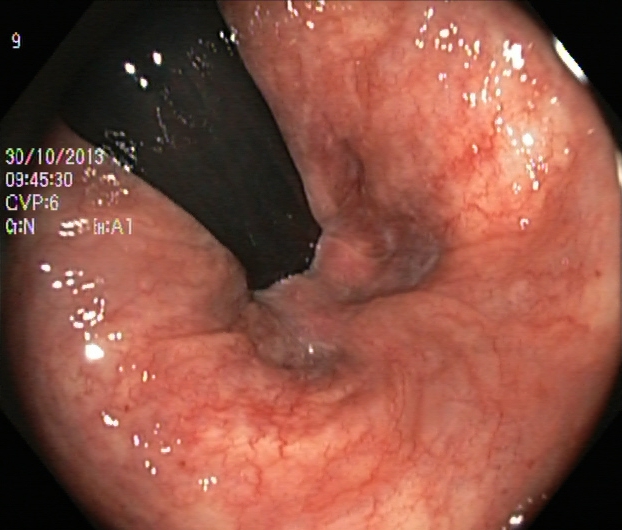}
    }
    \hspace{0mm}
     \subfloat[][\centering\figtxt Retroflex stomach]{
        \includegraphics[width=\figsz, height=\figsz]{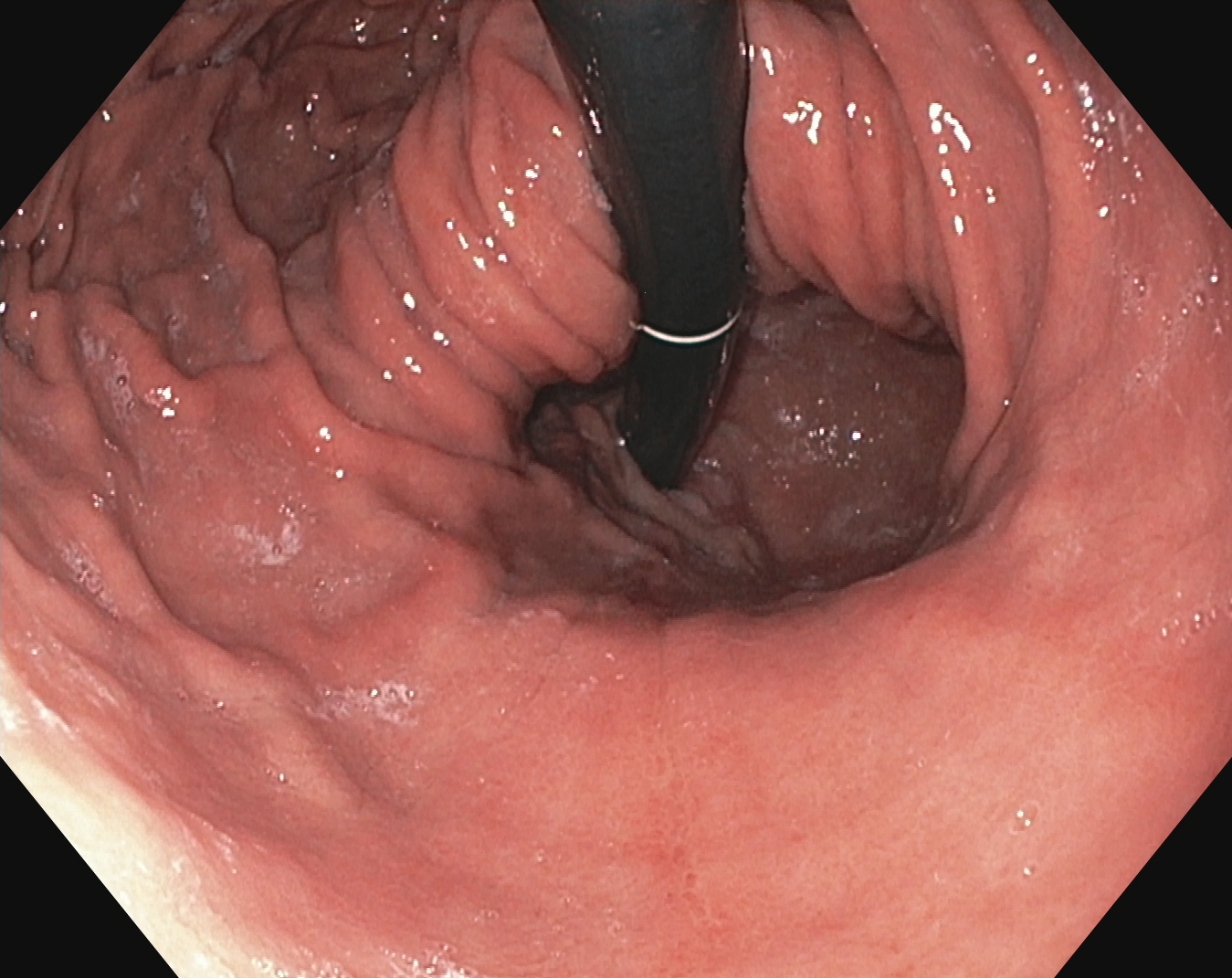}
    }
    \subfloat[][\centering\figtxt Stool inclusion]{
        \includegraphics[width=\figsz, height=\figsz]{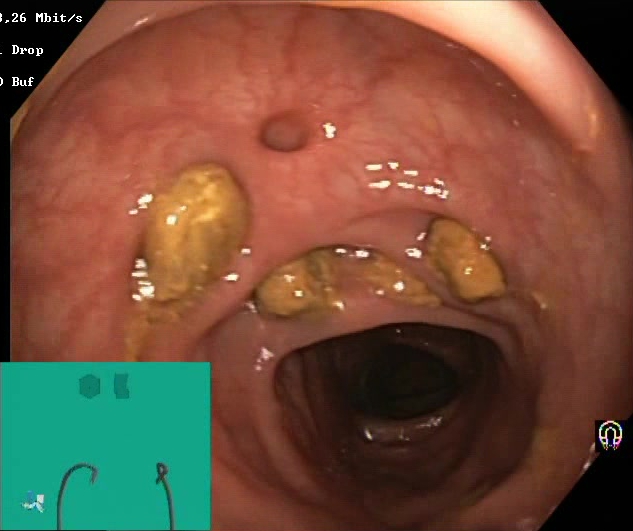}
        \label{fig:stool_inclusio}
   }
    \subfloat[][\centering\figtxt Stool plenty]{
        \includegraphics[width=\figsz, height=\figsz]{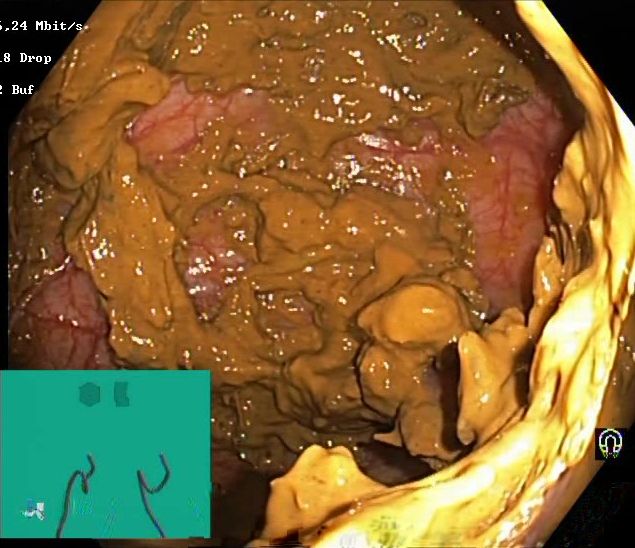}
        \label{fig:stool_prenty}
   }
    \subfloat[][\centering\figtxt Ulcerative colitis]{
        \includegraphics[width=\figsz, height=\figsz]{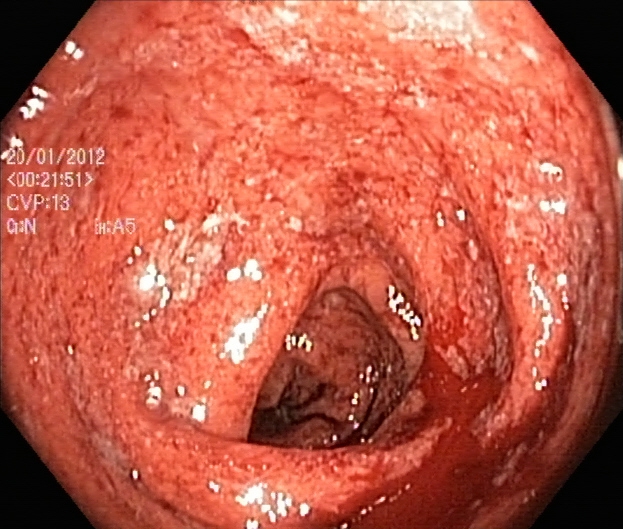}
  }
    \caption{Sample images of \ac{GI} findings. Each image represents one of the 16 classes from the dataset used for the Medico 2018 challenge~\cite{pogorelov2017kvasir,pogorelov2017nerthus}. }
    \label{fig:images_16}
    \vspace{-10pt}
\end{figure}
%%%%%%%%%%%%%%%%%%%%%%%%%%%%%%%%%%%%%%%%%%%%%%%%%

In order to study cross-dataset bias and metrics interpretation,
our contributions are as follows: 
\begin{enumerate}
    
    \item We present five \ac{ML} classification models to classify multi-class findings (anatomical landmark, pathological findings, polyp removal conditions, and normal findings)  of the \ac{GI} tract. Using a limited imbalanced dataset, we experiment with approaches ranging from \ac{GF} approaches to simple \ac{DNN} and complex \ac{DNN} approaches with transfer learning. Moreover, we present a detailed evaluation using six performance metrics to show the real classification performance of \ac{ML} models. In addition, we analyze and present detailed evaluation results of using multi-class classification \ac{ML} models for classifying binary classes (subclasses of the multi-class categories) with and without retraining to evaluate the generalizability of our models. We emphasize the difficulties of using well-performing \ac{ML} methods in cross-datasets as a result of the reluctance of \ac{ML} models to cross-dataset generalization. We present this negative impact with the aid of another evaluation using \ac{ROC} and \ac{PRC} curves of the best model. We also demonstrate when a \ac{ROC} curve is good to use and when it is better to use a \ac{PRC} curve. 
    
    \item With the above point, we emphasis the requirement of detailed cross-dataset evaluations to identify generalizability of \ac{ML} models before using them as universal models in live applications. Because good performance measures with a single dataset do not necessarily imply good real-world performance, we argue that researchers should present cross-dataset evaluations for building generalizable model rather than presenting performance values for the test datasets, which is separated from the same training data source. 
    
\end{enumerate}

Moreover, with respect to the 2018 Medico Task~\cite{pogorelov2018medico}, our best \ac{DNN} method achieved the highest \textbf{recall}, \textbf{specificity} and \textbf{accuracy} for multi-class classification of the \ac{GI} tract findings. We achieved an \ac{MCC} (0.0029 less) and an Rk correlation coefficient\footnote{The Rk correlation coefficient and the MCC were the most important considered metrics for winning the 2018 Medico Task.} (0.0001 less) nearly equal to the winning team. With this achievement, we demonstrate all the steps from designing to training and testing for reaching such performance using this model and its expandability using different pre-trained networks.

In the next section, we present related work and the performance of relevant existing solutions. Section~\ref{sec:global_feature} discusses the methodology used for our \ac{GF}-based approaches and the theoretical foundation for our work. The \ac{DNN}-based approaches are similarly described in Section~\ref{sec:deeplearning}. Our experimental results are presented and analyzed in Section~\ref{sec:results}, followed by a discussion in Section~\ref{sec:discuss} on how our results can be helpful for other researchers. Finally, in Section~\ref{sec:conclusion}, we conclude our findings. 

%============================
\section{Related Work}
\label{sec:related}

\begin{table}
\footnotesize
\caption{Overview of the related work. \textbf{REC = Recall (Sensitivity)}, \textbf{SPEC = Specificity}, \textbf{ACC = Accuracy,} \textbf{MCC = Matthews Correlation Coefficient}, 
\textbf{F1 = F1 Score}, \textbf{Rk = Rk correlation coefficient}, \textbf{FPS = Frame Per Second}. The results of the Medico Task may slightly vary compared to the proceeding note papers because of different ways of calculating the multi-class performance metrics by the organizers.}
\vspace{-10pt}

\begin{tabular}{llllllllll}
\toprule
\bf Reference &\bf Year &\bf  REC &\bf  PREC &\bf  SPEC &\bf ACC &\bf MCC    &\bf F1 &\bf Rk &\bf FPS\\ \midrule
Hwang et al.~\cite{hwang2007polyp}           & 2007 & 0.9600   & 0.8300      & -           & -        & -      & -        & -        & 15         \\ %\hline
Li et al.~\cite{li2012tumor}              & 2012 & 0.8860  & -         & 0.9620          & 0.9240    & -      & -        & -        & -          \\ %\hline
Zhou et al.~\cite{zhou2014polyp}            & 2014 & 0.7500   & -         & 0.9592      & 0.9077   & -      & -        & -        & -          \\ %\hline
Wang et al.\cite{wang2014part}             & 2014 & 0.8140  & -         & -           & -        & -      & -        & -        & 0.14       \\ %\hline
Mamonov et al.~\cite{mamonov2014automated}     & 2014 & 0.4700   & -         & 0.9000         & -        & -      & -        & -        & -          \\ %\hline
Wang et al.~\cite{wang2015polyp}            & 2015 & 0.9770  & -         & -           & 0.9570    & -      & -        & -        & 10         \\ %\hline
Riegler et al.\cite{riegler2016multimedia}    & 2016 & 0.9850  & 0.9388    & 0.7250       & 0.8770    & -      & -        & -        & $\sim$ 300        \\ %\hline
Shin et al.~\cite{shin2017comparison}       & 2017 & 0.9082 & 0.9271    & 0.9176      & 0.9126   & -      & -        & -        & -          \\ %\hline
Riegler et al.~\cite{riegler2017annotation}    & 2017 & 0.9850  & 0.9390     & 0.7250       & 0.8770    & -      & -        & -        & $\sim$ 75         \\ %\hline
Yu et al.~\cite{yu2017integrating}        & 2017 & 0.5005 & 0.4917    & -           & 0.9471   & -      & 0.4830    & 0.5357   & -          \\ %\hline
Pogorelov et al.~\cite{pogorelov2017comparison}  & 2017 & 0.8260  & 0.8290     & 0.9750       & 0.9570    & -      & 0.8260    & 0.8020    & 46         \\ %\hline
Agrawal et al.~\cite{agrawal2017scl}           & 2017 & -      & -         & -           & 0.9610    & 0.8260  & 0.8470    & -        & -          \\ %\hline
Naqvi et al.~\cite{naqvi2017ensemble}        & 2017 & -      & 0.7665    & 0.9660       & 0.9420    & 0.7360  & 0.7670    & -        & -          \\ %\hline
Petscharnig et al.~\cite{petscharnig2017inception} & 2017 & 0.7550  & 0.7550     & 0.9650       & 0.9390    & 0.7200   & 0.7550    & 0.7240    & -          \\ %\hline
Pogorelov et al.~\cite{pogorelov2017efficient}  & 2017 & 0.9060  & 0.9060     & 0.9810       & 0.9690   & -      &-    & -   & 30        \\ %\hline
Yuan et al.~\cite{yuan2018automatic}        & 2018 & 0.8180  & 0.7232    & -           & -        & -      & 0.7431   & -        & -          \\ %\hline
Wang et al.~\cite{wang2018development}  & 2018 & 0.9438  & -     & 0.9592      &    & -      &-    & -   & -  \\ %\hline
Mori et al.~\cite{mori2018detecting} & 2018 & >0.9000  & -   & >0.9000   & - & -      & -   & -  & - \\ %\hline
\midrule 
\multicolumn{9}{c}{\bf{MediaEval 2018 Medico Task}~\cite{pogorelov2018medico}} [ All the experiments below are done using 2018 Medico dataset] \\
\midrule 
Hoang et al.~\cite{RCNNTrungHieuHoang}       & 2018 & 0.9281 & \textbf{0.9426}    &\textbf{ 0.9963}      & \textbf{0.9932 }  & \textbf{0.9312} & \textbf{0.9342}   & \textbf{0.9398}   & 23    \\ %\hline
Hicks et al.~\cite{hicksdeep}                & 2018 & 0.9218 & 0.9378    & 0.9959      & 0.9924   & 0.9228 & 0.9236   & 0.9325   & 624     \\ %\hline
Borgli et al.~\cite{borgliautomatic}          & 2018 & 0.8572 & 0.8708    & 0.9956      & 0.9918   & 0.8555 & 0.8555   & 0.9280    & -          \\ %\hline
Kirker\o d et al.~\cite{kirkerodusing}                 & 2018 & 0.8433 & 0.8514    & 0.9944      & 0.9896   & 0.8366 & 0.8367   & 0.9082   & -          \\ %\hline
Dias et al.~\cite{diastransfer}             & 2018 & 0.8205 & 0.8414    & 0.9938      & 0.9885   & 0.8146 & 0.8114   & 0.8983   & 8.61     \\ %\hline
Taschwer et al.~\cite{taschwer2018early}        & 2018 & 0.8673 & 0.8826    & 0.9933      & 0.9876   & 0.8641 & 0.8662   & 0.8897   & -          \\ %\hline
Ostroukhova et al.~\cite{ostroukhovatransfer}      & 2018 & 0.8236 & 0.8281    & 0.9911      & 0.9835   & 0.8115 & 0.8145   & 0.8539   & 1E-100  \\ %\hline
Khan et al.~\cite{khanmajority}             & 2018 & 0.6203 & 0.7173    & 0.9767      & 0.957    & 0.6025 & 0.5868   & 0.6302   & 43329 \\ %\hline
Steiner et al.~\cite{NOATMichael}              & 2018 & 0.4219 & 0.5146    & 0.9717      & 0.9469   & 0.3901 & 0.3913   & 0.5368   & -          \\ %\hline
Ko et al.~\cite{TobeyWeighted}            & 2018 & 0.5005 & 0.4916    & 0.9715      & 0.9471   & 0.4608 & 0.4829   & 0.5357   & 0.5357     \\ %\hline
Thambawita et al. \textbf{(Ours)}~\cite{thambawita2018medico}     & 2018 & \textbf{0.9361} & 0.9319    & \textbf{0.9963 }     & \textbf{0.9932 }  & 0.9283 & 0.9297   & 0.9397   & -          \\ \bottomrule
\end{tabular}
 \label{tab:Overview_of_the_related_work}
 \vspace{-5mm}
\end{table}
 Many methods and algorithms have been proposed for \ac{GI} tract disease detection/classification using videos and images from colonoscopy and gastroscopy as input. The problem of polyp detection has by far received the most attention by researchers. Images and videos of polyps and other abnormalities inside the \ac{GI} tract are usually collected using a specific purpose camera and imaging system, like ScopeGuide from Olympus. 
The information gathered from these types of devices may be of great significance for later examination and must be handled with great care. Polyps generally have different characteristics to the normal surrounding healthy tissue and are often easy for clinicians to detect. There are several good datasets available for training and testing on polyps (the details about the available polyp dataset can be found in~\cite{debesh2020kvasir-seg,de2018methodology}), and binary classification methods are relatively straight forward to implement.

The other active research efforts include developing an automatic and real-time detection system for \ac{GI} bleeding, ulcerative lesion, blood-based abnormality, tumor, angiectasia, and for multi-class data of GI tract that comprise of anatomical landmarks (e.g., z-line, pylorus, and cecum), pathological findings (e.g., esophagitis and ulcerative colitis), and normality and regular findings (e.g., normal colon mucosa and stool). Suitable datasets for research in these areas are less developed and lack adequate content. Similarly, presented performance measures in these areas are not adequate because of not presenting enough performance metrics or not presenting cross-dataset evaluations. 

% \textcolor{blue}
Table~\ref{tab:Overview_of_the_related_work} gives an overview of important works related to \ac{GI} disease detection/classification and the 2018 Medico Task~\cite{pogorelov2018medico} using \ac{CAD}, from automatic polyp detection to multi-class disease detection and classification system. The dataset used for the experiments in the first half part of the Table~\ref{tab:Overview_of_the_related_work} is different. Therefore, the results can not be directly compared; however, the results on the lower half part can be compared as the algorithms are tested on the same dataset.

Most of the research in the medical field only focuses on designing an automated disease detection system for detecting or classifying specific disease or abnormality, like polyp detection or ulcer detection. Because patients may suffer from more than one type of disease at the time, a working multi-class disease detection system will help treatment. The performance of existing multi-abnormality detection systems is, however, not satisfactory and cannot assist doctors in \ac{CAD} in real-time while undergoing colonoscopies.  Furthermore, these research works have not evaluated all performance metrics at once to analyze the real behavior of their classification models. On the other hand, none of the above methods have performed cross-dataset evaluations to prove the capabilities to use the \ac{ML} models in real \ac{CAD} systems.

For \ac{HC} feature based methods, image descriptors like global or local image features (e.g., color, texture, and edges) are extracted, and later on, various \ac{ML} classifiers (for example logistic model tree ~\cite{thambawita2018medico}, random forest classifier~\cite{mamonov2014automated}, or \ac{SVM}~\cite{wang2014part}) are employed to perform analysis using these features. 
\ac{HC} descriptors (manually designed features) are useful for the gastroenterologist  while identifying specific abnormality regions inside the \ac{GI} tract. For instance, as blood has a particular range of chromaticity, we can specify a specific chromaticity range where features of bleeding abnormality seem to be concentrated~\cite{jia2017gastrointestinal}. \citet{riegler2017annotation} achieved an F1 score of 0.909 with a \ac{GF}-based approach and an F1 score of 0.875 with a \ac{DL} based approach with a multi-class \ac{GI} tract dataset. With ASU-Mayo polyp dataset, the \ac{GF}-based approach achieves an F1 score of 0.961 while the \ac{DL} based approach could make 0.936. They further suggested that the combination of both approaches may lead to improved performance. Also, the previous work by \citet{riegler2014reflects}, reveals that, while only detecting whether a frame contains an irregularity or not, \ac{GF}s can beat local features, i.e., at least reach as same results while concerning detection/classification and performs better than local features with regard to processing speed. In all of these works, researchers presented performance metrics using a test dataset selected from the same dataset used for the training data. Therefore, these results do not reflect the actual practical performance of the proposed methods. 
 
 A few past studies used the information such as color and texture of polyp to sketch \ac{HC} descriptors~\cite{ karkanis2003computer, iakovidis2005comparative, alexandre2008color, ameling2009texture, cheng2011automatic, tajbakhsh2016automated, iwahori2013automatic}. The other category of methods for automated polyp detection used shape, intensity, edge, and spatio-temporal information. For instance, \citet{hwang2007polyp} appropriated elliptical shape features to detect the occurrence of polyps in the colonoscopy videos. Bernal et al. \cite{bernal2012towards} proposed a polyp detection technique by utilizing polyp region descriptor, which is dependent on the depth of the valley image and introduced a region growing method to detect polyps in colonoscopy images. \citet{bernal2013impact} additionally used valley information and enhanced their approach by improving the polyp localization results to almost 30\%. \citet{bernal2015wm} also performed additional evaluations using valley information and demonstrated better performance, especially for smaller polyps and decreased polyp miss-rate. \citet{park2012colon} utilized the spatio-temporal features for automatic polyp detection. The recently completed related work that uses the cross-sectional profile to detect protruding polyps automatically is the polyp-detection system Polyp Alert~\cite{wang2015polyp}, which can provide near real-time feedback during colonoscopies. However, the system is limited to polyp detection and is slow for live examinations. \citet{tajbakhsh2016automated} proposed a method for automatic polyp detection from colonoscopy videos, which uses context information to remove non-polyp and shape information to localize polyp reliably. \citet{riegler2017annotation} utilized various global features and have achieved high precision and recall above 90\%. \citet{yuan2018automatic} have employed a bottom-up and top-down saliency approach for automated polyp detection. While these research works discuss improving the performance of \ac{ML} models, they have not evaluated the performance of the \ac{ML} models with cross-datasets. Then, the presented results might be influenced by data bias problems, which make restrictions to use the models in practice.  

As \ac{CNN} architectures have achieved exceptional gains in medical image and video analysis tasks, more recent work on polyp detection is mainly based on \acp{CNN}. \Citet{tajbakhsh2015automatic} proposed a 2D-\ac{CNN} method for polyp detection by learning discriminative spatial and temporal features. \Citet{yu2017integrating} have used 3D-\ac{CNN} to volumetric medical data for automated polyp detection in colonoscopy videos. \Citet{zhang2019real} suggest an enhanced~\ac{SSD} called SSD-GPNet for detecting gastric polyps, which have the potential of achieving real-time detection up to 50\, FPS using Nvidia Titan V. Furthermore, they use GPDNet~\cite{zhang2017gastric} to classify three classes of the precancerous gastric disease. 

Researchers are also comparing \ac{HC} and \ac{DL} methods. For instance, \citet{pogorelov2017efficient} and \citet{riegler2017annotation} compared several (\ac{HC} and \ac{DL}-based) localization methods. \citet{pogorelov2018deepp} evaluated their approach utilizing \ac{HC} and \ac{DL} methods on different available datasets for real-time polyp detection. Their best model with a Generative Adversarial Network (GAN) obtained detection specificity of 94\% and an accuracy of 90.9\%. The above research works present good performance for predicting polyps while \citet{pogorelov2018deepp} present evaluated results of the models with cross-datasets. However, having overlapped data sources in the cross-datasets, the shown results do not reveal the real performance in cross dataset evaluations.

The pre-trained models, along with transfer learning mechanisms, are also becoming popular because of their capability to outperform state-of-the-art algorithms even with less amount of the training data, where the limited size of the medical dataset for experiments has always been a problem to yield better results. For the detection and localization of the polyps~\cite{bernal2017comparative,tajbakhsh2016convolutional}, the pre-trained models with \ac{CNN} mechanism also achieve promising results. A comparison of \ac{DL} with global features for \ac{GI} tract disease detection has also been presented. \citet{pogorelov2017comparison} presented 17 different methods for multi-class classification of \ac{GI} tract data with the limited number of the training dataset. They have used both \acp{GF} and \ac{DL} approaches into their work. They achieved the best result with modified ResNet50 features approaching with \ac{LMT} classifier. They reached an R\textsubscript k value of 80.2\% and an F1-score of 82.6\% with 2000 training and 2000 test dataset. 

Comparing with the polyp detection approaches, the research on multi-class disease detection/classification on a complete GI tract system is minimal. However, for multi-class disease detection/classification (including polyp detection) inside the GI tract, we have listed out few contributions made in this area. For example, the authors of numerous papers \cite{petscharnig2017inception, agrawal2017scl, RCNNTrungHieuHoang, hicksdeep, borgliautomatic, kirkerodusing, diastransfer, taschwer2018early, ostroukhovatransfer, khanmajority,  NOATMichael, TobeyWeighted} have presented their approach in classifying disease inside the GI tract utilizing the Kvasir dataset and MediaEval Medico 2018 dataset. The latter is a combination of the Nerthus~\cite{nerthus2017mmsys} and Kvasir~\cite{pogorelov2017kvasir} datasets. 

\Citet{hicksdeep} show how fine-tuning a \ac{CNN} model using transfer-learning with data from different source domains affects classification performance. In their case, extending the generic ImageNet dataset with medical images from the LapGyn4 and Cataract-101 dataset, they obtained a high \ac{MCC} score of 0.9228. For the 2018 Medico Task, we proposed solutions based on \acp{GF} and \ac{DL} based methods for multi-class classification of \ac{GI} tract  findings~\cite{thambawita2018medico}. Our best model was a combination of two pre-trained networks: ResNet-152 and Densenet-161, along with a \ac{MLP}. Here, we obtained an \ac{MCC} of 94.21\%, a  F1 score of 94.58\%, and an accuracy of 99.32\%. This was one of the best results in the MediEval 2018 Medico Task Challenge. We discuss the model introduced by \citet{thambawita2018medico} in detail in this paper and reproduce similar results. Based on those models, we provide and discuss the requirement of detailed evaluations using multiple performance metrics and cross-dataset evaluations.

Recent related works show promising results in terms of evaluation metrics, i.e., both sensitivity and specificity despite various challenges (for example, difficulties arise due to a dataset obtained from different modalities). The limitation with most of the recent approaches is that they target only specific problems, like bleeding detection or polyp detection. Current systems are either (i) too narrow for a flexible, multi-disease detection/classification system; 
(ii) tested only on a limited datasets, too small to show whether the systems would work well in hospitals, 
(iii) providing low processing performance for a real-time system or ignore the system performance entirely; 
(iv) overfitting of the specific dataset can also be a problem and lead to unreliable results; or 
(v) tested using datasets that are not publicly available, making it difficult to compare the approaches with others. 

In some cases, \acp{GF}-based approaches produce better results. For some methods, deep-learning performs better. The \ac{CNN} approaches and pre-trained network with transfer-learning mechanism approaches have the best results in most of the cases. Reusing already existing \ac{DL} architectures and pre-trained models leads to excellent results in, for example, the ImageNet classification tasks. For example, the \ac{HC} feature based approach works well for \ac{TN} detection/classification tasks.

To reduce the damage of dataset bias problem, \citet{ECCV12_Khosla}
directed their experiments for classification tasks as well as detection problems. They used different datasets from different domains in the training stage to generalize the features extracted from their~\ac{ML} model. However,~\ac{SVM} was used as the main algorithm, and the~\ac{DNN} dataset bias problem was not addressed. 

With the goal of making researchers aware of the dataset bias problems, \citet{5995347} did informative research using basic datasets and basic \ac{ML} models with classification and detection task of computer vision. Initially, \citeauthor{5995347} trained a simple linear \ac{SVM} to make a simple classifier to name a given dataset from 12 different datasets, which are having closely the same categories. They have been inspired by the research done by \citet{dollar2009pedestrian} to detect pedestrians. The result of the experiment for dataset classification shows a clear diagonal in the confusion matrix. This implies that there are clear dataset bias features, while these datasets have the same categories. Therefore, researchers want to apply cross-dataset generalization for avoiding dataset bias behavior of \ac{ML} models. Moreover, they have discussed selection bias, capture bias, category or label bias, and negative bias as the main factors for the dataset bias. This directs our research to do additional experiments to identify the significant factors of the cross-domain data generalization in the medical domain, which is more critical than the general image classification.

The classification of \ac{GI} diseases is more complicated than a simple real-world object classification task where one detects faces or recognize characters. Typical \ac{GI} tract datasets are heavily imbalanced, e.g., the dataset of the 2018 Medico Task consists of 16 classes of anatomical landmarks, pathological findings, polyp removal cases, and normal and regular findings where the polyp class has a maximum of 613 images, and the instrument class has a minimum of only 4 images. Additionally, medical datasets are captured using different endoscopic instruments, and some of the images can be noisy, be blurry, be over-or under-exposed, be interleaved, have superfluous information within the image, contain borders, and be affected by specular reflections caused by instrument light source. Some of the images may have bleeding, while other images can be partially covered by stool or mucus. Moreover, the organs from mouth to anus can have multiple lesions showing different diseases, abnormalities, and internal injuries. Thus, the above situation leads to the necessity of distinguishing between various classes of \ac{GI} tract findings. In this scenario, not only high precision, and recall, but also high accuracy and \ac{MCC} becomes essential for developing an automated generalizable multi-class classification system. This implies the real requirement of measuring and analyzing all performance metrics at once. Furthermore, to prove the generalizability of models, cross-dataset evaluations are required.

%===============================

\section{Global Feature based Approaches}
\label{sec:global_feature}

\begin{figure}
\resizebox{!}{3cm}{%
\begin{tikzpicture}[
    in_out_node/.style={rectangle, draw=black!60, fill=black!5, very thick, minimum size=10mm, rounded corners=3mm},
    model_node/.style={rectangle, draw=blue!60, fill=blue!5, very thick, minimum height=2cm, minimum width=2cm, text width=2cm,align=center},
    model_out/.style={rectangle, draw=red!60, fill=red!5, very thick, minimum height=3cm, minimum width=1cm},
    normal_node/.style={rectangle, draw=black!60, fill=black!5, very thick, minimum size=10mm, rounded corners=3mm, text width=2cm,align=center},
]

\node[in_out_node]  (input)     {input};
\node[model_node, draw=green!60, fill=green!5]   (gf_extraction) [right=0.4cm of input]    {Global Feature Extraction (LIRE)};
\node[normal_node]  (gfs)   [right=0.4cm of gf_extraction]  {Extracted Features (6)};

\node[model_node]  at (8,1.2) (main_model_1)     {Simple Logistic Model};
\node[model_node] at (8,-1.2)   (main_model_2) {Logistic Model Tree};
\node[normal_node]  (output_1)   [right=0.4cm of main_model_1]  {output (Method 1)};
\node[normal_node]  (output_2)   [right=0.4cm of main_model_2]  {output (Method 2)};
%\node[model_out]    (model_out) [right=of model_resnet] {16};
%\node[out_only_node]  (output)   [right=of model_out] {output (method3)};
%\node[below=0.2cm of model_resnet] {Base Network};
%\node (e1) [below left of=model_out] {};
%\node (e2) [below left of=model_resnet] {};

%arrows
\draw[->] (input.east) -- (gf_extraction.west);
\draw[->] (gf_extraction.east) -- (gfs.west);
\draw[->] (gfs.east) -- (main_model_1.west);
\draw[->] (gfs.east) -- (main_model_2.west);
\draw[->] (main_model_1.east) -- (output_1.west);
\draw[->] (main_model_2.east) -- (output_2.west);
%\draw[->] (e1) -- node[anchor=south] {loss} (e2);

\end{tikzpicture}
}
\caption{Block diagram of the proposed method 1 and method 2. The pipeline start with the input of images. Global features are extracted using the LIRE framework. These features are then used for two different classification algorithms (simple logistic model for method 1 and logistic model tree for method 2).}
\label{fig:blockDiagram_method_1_2}
\vspace{-10pt}
\end{figure}
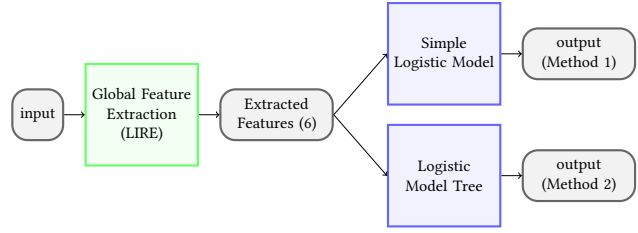

\acfp{GF} or descriptors are features computed over the whole image or covering a regular sub-section of an image. \acp{GF} represents the overall properties of an image and are often used in image retrieval, image compression, image classification, object detection, and image collection search and distance computing~\cite{pogorelov2017comparison}. Examples of \acp{GF} are shape matrix, Histogram Oriented Gradients (HOGs), Co-HOG and invariant moments (Hu, Zernike).  The LIRE~\cite{lux2016lire} framework can be used to extract \ac{HC} \acp{GF} like texture, color distribution, and the histogram of brightness. The most commonly used \acp{GF} include Joint Composite Descriptor (JCD), Tamura, Color Layout (CL), Edge Histogram (EH), Auto Color Correlogram (ACC), Pyramid Histogram of Oriented Gradients (PHOG), Color and Edge Directivity Descriptor (CEDD), CL, Local Binary Patterns, and Scalable Color (SC). Figure~\ref{fig:blockDiagram_method_1_2} shows the architecture of the proposed \ac{GF}-based methods (1 and 2). These methods use six selected \acp{GF} and the best \ac{ML}  classifiers for the provided dataset.

Feature engineering is among the most crucial and challenging part for approaching any \ac{ML} and computer vision problem.  Based on the findings of \citet{pogorelov2017comparison} and \citet{riegler2017multimedia}, we choose to used JCD, Tamura, CL, EH, ACC, and PHOG. The combinations of these features represent the overall properties of the images. We can even add more \acp{GF}, but it may increase the noise to the image features, which again would hurt the classification performance.
Moreover, we have formulated the problem of \ac{GI} tract anomaly classification as a multi-class (sixteen-class) classification of different findings including anomalies, landmarks and clinical markings. With the provided dataset, we have computed the \acp{GF} of each image. A multi-class classification problem is a general and well-studied \ac{ML} problem, and there are a variety of methods available to solve this issue with higher performance. Therefore, we have sent the extracted \acp{GF} to many available \ac{ML} classifiers. The whole experiment was completed with the development dataset. The 2018 Medico Task~\cite{pogorelov2018medico} shows best classification rates with Simple Logistic (SL)~\cite{landwehr2005logistic} and Logistic Model Tree (LMT)~\cite{landwehr2005logistic} classifiers.

%---------------------------------------------------------------------

%---------------------------------------------------------------------
\subsection{Method 1: The SimpleLogistic classifier}

In method 1, we combine the \ac{SL} classifier from the Weka software~\cite{hall2009weka} to build a linear logistic regression model with the LogitBoost~\cite{friedman2000additive} utility for determining attributes. 
The \ac{SL} classifier can deal with both binary class classification, multi-class classification, missing class, and nominal class. It can handle different types of attributes such as binary attributes, nominal attributes, date attributes, missing values, unary attributes, and empty nominal attributes~\cite{landwehr2005logistic}. In a linear logistic regression classifier, a simple (linear) model fits the data, and the method of model fitting is pretty stable, leading to low variances. 

LogitBoost is utilized for determination of the most appropriate attributes in the data at the time of executing logistic regression, which is done by performing a simple regression in every iteration before it converges to a solution of maximum likelihood. Therefore, LogitBoost, with a simple regression function that acts as a base learner, is utilized for fitting the logistic models. The optimum number of iterations associated with the LogistBoost algorithm to function is cross-validated, which leads to the automatic selection of the attribute~\cite{sumner2005speeding}. 
The \ac{SL}  classifier has a built-in attribute selection (if the default parameter is not changed): it stops computing Simple Linear Regression models (i.e., performing LogitBoost iterations) when the cross-validated classification error no longer decreases. With the extracted features using LIRE, the \ac{SL} classifier has not only the highest classification accuracy, but also it takes lowest classification time (i.e., lowest computational complexity) when compared with other \ac{ML} classification algorithms. 

\subsection{Method 2: The Logistic model tree }

In method 2, we use the \ac{LMT} classifier from the Weka software. The \ac{LMT} is a classification model related to a supervised training algorithm, which is a combination of Logistic Regression (LR) and decision tree learning techniques~\cite{landwehr2005logistic, salzberg1994c4}. Thus, the \ac{LMT} is considered an analogue model for solving classification problems. In the logistic variant, information gain is utilized for splitting, the LogitBoost algorithm generates an LR model at each node in the tree, and the CART algorithm \cite{salzberg1994c4} is utilized for pruning the tree.    

The \ac{LMT} uses a cross-validation (CV) technique to find several LogitBoost iterations to  prevent overfitting of the training data. The LogistBoost algorithm accomplishes additive logistic regression which is achieved by least-squares fits for every class M, \cite{doetsch2009logistic} which is shown in Equation~\ref{eq:logistic model tree}:
\begin{equation}
\label{eq:logistic model tree}
    L_{M}(x) = \sum_{i=1}^{n}{\beta_i}+ {\beta}_{0}
\end{equation}
Here, $\beta_i $ denotes the coefficient of the \textit{i}th component of the vector \textit{x}, and \textit{n} denotes number of features. The \ac{LMT} model uses the Linear logistic regression method to calculate the posterior probabilities of the leaf nodes~\cite{landwehr2005logistic}  which is shown in Equation~\ref{eq:Logistic model tree1}:
\begin{equation}
\label{eq:Logistic model tree1}
     P\left({M}\mid X\right) = -\frac{\exp\left(L_{M}(X)\right)} {\sum_{M=1}^{D}\exp \left(L_{M}(X)\right)}
\end{equation}
%%%%%%%%%%%%%%%%%%%%%%%%%%%%%%%%%%%%%%%%%%%%%%%%%

Here, \textit{D} denotes the number of classes, and L \textsubscript{M}(X) stands for the least-square fits. The least-square fits L \textsubscript{M}(X) are transformed in such a way that ${\sum_{M=1}^{D}\exp (L_{M}(X)}\big)$
is equal to zero. 

%==============================

\section{Deep Learning Approaches}
\label{sec:deeplearning}

For our transfer-learning approaches, we selected two \acp{DNN}: ResNet-152~\cite{resnet} and Densenet-161~\cite{densenet} based on the top-1-error rate and top-5-error rate for the ImageNet~\cite{imagenet_1, imagenet_2} classification as given in the Pytorch documentation~\cite{pytorch_models}. Then, we chose ResNet-152 as the base model of the first \ac{DL} approach and this base model experiment has been done under the method 3 (the model is illustrated in Figure~\ref{fig:deep_learning_flow_basic}). This selection has been made based on preliminary experiments. In the preliminary experiments, the ResNet-152 showed better performance than Densenet-161. This Densenet-161 was in the second place in the performance ranking when we compared stand-alone pre-trained \ac{DL} models.

In \ac{DL} methods 4 and 5 (as illustrated in figures \ref{fig:deep_learning_flow_averaging} and \ref{fig:deep_learning_flow_mlp}), we have used both pre-trained ResNet-152 and Densenet-161 using the ImageNet dataset. In the following sections, we discuss data pre-processing mechanisms and training mechanisms used for all three \ac{DL} methods.  In the later sections, we discuss these methods one by one with its fine-tuning mechanisms with more comprehensive explanations. 

For the transfer learning methods, we use the data preprocessing tool of Pytorch library to 1) resize input images, 2) crop marginal annotations of the medical images, 3) normalize the pixel values of input images, and 4) apply random image transformations. Regarding image resizing, all images of the dataset were resized into $224 \times 224$ because ResNet-152 and Densenet-161 accept images with these dimensions. By applying the central-cropping transformation of Pytorch, we minimized unnecessary effects for the final predictions of \acp{DNN} affected from annotated marks (green boxes) of the medical images as shown in Figures~\ref{fig:colon_clear},~\ref{fig:stool_inclusio}, and~\ref{fig:stool_prenty}.
Center cropping did not remove important information from the images because we cropped down to $ 224 \times 224 $ from $ 256 \times 256 $. 
Our experiments show that removing the whole green box, such as ones in Figures \ref{fig:colon_clear}, \ref{fig:stool_inclusio}, and \ref{fig:stool_prenty}, from the images by applying a larger crop size is not advisable because, for some images, too much content of the finding is lost with a large crop size.
When applying the normalization function to the input images, a standard deviation ($\sigma$) of $0.5$ and a mean ($\mu$) of $0.5$ were used with the normalization function in the Pytorch. The mathematical equation used in this function is given in Equation~\ref{eq:normalization} and $c$ represents the three channels R, G, and B of input images. The $input$ represents a tensor of pixel values of each layer.  We have used random transformations, random horizontal flips, random vertical flips, and random rotations from Pytorch as data augmentation techniques.

\begin{equation}
\label{eq:normalization}
    input_c = \frac{input_c - \mu_c}{\sigma_c}; \quad where \quad c= [0,1,2]
   % \vspace{-5pt}
\end{equation}

For training all \acp{DNN}, the transfer-learning mechanism was used. Then, we used the cross-entropy loss~\cite{cross_entropy_1} with weighted classes as given in Equation~\ref{eq:cross_entroy_weighted} to calculate the loss values of the \acp{DNN}:
\begin{equation}
\label{eq:cross_entroy_weighted}
    loss(x, class) = weight[class]\times\left( -x[class] + 
    \ln\left(\sum_j \exp( x[j] )  \right)\right)
\end{equation}
In this equation, the weight parameter value is calculated inversely proportional to the image count in the corresponding class. In other words, class weight values are high when the classes have less number of images. However, the inbuilt cross-entropy function given in the Pytorch is used instead of implementing it from scratch. While doing preliminary experiments, we observed that there was not any effect from weighted cross-entropy loss. Then, we used the normal cross-entropy loss (Equation~\ref{eq:cross_entropy}) function for calculating the loss of the \acp{DNN}: 
% Equations - cross-entropy with weight and 
\begin{equation}
\label{eq:cross_entropy}
    loss(x, class) = -\ln\left(\frac{\exp(x[class])}{\sum_j\exp(x[j])}\right) = -x[class] + \ln\left(\sum_j \exp{(x[j])}\right)
\end{equation}
As the optimizer of all the \acp{DNN}, the stochastic gradient descent~\cite{sgd} method with a momentum~\cite{sgd-momentum} was applied.  We selected this optimizer because of its stable learning mechanism in contrast with the highly unstable learning pattern of other methods~\cite{adadelta, adam, optimizations_overview} while they show fast convergence. 
%%%%%%%%%%%%%%%%%%%%%%%%%%%%%%%%%%%%%%%%%%%%%%%

During the training procedure, we changed the learning rate manually based on the progress of learning curves rather than using the inbuilt learning rate schedulers of Pytorch. Initially, we began with a high learning rate. Then, the learning rate is reduced by the factor of 10 if the training process did not show good progress in the learning curves.  Finally, model weights of the best epoch based on the best validation accuracy were saved to use in the inference stage. 

% Write about method 3
\subsection{Method 3: \ac{DNN} approach based on ResNet-152 }

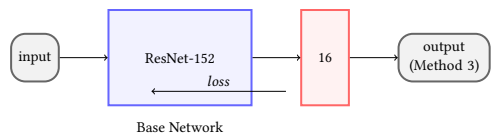
\begin{figure}
\resizebox{!}{1.7cm}{%
\begin{tikzpicture}[
    in_out_node/.style={rectangle, draw=black!60, fill=black!5, very thick, minimum size=10mm, rounded corners=3mm},
    model_node/.style={rectangle, draw=blue!60, fill=blue!5, very thick, minimum height=2cm, minimum width=3cm},
    model_out/.style={rectangle, draw=red!60, fill=red!5, very thick, minimum height=2cm, minimum width=1cm},
    out_only_node/.style={rectangle, draw=black!60, fill=black!5, very thick, minimum size=10mm, rounded corners=3mm, text width=1.7cm,align=center},
]

\node[in_out_node]  (input)     {input};
\node[model_node]   (model_resnet) [right=of input]    {ResNet-152};
\node[model_out]    (model_out) [right=of model_resnet] {16};
\node[out_only_node]  (output)   [right=of model_out] {output (Method 3)};
\node[below=0.2cm of model_resnet] {Base Network};
\node (e1) [below left of=model_out] {};
\node (e2) [below left of=model_resnet] {};

%arrows
\draw[->] (input.east) -- (model_resnet.west);
\draw[->] (model_resnet.east) -- (model_out.west);
\draw[->] (model_out.east) -- (output.west);
\draw[->] (e1) -- node[anchor=south] {$loss$} (e2);

\end{tikzpicture}
}
\caption{Block diagram of method 3. The input is an image which is passed to a ResNet-152 neural network. A final softmax layer outputs the scores for the $16$ classes.}
\label{fig:deep_learning_flow_basic}
\vspace{-10pt}
\end{figure}

Method 3  is the base method that uses only ResNet-152. A block diagram of this is illustrated in Figure~\ref{fig:deep_learning_flow_basic}. In this method, the last layer of ResNet-152 has been modified to output 16 classes of the 2018 Medico Task from 1000 classes of the ImageNet. Usually, we freeze first layers (there is not a logical way to select the number of layers to freeze) of pre-trained networks when we do transfer learning. Then, we train the last and the new layers using the new domain data. Finally, the entire network is trained after unfreezing all parameters of the network (a method known as fine-tuning).

We performed preliminary experiments to identify the influence of the above described freezing-unfreezing technique compared to using simple fine-tuning. Both techniques showed the same performance at the end of the training process, and we could not gain any performance benefit from the freezing-unfreezing method while using the simple fine-tuning method was faster. Therefore, we decided to use the simple fine-tuning method for all experiments.

In this method 3, we started the training process with a learning rate of 0.001. Then, the learning rate was decreased by a factor of 10 if we could not see any performance improvement for the validation dataset. We repeated this change of learning rate until the model came to a good stable position. In this experiment, the stochastic gradient descent method was used as the optimization method with a momentum of 0.9.

%\todo{Write about method 4}
\subsection{Method 4: \ac{DNN} approach based on ResNet-152 and Densenet-161 }

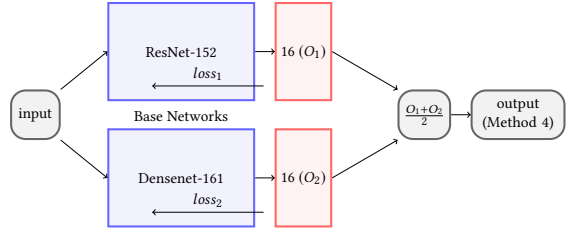
\begin{figure}
\resizebox{!}{3cm}{%
\begin{tikzpicture}[
    in_out_node/.style={rectangle, draw=black!60, fill=black!5, very thick, minimum size=10mm, rounded corners=3mm},
    model_node/.style={rectangle, draw=blue!60, fill=blue!5, very thick, minimum height=2cm, minimum width=3cm},
    model_out/.style={rectangle, draw=red!60, fill=red!5, very thick, minimum height=2cm, minimum width=1cm},
    out_only_node/.style={rectangle, draw=black!60, fill=black!5, very thick, minimum size=10mm, rounded corners=3mm, text width=1.7cm,align=center},
]

% Resnet 152
\node[in_out_node]  at (0,0) (input)     {input};
\node[model_node]  at (3,1.3) (model_resnet)   {ResNet-152};
\node[model_out]    (model_out_1) [right=0.4cm of model_resnet] {16 $(O_1)$};

\node[below=0.08cm of model_resnet] {Base Networks};
\node (e1) [below left of=model_out_1] {};
\node (e2) [below left of=model_resnet] {};

%Densenet-161
\node[model_node]  at (3, -1.3) (model_densenet)    {Densenet-161};
\node[model_out]    (model_out_2) [right=0.4cm of model_densenet] {16 $(O_2)$};

\node (e3) [below left of=model_out_2] {};
\node (e4) [below left of=model_densenet] {};

%out nodes
\node[in_out_node]  at (8,0) (output_1)   {$\frac{O_1+O_2}{2}$};
\node[out_only_node]  (output_2)  [right=0.4cm of output_1] {output (Method 4)};

%arrows
\draw[->] (input.north east) -- (model_resnet.west);
\draw[->] (input.south east) -- (model_densenet.west);

% arrows Resnet-152
\draw[->] (model_resnet.east) -- (model_out_1.west);
\draw[->] (model_densenet.east) -- (model_out_2.west);
\draw[->] (model_out_1.east) -- (output_1.north west);
\draw[->] (output_1.east) -- (output_2.west);
\draw[->] (model_out_2.east) -- (output_1.south west);
\draw[->] (e1) -- node[anchor=south] {$loss_1$} (e2);
\draw[->] (e3) -- node[anchor=south] {$loss_2$} (e4);

\end{tikzpicture}
}
\caption{Block diagram of method 4. The input image is in parallel passed to a ResNet-152 and a Densenet-161 neural network. Two separate softmax layers calculate separate $16$-class scores, which are finally combined.}
\label{fig:deep_learning_flow_averaging}
\vspace{-10pt}
\end{figure}

In method 4, as illustrated in Figure~\ref{fig:deep_learning_flow_averaging}, we used two pre-trained networks on ImageNet: ResNet-152 and Densenet-161. These networks were retrained separately into the Medico dataset using the same procedure used in method 3. Before this retraining, the networks were modified to classify the 16 classes. Then, we calculated an average probability of the two probability vectors ($V_{Resnet\_152}$ and $V_{Densenet\_161}$) output by the two separate networks: ResNet-152 and Densenet-161.  By calculating the average of these two probability vectors $(V_{answer} = {}^1{\mskip -5mu/\mskip -3mu}_2 \left(V_{Resnet\_152} + V_{Densenet\_161}\right)$, we accepted cumulative probability decision rather than the individual decision. Using the average from these two networks, we expected to have a good decision with high confidence. For example, if the two networks return high probability values for the same class, the class probability value (confidence of classifying to that class) is high. On the other hand, when one network has a high probability, and the other network has a low probability for a specific class, then the final probability value is around $0.5$. This value infers that confidence about the particular class is not good enough for the final decision. 

In this model, the probability of the final answer depends on the average values rather than the highest probability value returned from one of the two models. Here, the problem is that the prediction suggested from the highest probability value of one model may be the correct class compared to the selected category from the average.  Finally, we trained the model using a learning rate of 0.001.  Also, we decreased the learning rate by factor 10, when the model did not show convergence. A momentum of 0.9 with \ac{SGD} was used as same as method 3.

%%%%%%%%%%%%%%%%%%%%%%%%%%%%%

%%%%%%%%%%%%%%%%%%%%%%%%%%%%%%%%%%

%\todo{write about method 5}
\subsection{Method 5: \ac{DNN} approach based on ResNet-152, Densenet-161 and MLP}
\label{sec:method_5}

\begin{figure}
\resizebox{!}{3cm}{%
\begin{tikzpicture}[
    in_out_node/.style={rectangle, draw=black!60, fill=black!5, very thick, minimum size=10mm, rounded corners=3mm},
    out_only_node/.style={rectangle, draw=black!60, fill=black!5, very thick, minimum size=10mm, rounded corners=3mm, text width=1.7cm,align=center},
    model_node/.style={rectangle, draw=blue!60, fill=blue!5, very thick, minimum height=2cm, minimum width=3cm},
    model_out/.style={rectangle, draw=red!60, fill=red!5, very thick, minimum height=2cm, minimum width=1cm},
    fc/.style={rectangle, draw=green!60, fill=green!5, very thick, minimum height=2cm, minimum width=1cm},
]

% Resnet 152
\node[in_out_node]  at (0,0) (input)     {input};
\node[model_node]  at (3,1.3) (model_resnet)   {ResNet-152};
\node[model_out]    (model_out_1) [right=0.4cm of model_resnet] {16 $(O_1)$};

\node[below=0.08cm of model_resnet] {Base Networks};
\node (e1) [below left of=model_out_1] {};
\node (e2) [below left of=model_resnet] {};

%Densenet-161
\node[model_node]  at (3, -1.3) (model_densenet)    {Densenet-161};
\node[model_out]    (model_out_2) [right=0.4cm of model_densenet] {16 $(O_2)$};

\node (e3) [below left of=model_out_2] {};
\node (e4) [below left of=model_densenet] {};

%out nodes
\node[fc] at (7.5,0) (fc1) {fc1};
\node[fc] (fc2) [right=0.4cm of fc1] {fc2};
%\node[in_out_node]  (output_1) [right=0.4cm of fc2]  {$\frac{O_1+O_2}{2}$};
\node[out_only_node]  (output_2)  [right=0.4cm of fc2] {output \\(Method 5)};

%arrows
\draw[->] (input.north east) -- (model_resnet.west);
\draw[->] (input.south east) -- (model_densenet.west);

% arrows Resnet-152
\draw[->] (model_resnet.east) -- (model_out_1.west);
\draw[->] (model_densenet.east) -- (model_out_2.west);
\draw[->] (model_out_1.east) -- (fc1.north west);
%\draw[->] (output_1.east) -- (output_2.west);
\draw[->] (model_out_2.east) -- (fc1.south west);
\draw[->] (e1) -- node[anchor=south] {$loss_1$} (e2);
\draw[->] (e3) -- node[anchor=south] {$loss_2$} (e4);
\draw[->] (fc1) -- (fc2);
\draw[->] (fc2) -- (output_2);

% arrow Densenet-161

\end{tikzpicture}
}
\caption{Block diagram of method 5. It is similar to method 4, but instead of a single step to combine the output scores of the two neural networks, two fully connected layers are utilized.}
\label{fig:deep_learning_flow_mlp}
\vspace{-10pt}
\end{figure}
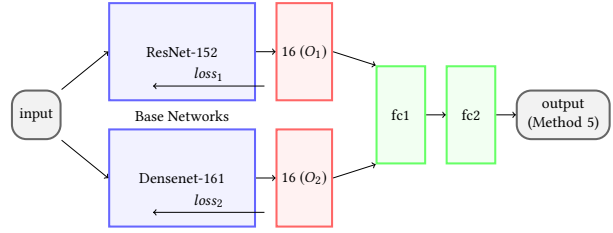

Method 5 was designed to overcome the problem of method 4. The block diagram of this method is illustrated in Figure~\ref{fig:deep_learning_flow_mlp}. The simple averaging method was not enough to make a final decision when the two networks provide two different answers.  As a solution, a \ac{MLP} was introduced instead of the simple averaging method. Then, we trained only this \ac{MLP} with the pre-trained ResNet-152, and Densenet-161 for the Medico dataset to decide the final prediction based on the probabilities that come from two networks. More details about designing this complex model are discussed under Sections ~\ref{subsec:extendible_idea} and~\ref{subsec:method5_more}.

%%%%%%%%%%%%%%%%%%%%%%%%%%%%%

%%%%%%%%%%%%%%%%%%%%%%%%%%%

\subsubsection{Extendable method 5}
\label{subsec:extendible_idea}

\begin{figure}
\resizebox{!}{5cm}{%
\begin{tikzpicture}[
    in_out_node/.style={rectangle, draw=black!60, fill=black!5, very thick, minimum size=10mm, rounded corners=3mm},
    out_only_node/.style={rectangle, draw=black!60, fill=black!5, very thick, minimum size=10mm, rounded corners=3mm, text width=1.7cm,align=center},
    model_node/.style={rectangle, draw=blue!60, fill=blue!5, very thick, minimum height=2cm, minimum width=3cm},
    model_out/.style={rectangle, draw=red!60, fill=red!5, very thick, minimum height=2cm, minimum width=1cm},
    fc/.style={rectangle, draw=green!60, fill=green!5, very thick, minimum height=2cm, minimum width=1cm},
]

% Resnet 152
\node[in_out_node]  at (0,0) (input)     {input};
\node[model_node]  at (3,2) (model_resnet)   {$net_1$};
\node[model_out]    (model_out_1) [right=0.4cm of model_resnet] {$P_{O_1}$};

\node[below=0.7cm of model_resnet] {Base Networks};
\node (e1) [below left of=model_out_1] {};
\node (e2) [below left of=model_resnet] {};

%Densenet-161
\node[model_node]  at (3, -2) (model_densenet)    {$net_N$};
\node[model_out]    (model_out_2) [right=0.4cm of model_densenet] {$P_{O_N}$};

\node (e3) [below left of=model_out_2] {};
\node (e4) [below left of=model_densenet] {};

%out nodes
\node[fc] at (7.5,0) (fc1) {$fc_1$};
\node[fc] (fc2) [right=0.4cm of fc1] {$fc_M$};
%\node[in_out_node]  (output_1) [right=0.4cm of fc2]  {$\frac{O_1+O_2}{2}$};
\node[out_only_node]  (output_2)  [right=0.4cm of fc2] {$P_{out}$};

\node (text1) at (10.5, -3.75) {Post-training};
\node (text1) at (5, -3.3) {Pre-training};

%arrows
\draw[->] (input.north east) -- (model_resnet.west);
\draw[->] (input.south east) -- (model_densenet.west);

% arrows Resnet-152
\draw[->] (model_resnet.east) -- (model_out_1.west);
\draw[->] (model_densenet.east) -- (model_out_2.west);
\draw[->] (model_out_1.east) -- (fc1.north west);
%\draw[->] (output_1.east) -- (output_2.west);
\draw[->] (model_out_2.east) -- (fc1.south west);
\draw[->] (e1) -- node[anchor=south] {$loss_1$} (e2);
\draw[->] (e3) -- node[anchor=south] {$loss_N$} (e4);
\draw[->] (fc1) -- (fc2);
\draw[->] (fc2) -- (output_2);

%Draw boarders
\draw[black, very thick,densely dotted] (1,-4) rectangle (12,4);
\draw[black, very thick, dotted] (1.2,-3.5) rectangle (6.7,3.5);
\draw[red, very thick, dotted] (model_out_1.south) --  (model_out_2.north);
\draw[blue, very thick, dotted] (model_resnet.south) --  (model_densenet.north);

\end{tikzpicture}
}
\caption{Block diagram of the proposed parallel DNN merging. The training process is split into a pre-training (pre-training of individual models) and post-training step (training the whole network architecture).}
\label{fig:extendible_idea}
\vspace{-10pt}
\end{figure}
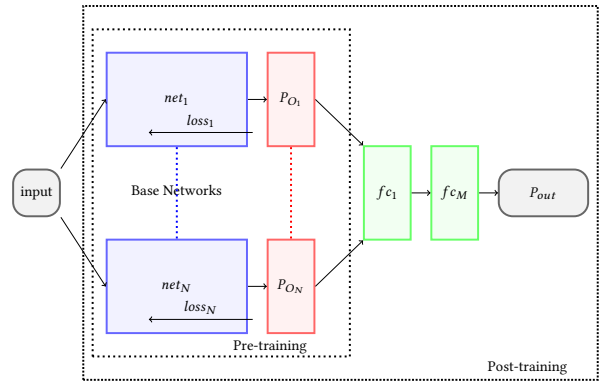

In this section, we show how we can improve accuracy using multiple cumulative probabilistic decisions by extending method 5 into $N\geq2$ \acp{DNN}. In this general model, as illustrated in Figure ~\ref{fig:extendible_idea}, we divide the whole training process into the following four steps: 1) pre-training of individual models, 2) model selection for merging, 3) merging models with a \ac{MLP} and 4) post-training and fine-tuning. Let  $NETS = \{net_1, net_2, \ldots, net_N\}$  be the set of pre-trainer networks using the ImageNet dataset, and 
 $P_{O_i}$ be the returned probability vector for model $net_i$. 

In the pre-training step (1), we train each \ac{DNN}  $net_i \in NETS$ as much as possible using the transfer learning mechanism until it gives the best predictions as described in method 3 (using different loss functions; $loss_1$ to $loss_N$). The \acp{DNN} have their unique prediction capabilities within the given classification problem. Then, we analyze the \ac{CM} of the best outcome of each \ac{DNN}.

At the selection step (2), we select networks that give different diagonals of \ac{CMs} (diagonal of a \ac{CM} represents correct classifications) compared to other \ac{CMs} of selected \acp{DNN}. If the diagonal of \ac{CM} of network $net_i = CM_i$, then we select networks which are having $CM_i \neq CM_j; j=[1,2,...,i-1, i+1, ..., N]$. The goal of this comparison is to identify \ac{DNN} models which have different classification performances compared to each other. 
Equal diagonals of \ac{CMs} do not imply that the networks are identical for their classifications, because there might be models that give the same diagonal numbers but lead to different classifications for a given image. 
If the case of equal diagonals occurs, we have to compare correctly classified images to identify the differences. The number of \acp{DNN} selected for the final training may or may not be equal to the initial number of pre-trained \acp{DNN} depending on similarities in some the \ac{CMs}. 

In the merging step (3), we use an \ac{MLP} to merge all the outputs of the selected \acp{DNN}. The \ac{MLP} consists of $M$ layers which take $ \sum_{i=1}^{N} length\_of(P_{O_i})$ number of inputs and output $P_{out}$ probability vector according to the given classification problem. Then, step (4) can be started by freezing all the pre-trained \acp{DNN} and training only the new \ac{MLP} until it shows a good validation performance. Optionally, we can re-train the whole model without freezing any layer if we cannot achieve a performance improvement by training only the new \ac{MLP}. 

%-------------------

%-------------------------------

\subsubsection{Method 5 used by this research work}
\label{subsec:method5_more}

According to the procedure discussed in Section~\ref{subsec:extendible_idea}, our implementations of method 5 was designed using two parallel networks ($N=2$): ResNet-152 and Densenet-161. Then, we analyzed two \acp{CM}, which came from ResNet-152 and Densenet-161. These two networks were pre-trained according to the given classification problem. Because the $CM_{Resnet\_152} \neq CM_{Densenet\_161}$, then we combined the two networks with a \ac{MLP}. This comparison of \ac{CMs} has been done visually using colormaps. However, if the visual inspection of \ac{CMs} is hard, mathematical operations can be used. Moreover, if the \ac{CMs} are equal completely, a manual inspection of the classified images is required to identify the differences of model classifications. After combining, we freeze two \acp{DNN} to proceed for the post-training step.  In our experiments, the input layer of the \ac{MLP} consists of 32 input nodes. The output of the \ac{MLP} is a probability vector with 16 values, which is equal to the number of classes of the Medico dataset. We used two fully connected layers, with 32 neurons and 16 neurons. In the post-training step, we started training only the \ac{MLP} with a learning rate of 0.01. To do the post-training, the multiclass cross-entropy loss, and \ac{SGD} were used.

%=============================

\section{Results}
\label{sec:results}
In this section, we discuss the experimental setup, datasets, and results obtained from our experiments. Using these presented results, we emphasize that high scores for performance metrics do not always show the actual performance of \ac{ML} methods. To show this, we present well-performing \ac{ML} models, which achieved good results for their performance values. Using cross-dataset testing, we present a detailed analysis of evaluation metrics to emphasize that they are not always representative to identify the real performance of models.

For all experiments, we used the same hardware platform with an Intel Core i7 $8^{th}$ generation processor with 16\,GB DDR4 RAM, and 8\,GB NVIDIA Geforce 1080 GPU. On the other hand, we practiced two different software frameworks for implementing our methods. To implement the \ac{GF}-based  methods 1 and 2, we have used the WEKA framework~\cite{hall2009weka}. We have used the Pytorch framework for the \ac{DNN}-based methods 3, 4, and 5.

\subsection{Datasets}
\label{subsec:dataset}

For the work performed in this paper, we used the following four datasets: 1) the 2018 Medico dataset~\cite{Pogorelov2018}, 2) CVC-356-plus (modified version of CVC-356~\cite{cvc_612_1,bernal2012towards,bernal2015wm}), 3) CVC-612-plus (modified version of CVC-612~\cite{cvc_612_1,bernal2015wm, bernal2012towards}), and 4) CVC-12k~\cite{CVC_12k_1, cvc-12k_2}. The training and testing datasets of the 2018 Medico Task have been derived from the Kvasir Dataset~\cite{pogorelov2017kvasir} and Nerthus Dataset~\cite{nerthus2017mmsys}. It consists of 16 classes as shown in Table~\ref{tbl:medico_dataset_summary}. These images consist of different anatomical landmarks (z-line, pylorus, cecum), pathological findings (esophagitis, polyps, ulcerative colitis), endoscopic polyp removal cases (dyed and lifted polyp, dyed resection margin) and normal findings (normal colon mucosa, stool) in the \ac{GI} tract. The dataset also contains images with a different degree of Boston Bowel Preparation Scale (BBPS) ranging from 0 to 3. Some of the original images contain the endoscope position marking probe. These are seen as a small green box located in the bottom corners, showing its configuration and location of the image frame. The images used in the study are captured using an electromagnetic imaging system (Scopeguide, Olympus, Europe)~\cite{pogorelov2017kvasir}. On the other hand, in Table~\ref{tab:all_datasets}, we present a summary of the uses of the 2018 Medico dataset and other datasets for polyps and non-polyps classifications. 

The Medico development dataset was used to train our \ac{ML} models in the first stage. However, this dataset consists of a highly imbalanced number of images, as summarized in Table~\ref{tbl:medico_dataset_summary}. Within this, the out-of-patient class had only 4 images to train our models. Therefore, only in the first stage, we used an additional 30 images that were selected randomly from the Internet to fill this class in the training dataset. These were images of flowers, vehicles, and other general stuff in our everyday life and did not have any relationship with this class. The advantage of this technique is discussed in the discussion of Section \ref{sec:discuss}.

%%%%%%%%%%%%%%%%%%%%%%%%%%%%%%%%%%%%%%%%%%%%%%%
\begin{table}
\footnotesize
\caption{Summary of the 2018 Medico dataset. Column one shows the names of the different findings. Column two and three list the number of images in the development and test set.}
\vspace{-5pt}
\label{tbl:medico_dataset_summary}
\begin{tabular}{lcc}
    \toprule
    Type & \# of images in Development Set & \# of images in Test Set \\
    \midrule
    Blurry-nothing & 176 &  39\\
    Colon-clear & 267 &  1070\\
    Dyed-lifted-polyps  & 457 &  590\\
    Dyed-resection-margins & 416 &  583\\
    Esophagitis & 444 &  483\\
    Instruments  & 36 &  165\\
    Normal-cecum & 416 &  604\\
    Normal-pylorus & 439 &  569\\
    Normal-z-line & 437 &  636\\
    Out-of-patient  & 4 &  6\\
    Polyps & 613 &  423\\
    Retroflex-rectum & 237 &  194 \\
    Retroflex-stomach & 398 &  399\\
    Stool-inclusions & 130 &  508\\
    Stool-plenty & 366 &  1920\\
    Ulcerative-colitis & 457 & 551\\
    \bottomrule
\end{tabular}
\end{table}

%%%%%%%%%%%%%%%%%%%%%%%%%%%%%%%%%%%%%%%%%%%%%%%%%%%%%%%%%%%%%%%%%%%%%%%%%%%%%%%%%%%%%%%
\begin{table}
\footnotesize
    \centering
     \caption{Overview of the datasets used for our experiments. In total, we have five different datasets, but the Medico dataset is split into a development part and a test part for the challenge. The training and testing column indicates how the dataset was used in the experiments. Polyps and non-polyps indicate the number of findings. Medico and CVC-356-plus represent a bias towards non-findings. CVC-612-plus is a quite balanced dataset and CVC-12k presents a bias towards findings. Dataset were chosen based on these distributions to represent common cases in medical imaging dataset.}
     \vspace{-5pt}
     \caption*{* -  We have replaced this image set with a new image set (with 1,171 images) extracted from a clear colon video collected from the B\ae rum Hospital, Norway in the second stage of this research to avoid the overlap between the training data and the testing data.}
     \vspace{-5pt}
    \begin{tabular}{lccccc}
    \toprule 
    Dataset&Training&Testing&\# images&\# polyps&\# non-polyps \\
    \midrule
         2018 Medico - Development    & X & - & 5906      & 613   & 5293\\
         2018 Medico - Testing        & - & X & 8740     & 423   &   8317\\
         CVC-356-plus               & X & X & 2285      &  356     &  1929$^*$ \\
         CVC-612-plus               & X & X &    1316   & 612   & 704\\
         CVC-12k                    & - & X &    11954  & 10025 &   1929\\
         \bottomrule
    \end{tabular}
   
    \label{tab:all_datasets}
    \vspace{-10pt}
\end{table}

%%%%%%%%%%%%%%%%%%%%%%%%%%%%%%%%%%%%%%%%%%%%%%%%%%%%%%%%%%%%%%%%%%
When we discuss the \ac{ML} models' generalizability in the second part of the paper, we used the CVC datasets to retrain and test our models. The CVC-356-plus dataset is the modified version of the CVC-356~\cite{cvc_612_1,bernal2012towards,bernal2015wm} dataset which has only polyp images. In that modification, we added 1929 non-polyp images from the CVC-12k \cite{CVC_12k_1, cvc-12k_2} dataset to the CVC-356 dataset and created a new dataset called CVC-356-plus. Similarly,  CVC-612-plus dataset was created by extending the CVC-612 dataset~\cite{cvc_612_1,bernal2015wm, bernal2012towards}. For this CVC-612-plus dataset, we added 704 non-polyp images extracted from new \ac{GI} tract videos collected by the B\ae rum Hospital which is part of the Vestre Viken Hospital Trust in Norway. The content of the CVC-12k dataset underwent a minor reorganization by filtering and grouping polyps and non-polyps images into two separate folders. However, the content and number of images in CVC-12k were not otherwise changed. Therefore, we refer to it by its common name. 

In the second part of our research, we have used the CVC-356-plus and CVC-612-plus datasets for retraining our models to classify polyps and non-polyps. Only in this part of the research, we replaced 1,929 non-polyp images of the CVC-356-plus dataset with newly extracted 1,171 images from a clean and healthy colon video collected from the same hospital. We have done this modification to avoid the overlap between the non-polyp images of the CVC-356-plus training dataset and CVC-12k testing dataset.

\begin{figure}[!t]
    \centering
    \includegraphics[width=1\linewidth]{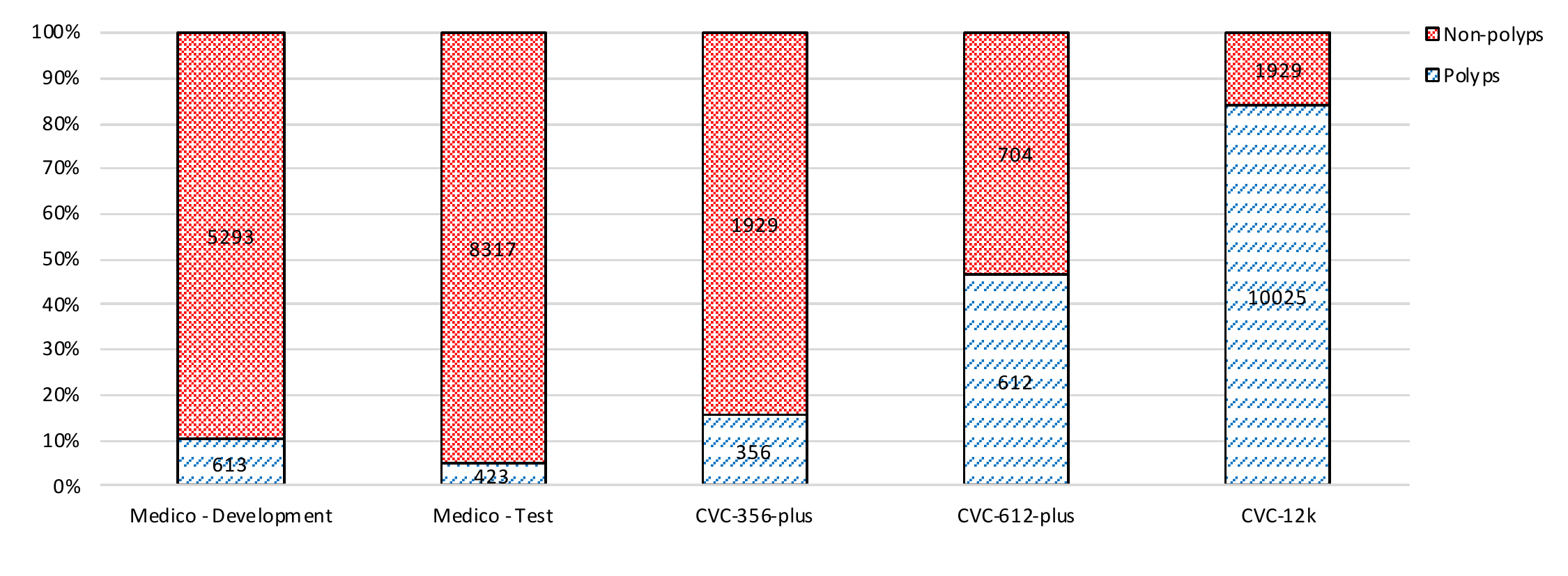}
    \caption{Ratios of findings to non-findings in the datasets (polyp/non-polyp). The X-axis represents the different datasets used for the binary classification. The Y-axis represents the percentage of polyps and non-polyps. The numbers inside the bars show the actual number of polyps and non-polyps images.}
    \label{fig:polyps_nonpolyps_dataset_comparisons}
   \vspace{-10pt}
\end{figure}

For the dataset preparation stage, we focused on the number of polyps and non-polyps images in each dataset to analyze the correlation between the data distribution and the model performance. A bar graph of this data distribution is illustrated in Figure~\ref{fig:polyps_nonpolyps_dataset_comparisons}.
We chose to include different proportions for the number of polyps and non-polyps to keep a diversity of data percentages in each test case. 
In the CVC-356-plus dataset, the polyps percentage is low compared to the non-polyp percentage. 
In the CVC-612-plus dataset, percentages of polyps and non-polyps are around 50\%. In contrast to this, the CVC-12k dataset has a higher polyps percentage than the non-polyps percentage. Due to this, we can study  the effects of data imbalance in the training and testing datasets on the performance and interpretability of the metrics.

%%%%%%%%%%%%%%%%%%%%%%%%%%%%%%%%%%%%%%%%%%%%%%%%%%%%%%%%%%%%%%%%%%%%%%%%%%%
%%%%%%%%%%%%%%%%%%%%%%%%%%%%%%%%%%%%%%%%%%%%%%%%%%%%%%%%%%%%%%%%%%%%%%%%%%%
% add a plot about portions of polyps and non polyps

\subsection{Analyzing results}
% added by Vajria
We discuss our results in two main sections: 1) the 16 class classification task based on the 2018 Medico Task and 2) the polyps and non-polyps classification task to analyze generalizability of \ac{ML} models.

\subsubsection{16-class classification}
In this 16-class classification task, the training dataset of the 2018 Medico Task was split into a $70\%$ training dataset and a $30\%$ validation dataset. Then, the test data given by the organizers was used to test the performance of five methods for classifying 16 classes of the \ac{GI} tract findings.

% Test Results
\begin{table}[!t]
\footnotesize
%[htbp]
%\vspace{-12pt}
%\tiny 
  \caption{Evaluation results of the 2018 Medico Task (as provided by the organizers of the 2018 Medico Task)~\cite{thambawita2018medico} for the five methods used in this article. Based on the official results, method 5 was the best one based on the MCC score. }
  %\vspace{-10pt}
 % \caption*{M = Method, Rec = Recall, SPEC = Specificity, ACC = Accuracy, MCC = Matthews Correlation Coefficient, F1 = F1 Score }
%\vspace{-10pt}
  \label{tab:org_resutls}
  \begin{tabular}{cccccccc}
    \toprule
    Method&REC&PREC&SPEC&ACC&MCC&F1\\
    \midrule
    1 & 0.8457 & 0.8457 & 0.9897 & 0.9807 & 0.8353 & 0.8456 \\
    2 & 0.8457 & 0.8457 & 0.9897 & 0.9807 & 0.8350 & 0.8457 \\
    3 & 0.9376 & 0.9376 & 0.9958 & 0.9922 & 0.9335 & 0.9376 \\
    4 & 0.9400 & 0.9400 & 0.9960 & 0.9925 & 0.9360 & 0.9400 \\
    5 & 0.9458 & 0.9458 & 0.9964 & 0.9932 & 0.9421 & 0.9458 \\
  \bottomrule
\end{tabular}
%\vspace{-10pt}
\end{table}

% Performance plot - 16 class identifications - all five methods
%%%%%%%%%%%%%%%%%%%%%%%%%%%%%%%%%%%%%%%%%%%%%%%%%%%%%%%%%%%%%%%%%%%%%%%%%%%%%%%%%%%%%%%%%%%%%%%%%%%%%%%%%%%%%%%%%%%%
\begin{figure}
    \centering
    \includegraphics[width=.48\linewidth, trim=2.5cm 0cm 2.5cm 0cm, clip]{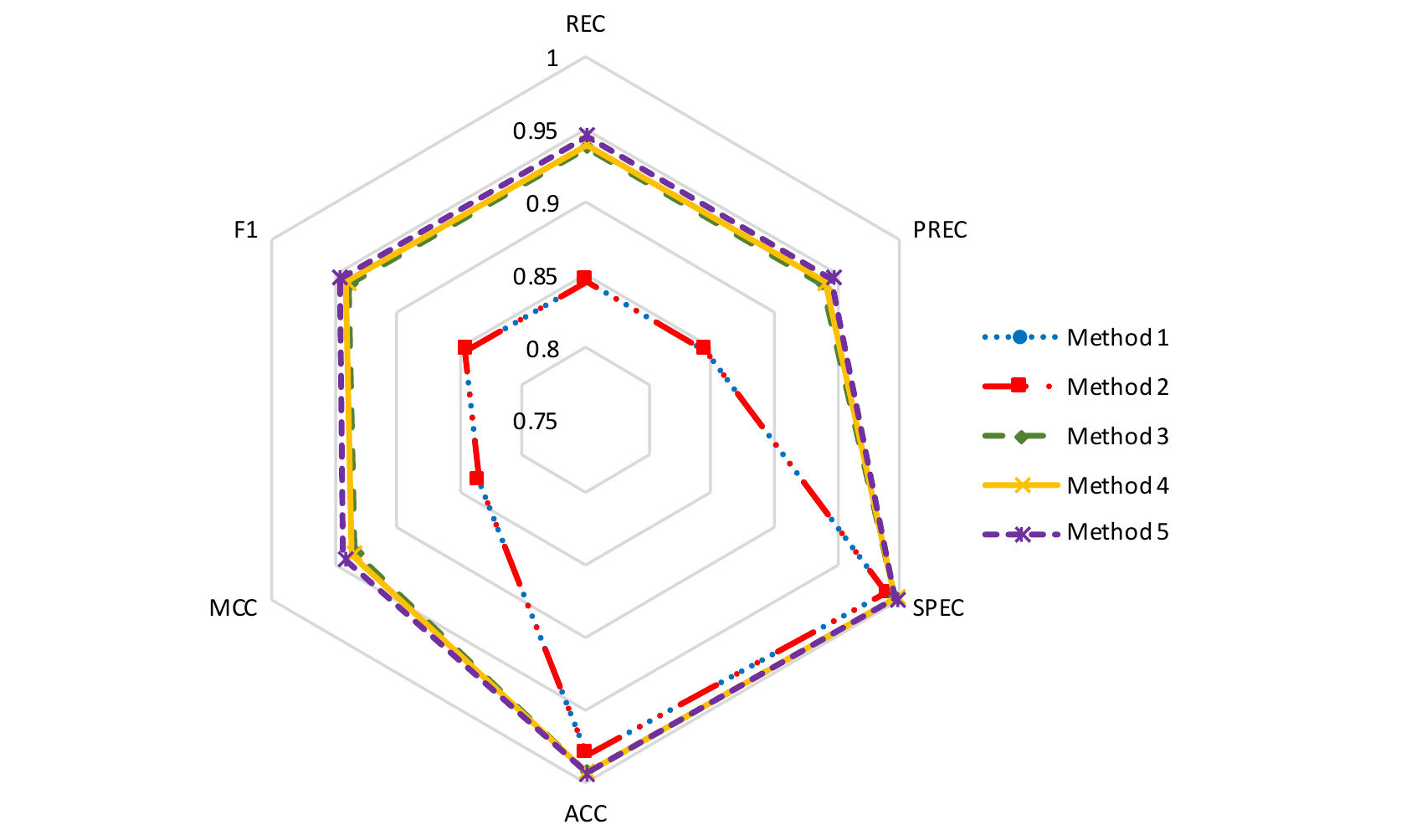}
    \caption{Performance comparison of all five classification models for the 16 classes of the 2018 Medico test dataset. Method 1 and 2 are similar in results but different from the other three methods (note that measurements start from $0.75$).}
    \label{plot:performace_comparison_16_class}
    \vspace{-10pt}
\end{figure}

%%%%%%%%%%%%%%%%%%%%%%%%%%%%%%%%%%%%%%%%%%%%%%%
\begin{table}
\footnotesize
\centering
%\scriptsize
  \caption{Confusion matrix of method 5 (our best model) based on the Medico test dataset. The diagonal value represents true predictions (number of images) of the model. A to P are the classes corresponding to the class names in the first column. Most confusion can be observed between class B and C, and class D and E. Looking at the images shows that they are quite similar in their visual features (colors, texture, etc.)}
  
  % \PH{There is something wrong with thes sizes of the cells, right margin...}}  
  \vspace{-10pt}
 % \caption*{\footnotesize \textmd{\textbf{A}:blurry-nothing, \textbf{B}:colon-clear, \textbf{C}:dyed-lifted-polyps, \textbf{D}:dyed-resection-margins, 
%  \textbf{E}:esophagitis,\textbf{F}:instruments, \textbf{G}:normal-cecum, \textbf{H}:normal-pylorus, 
%  \textbf{I}:normal-z-line, \textbf{J}:out-of-patient, \textbf{K}:polyps, \textbf{L}:retroflex-rectum, 
%  \textbf{M}:retroflex-stomach, \textbf{N}:stool-inclusions, \textbf{O}:stool-plenty, \textbf{P}:ulcerative-colitis}}
  %\vspace{-12pt}
  \label{tab:cm}
  \begin{tabular}{P{0.5mm}lP{2mm}P{2mm}P{2mm}P{2mm}P{2mm}P{2mm}P{2mm}P{2mm}P{2mm}P{2mm}P{2mm}P{2mm}P{2mm}P{2mm}P{2mm}P{2mm}}
                    \toprule
                    & & \multicolumn{16}{c}{Actual class}   \\
                 
                    &   & \textbf{A}  & \textbf{B}  & \textbf{C}   & \textbf{D}   & \textbf{E}   & \textbf{F}  & \textbf{G}   & \textbf{H}   & \textbf{I}   & \textbf{J} & \textbf{K}   & \textbf{L}  & \textbf{M}   & \textbf{N}  & \textbf{O}   & \textbf{P}   \\
                    \midrule 
\parbox[t]{1mm}{\multirow{18}{*}{\rotatebox[origin=c]{90}{Predicted class}}} & \textbf{Ulcerative-colitis (A)} & 500 & - & - & - & - & - & - & - & 39 & - & 3 & - & 1 & 1 & - & 7  \\
                    & \textbf{Esophagitis (B)} & 3 & 432 & 48 & - & - & - & - & - & - & - & - & - & - & - & - & -   \\
                    & \textbf{Normal-z-line (C)} &  1 & 121 & 513 & - & - & - & - & - & - & - & - & - & 1 & - & - & - \\
                    & \textbf{Dyed-lifted-polyps (D)} & 1 & - & - & 522 & 31 & - & - & - & - & - & 2 & - & - & - & - & 34   \\
                    & \textbf{Dyed-resection-margins (E)} & - & - & - & 33 & 532 & - & - & - & - & - & 1 & - & - & - & - & 17   \\
                    & \textbf{Out-of-patient (F)} & - & - & - & - & 1 & 5 & - & - & - & - & - & - & - & - & - & -   \\
                    & \textbf{Normal-pylorus (G)} & 3 & 3 & 2 & - & - & - & 559 & - & - & - & 2 & - & - & - & - & -  \\
                    & \textbf{Stool-inclusions (H)} & - & - & - & - & - & - & - & 501 & 7 & - & - & - & - & - & - & -  \\
                    & \textbf{Stool-plenty (I)} & 1 & - & - & - & - & - & - & - & 1918 & - & - & - & - & - & - & 1  \\
                    & \textbf{Blurry-nothing (J)} & 1 & - & - & - & - & - & - & - & 1 & 37 & - & - & - & - & - & -   \\
                    & \textbf{Polyps (K)} & 10 & - & - & 1 & - & - & 1 & - & - & - & 358 & 6 & - & 1 & - & 46  \\
                    & \textbf{Normal-cecum (L)} & 18 & - & - & - & - & - & - & - & - & - & 6 & 578 & - & - & - & 2   \\
                    & \textbf{Colon-clear (M)} & 1 & - & - & - & - & - & - & 5 & - & - & - & - & 1063 & - & 1 & -   \\
                    & \textbf{Retroflex-rectum (N)} & 3 & - & - & - & - & - & - & - & - & - & 2 & - & - & 188 & 1 & -   \\
                    & \textbf{Retroflex-stomach (O)} & - & - & - & - & - & - & 1 & - & - & - & - & - & - & 2 & 395 & 1   \\
                    & \textbf{Instruments (P)} & - & - & - & - & - & - & - & - & - & - & - & - & - & - & - & 165 \\
                \bottomrule
\end{tabular}
\end{table}

%%%%%%%%%%%%%%%%%%%%%%%%%%%%%%%%%%%%%%%%%%%%%%%%%%%
% Discussion about medico task results
%%%%%%%%%%%%%%%%%%%%%%%%%%%%%%%%%%%%%%%%%%%%%%%%%%%%
% own subsection???? - No

We evaluated our five models based on the results collected by the organizers. This evaluated results of the main five models are tabulated in Table~~\ref{tab:org_resutls}. With an \ac{MCC} score of 0.9421, method~5 showed the best performance for classifying the 16 classes of \ac{GI} tract findings. However, our \ac{GF}-based approaches did not show results competitive with the \ac{DNN} methods. The \ac{GF} model introduced in method~1 could reach an \ac{MCC} score of 0.8353. This result showed the best performance record for a \ac{GF}-based method. A clear performance difference between \ac{GF}-based methods and the \ac{DNN}-based methods can be seen as depicted in Figure~\ref{plot:performace_comparison_16_class}. In this plot, we compared this performance difference using six performance measures: \ac{REC}, \ac{PREC}, \ac{SPEC}, \ac{ACC}, \ac{MCC} and \ac{F1}. According to this plot, it is clear that the areas of the hexagons covered by the \ac{GF} methods are smaller than the areas covered by \ac{DNN} methods. With these results, it implies that three \ac{DL} methods outperform two \ac{GF} methods.

The \ac{CM} of method~5 collected from the organizers of the 2018 Medico Task is tabulated in Table~\ref{tab:cm} for the in-depth investigation.  According to the \ac{CM}, we can identify two main bottlenecks to improve the performance of method~5. The first one is misclassification between esophagitis and normal-z-line, and the second one is misclassification between dyed-lifted-polyps and dyed-resection-margins. Therefore, images from these classes were manually examined to identify the reasons for these misclassifications. For the conflict between esophagitis and normal-z-line,  the reason is very close locations of these two landmarks in the \ac{GI} tract. On the other hand, the confusion between dyed-lifted-polyps vs. dyed-restrictions caused because of the same color patterns and the same texture structures of both types of images. With these limitations, method 5 showed the best performance with an \ac{MCC} of $0.9421$, which was the important measurement to win the 2018 Medico Task. Based on the \ac{MCC} value, we won second place in the 2018 Medico Task. The winning team~\cite{RCNNTrungHieuHoang} relabeled the development dataset and also generated more images out of the provided instruments class by placing the instrument as a foreground over the images of dyed-lifted-polyps, dyed-resection-margins, and ulcerative colitis to balance the instrument class for improving the performance. However, we developed the model by only using the images provided by the task organizers for a fair comparison of the approaches with the limited dataset.  Then, our next experiments were conducted to find the re-usability of these well-performed models in different datasets with polyps and non-polyps categories (subcategories of the 16 classes of primary tasks).

%%%%%%%%%%%%%%%%%%%%%%%%%%%%%%%%%%%%%%%%%%%%%%%%%%%%%%%%%%%%%%

\subsubsection{Polyp and non-polyp classification using the pre-trained models}

The following analysis was performed to identify the polyp classification ability of our five models on the same test dataset and different CVC datasets. The 16-class classification results collected from the Medico Task organizers were analyzed to calculate polyp detection performance in the Medico test data. Moreover, our models were tested with CVC-356-plus, CVC-612-plus, and CVC-12k datasets without any modifications to the five models to compare the performance of polyp detection. 

\begin{figure}
    \centering
    \includegraphics[width=1\linewidth]{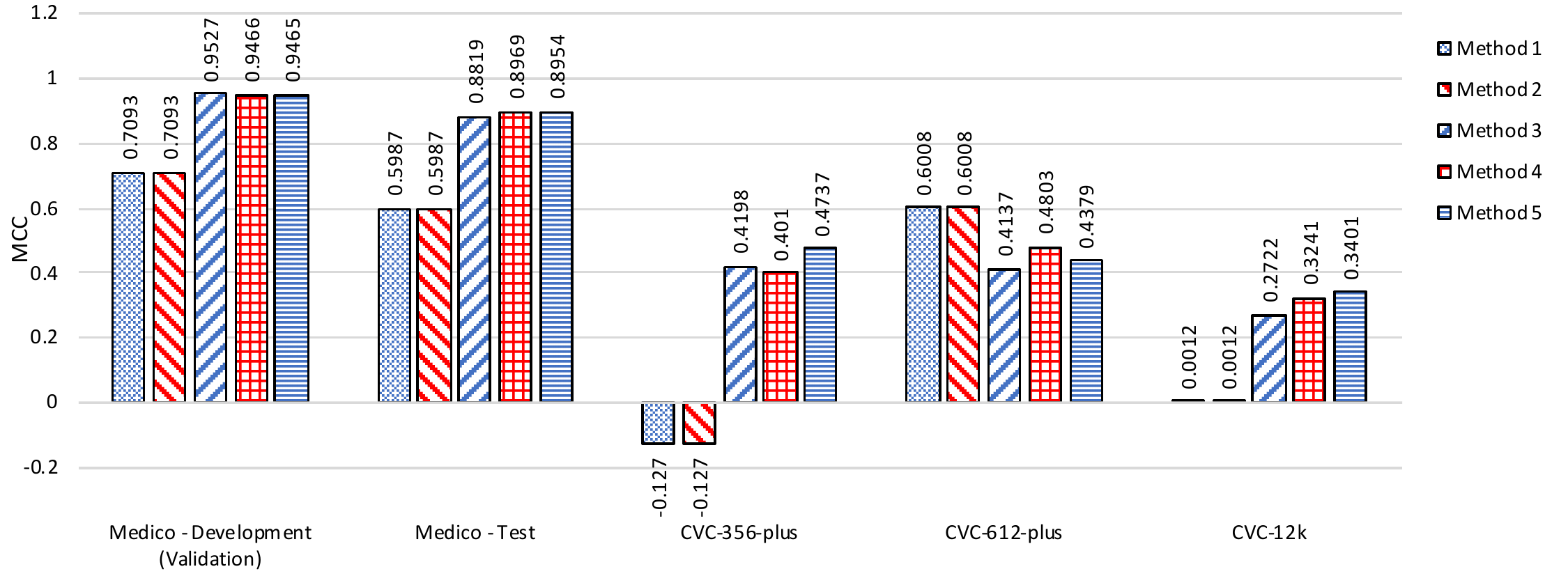}
    \caption{Polyps and non-polyps classification capabilities (based on MCC) of all five methods which were trained using 2018 Medico - development data to classify 16 classed. For most cases method 3 to 5 performs best. For CVC-612-plus test data, methods 1 and 2 are best performing.}
    \label{plot:before_retaining_polyps_detections}
    \vspace{-13pt}
\end{figure}

According to the correct and incorrect classifications of polyps and non-polyps in the test datasets, the first large column of Table~\ref{tab:cvc_resutls} was calculated to measure the polyps detection performance of five models. In this evaluation process, all the 15 classes except the polyp class were considered as the non-polyp classification because the number of outputs is 16 in the first models. For comparison, the MCC values of these tests are plotted in Figure \ref{plot:before_retaining_polyps_detections}. This graph shows that the polyp detection performance of the same dataset (the testing dataset of Medico Task) is higher than on the completely new datasets (CVC-356-plus, CVC-612-plus, and CVC-12k) for both the \ac{GF}-based approaches and the \ac{DNN} approaches. This is the first analysis, and we emphasize that it shows that researchers need to do cross-dataset evaluations to prove the real capabilities of \ac{ML} models.

%% First reference of table 6 is in the above para

%%%%%%%%%%%%%%%%%%%%%%%%%%%%%%%%%%%%%%%%%%%%%%%%%%%%%%%%%%%%%%%%%%%%%%%%%

% Validation accuracy for polyp detection after re-training
\begin{table} %[htbp]
\footnotesize
%\vspace{-12pt}
%\tiny 
  \caption{Polyp classification results with and without retraining for all datasets and methods (column two M=Method). For training, 2018 Medico - development data was used. We can observe that for some datasets, retraining seems to improve the performance. }
%\vspace{-10pt}
  \label{tab:cvc_resutls}
    \begin{tabular}{P{1mm}P{1mm}P{6mm}P{6mm}P{6mm}P{6mm}P{6mm}P{7mm}|P{6mm}P{6mm}P{6mm}P{6mm}P{6mm}P{6mm}}
  %\begin{tabular}{llllllll|llllll}
    \toprule
     & & \multicolumn{6}{c}{Without retraining} & \multicolumn{6}{c}{With retraining to 2 class classification}   \\
    &M&REC&PREC&SPEC&ACC&MCC&F1&REC&PREC&SPEC&ACC&MCC&F1\\
    \midrule
    \multirow{5}{*}{\rotatebox[origin=c]{90}{Test Dataset}} 
    & 1 & 0.7834 & 0.4899 & 0.9635 & 0.9558 & 0.5987 & 0.6028 & 0.9550  & 0.9630 & 0.6740 & 0.9553  & 0.5430  & 0.9590 \\
    & 2 & 0.7834 & 0.4899 & 0.9635 & 0.9558 & 0.5987 & 0.6028 &  0.9540  & 0.9630   & 0.6840 & 0.9537   &  0.5400   &  0.9580 \\
    & 3 & 0.9733 & 0.8088 & 0.9897 & 0.9890 & 0.8819 & 0.8835 & 0.9813 & 0.6577 & 0.9772 & 0.9773  & 0.7934 & 0.7876\\
    & 4 & 0.9599 & 0.8467 & 0.9922 & 0.9908 & 0.8969 & 0.8997 & 0.9813 & 0.7384 & 0.9845 & 0.9843  & 0.8440 & 0.8427\\
    & 5 & 0.9572 & 0.8463 & 0.9922 & 0.9907 & 0.8954 & 0.8984 & 0.9706 & 0.7516 & 0.9857 & 0.9850  & 0.8470 & 0.8471\\
    
    \midrule
    \multirow{5}{*}{\rotatebox[origin=c]{90}{CVC-356-plus}} 
    & 1 & 0.3089 & 0.1053 & 0.5158 & 0.4835 & $-$0.127 & 0.1571 & 0.8450 & 0.7990 & 0.1700 & 0.8446  & 0.0750 & 0.7780\\
    & 2 & 0.3089 & 0.1053 & 0.5158 & 0.4835 & $-0.127$ & 0.1571 & 0.8510 & 0.8420 & 0.2070 & 0.8512 & 0.1930 & 0.7930\\
    & 3 & 0.7865 & 0.3738 & 0.7569 & 0.7615 & 0.4198 & 0.5068 & 0.8118 & 0.5547 & 0.8797 & 0.8691 & 0.5978 & 0.6591\\
    & 4 & 0.6713 & 0.4003 & 0.8144 & 0.7921 & 0.4010 & 0.5016 & 0.6517 & 0.4150 & 0.8305 & 0.8026 & 0.4068 & 0.5071\\
    & 5 & 0.6685 & 0.4837 & 0.8683 & 0.8372 & 0.4737 & 0.5613 & 0.6713 & 0.6408 & 0.9305 & 0.8902 & 0.5906 & 0.6557\\
    
     \midrule
    \multirow{5}{*}{\rotatebox[origin=c]{90}{CVC-612-plus}} 
    & 1 &  0.7696 & 0.7969 & 0.8295 & 0.8016  & 0.6008  & 0.7830 &  0.6980 &  0.8070 & 0.6530 & 0.6983 & 0.4740 & 0.6590\\
    & 2 & 0.7696 & 0.7969 & 0.8295 & 0.8016 &0.6008 & 0.7830 & 0.7220 & 0.8170 & 0.6800 & 0.7218  & 0.5140 &  0.6910\\
    & 3 & 0.8415 & 0.6242 & 0.5597 & 0.6907 & 0.4137 & 0.7168 & 0.8382 & 0.6136 & 0.5412 & 0.6793 & 0.3932 & 0.7086\\
    & 4 & 0.8627 & 0.6559 & 0.6065 & 0.7257 & 0.4803 & 0.7452 & 0.8578 & 0.6890 & 0.6634 & 0.7538 & 0.5265 & 0.7642\\
    & 5 & 0.8137 & 0.6501 & 0.6193 & 0.7097 & 0.4379 & 0.7228 & 0.8007 & 0.7061 & 0.7102 & 0.7523 & 0.5104 & 0.7504\\
    
     \midrule
    \multirow{5}{*}{\rotatebox[origin=c]{90}{CVC-12k}} 
    & 1 & 0.4858 & 0.8391 & 0.5158 &  0.4907 & 0.0012 & 0.6154 &  0.1650 &  0.7880 & 0.8370 &  0.1651 & 0.0130 &  0.0530\\
    & 2 & 0.4858 & 0.8391 & 0.5158 &  0.4907 & 0.0012 & 0.6154 &  0.1650 & 0.8210 & 0.8380 & 0.1699 & 0.0290 & 0.0630\\
    & 3 & 0.6112 & 0.9289 & 0.7569 & 0.6347 & 0.2722 & 0.7373 & 0.6033 & 0.9631 & 0.8797 & 0.6479 & 0.3558 & 0.7419\\
    & 4 & 0.6236 & 0.9458 & 0.8144 & 0.6544 & 0.3241 & 0.7517 & 0.6459 & 0.9519 & 0.8305 & 0.6757 & 0.3539 & 0.7696\\
    & 5 & 0.5936 & 0.9591 & 0.8683 & 0.6379 & 0.3401 & 0.7333 & 0.5576 & 0.9766 & 0.9305 & 0.6178 & 0.3595 & 0.7099\\
  \bottomrule
\end{tabular}
%\vspace{-13pt}
\end{table}

%%%%%%%%%%%%%%%%%%%%%%%%%%%%%%%%%%%%%%%%%%%%%%%

From the first column of Table \ref{tab:cvc_resutls} and Figure~\ref{plot:before_retaining_polyps_detections}, it is clear that the performance of the \ac{GF} methods for different datasets (CVC-356plus, CVC-612-plus, and CVC-{$(356$+$612$)} dataset) is unpredictable because it presents huge value fluctuations in the graph with negative \ac{MCC} value. This shows the incapability of \ac{GF} methods to make predictions on different datasets.
The negative values of \ac{MCC} in this experiment like $ -0.127 $ for the CVC-356-plus dataset indicate that there is no agreement or only a not relevant relationship between target and prediction. An \ac{MCC} around 0 would mean that the classifier is deciding random and MCCs above 0 would indicate correct classification. The closer to -1 or 1 the stronger is the indication for begin wrong or correct, respectively. However, the polyp detection performance of the \ac{GF}-based methods in the CVC-612-plus dataset outperforms the \ac{DNN} methods with the \ac{MCC} value of 0.6008 while the best \ac{DNN} method shows the MCC value of 0.4803. This prediction accuracy of the \ac{GF} methods can be identified as an erroneous prediction because the performance of this method for the other two CVC datasets shows poor \ac{MCC} scores than the \ac{DNN} based approaches. Moreover, the \ac{DNN} based approaches show considerable steady \ac{MCC} values for all new datasets, and it implies that \ac{DNN} methods are more generalizable than the \ac{GF} methods.

Because the performance gap between the 16-class classification and polyps classification showed differences, we retrained our models to classify only the polyps and non-polyps classes. Therefore, our next experiments were performed to test how retraining our five \ac{ML} models to classify polyps and non-polyps will influence the performance.

%%%%%%%%%%%%%%%%%%%%%%%%%%%%%%%%%%%%%%%%%%%%%%%
% Plot before retaining - polyps detections

%%%%%%%%%%%%%%%%%%%%%%%%%%%%%%%%%%%%%%%%%%%%%%%

For the retraining experiments, we first retrained the two \ac{GF} methods with new ARFF files generated for polyps and non-polyps categories. Second, in the retraining stage of the three \ac{DNN} methods, we changed only the last layer into two outputs. However, we did not change the loss function from categorical cross-entropy into binary cross-entropy because two-class categorical cross-entropy is equal to binary cross-entropy.  Moreover, we retained the original optimization functions as it is. Then, we retrained all five models using the same Medico dataset, which has only polyps and non-polyps classes. At the end, the results of these experiments are tabulated in the right columns of Table \ref{tab:cvc_resutls}.

The results in Table \ref{tab:cvc_resutls} show that it can be difficult to evaluate the models and interpret the results after retraining for two class classification.
All MCC values of the five methods tested on the CVC-356-plus data show improvements. Similarly, for the CVC-612-plus test data, method 4 and 5 show performance improvements from \ac{MCC} values of $0.4803$ and $0.4379$ to $0.5265$ and $0.5104$, respectively. 
In contrast, methods 1, 2 and 3 show a performance drop which is indicated by MCC values 0.6008, 0.6008 and 0.4137 reduced to 0.4740, 0.5140 and, 0.3932, respectively.  
Therefore, we extended our experiment by introducing additional retraining options with the CVC-356-plus and CVC-612-plus datasets. After that, the retraining process can be categorized as retraining the models to classify polyps and non-polyps using 1) only the same Medico training dataset (as tabulated in Table \ref{tab:cvc_resutls}), 2) the Medico dataset with the CVC-356-plus dataset, 3) the Medico dataset with the CVC-612-plus dataset, and 4) the Medico dataset with the CVC-356-plus  and CVC-612-plus datasets. Then, our testing datasets are limited to two datasets: the Medico test dataset and the CVC-12k dataset. Results related to these new retraining processes can be seen in Table~\ref{tab:cvc_retraining}.
When the models are trained using the balanced CVC-612-plus dataset in combination with the 2018 Medico development data, the \ac{DNN} models show better MCC values ($0.8189$, $0.8555$, and $0.8606$) for methods 3, 4, and 5, respectively. This is true for the Medico test data and the two smaller CVC datasets. Moreover, the MCC values for the CVC-12k test data also achieves the best MCC values of $0.1421$, $0.1418$, and $0.1802$ for method 3, 4, and 5. An important observation from the CVC-12k dataset is also that looking at all other metrics but MCC and specificity could mislead to the assumption that the results are good. For example, scores above $0.8$ for accuracy which is often used as the only indicator for performance in similar studies.

% New tables are here
%%%%%%%%%%%%%%%%%%%%%%%%%%%%%%%%%%%%%%%%%%%%%%%
\newcommand{\spheading}[2][10em]{% \spheading[<width>]{<stuff>}
  \rotatebox{90}{\parbox{#1}{\raggedright #2}}}
% Validation accuracy for polyp detection after re-training
\begin{table} %[htbp]
\footnotesize
%\vspace{-12pt}
%\tiny
  \caption{Evaluation results on using CVC-356-plus and CVC-612-plus combined as training data with retraining to classify polyps and non-polyps. The 2018 Medico test dataset and the CVC-12k dataset are the test datasets. Using the balanced CVC-612-plus as training data, we achieve the best results. Combining the CVC-356-plus and the CVC-612-plus does not improve the performance. Overall the performance is better on the Medico test dataset. }
%\vspace{-10pt}
  \label{tab:cvc_retraining}
\begin{tabular}{P{1mm}P{5.5mm}|P{1mm}P{6mm}P{6mm}P{6mm}P{6mm}P{6mm}P{7mm}|P{6mm}P{6mm}P{6mm}P{6mm}P{6mm}P{6mm}}
  %\begin{tabular}{llllllll|llllll}
    \toprule
    & & & \multicolumn{6}{c}{Medico test data} & \multicolumn{6}{c}{CVC-12k}   \\
    & &M&REC&PREC&SPEC&ACC&MCC&F1&REC&PREC&SPEC&ACC&MCC&F1\\
    \midrule
     \multirow{15}{*}{\rotatebox[origin=c]{90}{Retraining datasets with Medico data}} &
    \multirow{5}{*}{\rotatebox[origin=c]{90}{\parbox{1cm}{\centering CVC-356-plus}}} 
     & 1 &   0.9550  & 0.9610 & 0.6230 &  0.9549 &  0.5160  &  0.9570  & 0.5840 & 0.7040 & 0.3090 & 0.5836  & $-0.084$ & 0.6320 \\
   & & 2 &   0.9520   &  0.9620  & 0.6710 &  0.9521  &  0.5260   &   0.9560 & 0.5810 & 0.7100 & 0.3360 & 0.5807  & $-0.065$ & 0.6310 \\
   & & 3 & 0.9626 & 0.6630 & 0.9781 & 0.9775 & 0.7887 & 0.7852 & 0.8423 & 0.8565 & 0.2665 & 0.7494  & 0.1052 & 0.8493\\
   & & 4 & 0.9599 & 0.7526 & 0.9859 & 0.9848 & 0.8427 & 0.8437 & 0.9192 & 0.8481 & 0.1441 & 0.7941  & 0.0810 & 0.8822\\
   & & 5 & 0.9706 & 0.7773 & 0.9876 & 0.9868 & 0.8623 & 0.8633 & 0.8694 & 0.8507 & 0.2068 & 0.7625  & 0.0802 & 0.8599\\

     \cmidrule{2-15}
   &  \multirow{5}{*}{\rotatebox[origin=c]{90}{\parbox{1cm}{\centering CVC-612-plus}}} 
     & 1 & 0.9510  &  0.9590   & 0.6270 & 0.9508 &  0.4970 &   0.9540   & 0.5840 & 0.7030 & 0.3040 & 0.5842  & $-0.087$ & 0.6320 \\
   & & 2 &   0.9530 &  0.9610 & 0.6430 &  0.9530 & 0.5160   & 0.9560 & 0.6400 & 0.6970 & 0.2240 & 0.6395  & $-0.117$ & 0.6660 \\
   & & 3 & 0.9652 & 0.7092 & 0.9823 & 0.9816 & 0.8189 & 0.8177 & 0.9325 & 0.8546 & 0.1752 & 0.8103  & 0.1421 & 0.8918\\
   & & 4 & 0.9572 & 0.7766 & 0.9877 & 0.9864 & 0.8555 & 0.8575 & 0.9336 & 0.8544 & 0.1731 & 0.8109  & 0.1418 & 0.8922\\
   & & 5 & 0.9626 & 0.7809 & 0.9879 & 0.9868 & 0.8606 & 0.8623 & 0.9486 & 0.8571 & 0.1778 & 0.8242  & 0.1802 & 0.9005\\

     \cmidrule{2-15}
   & \multirow{5}{*}{\rotatebox[origin=c]{90}{\parbox{1cm}{\centering CVC-\{356+612\}}}} 
    & 1 & 0.9500  &  0.9600  & 0.6480 & 0.9503 &   0.5050  &  0.9540  & 0.6180 & 0.6930 & 0.2280 & 0.6179  & $-0.129$ & 0.6520 \\
  & & 2 &  0.9500 &  0.9610 & 0.6710 &  0.9503  &  0.5170  &   0.9550  & 0.7200 & 0.7010 & 0.1820 & 0.7199  & $-0.105$ & 0.7100\\
  & & 3 & 0.9733 & 0.5909 & 0.9699 & 0.9700 & 0.7458 & 0.7354 & 0.9537 & 0.8479 & 0.1109 & 0.8177  & 0.1028 & 0.8977\\
  & & 4 & 0.9545 & 0.7596 & 0.9865 & 0.9851 & 0.8443 & 0.8460 & 0.9543 & 0.8463 & 0.0995 & 0.8164  & 0.0874 & 0.8971\\
  & & 5 & 0.9599 & 0.7771 & 0.9877 & 0.9865 & 0.8571 & 0.8589 & 0.9278 & 0.8462 & 0.1239 & 0.7981  & 0.0699 & 0.8851\\
    
  \bottomrule
\end{tabular}
\vspace{-10pt}
\end{table}

%%%%%%%%%%%%%%%%%%%%%%%%%%%%%%%%%%%%%%%%%%%%%%%%%%%%%
% New plot series to replace old plots
%%%%%%%%%%%%%%%%%%%%%%%%%%%%%%%%%%%%%%%%%%%%%%%
\begin{figure}[htbp]
    \centering
    \rotatebox[origin=c]{0}{(a)}\quad
    \subfloat{
        \includegraphics[width=.4\linewidth, trim=0cm 1.3cm 0cm 1.3cm, clip,valign=c]{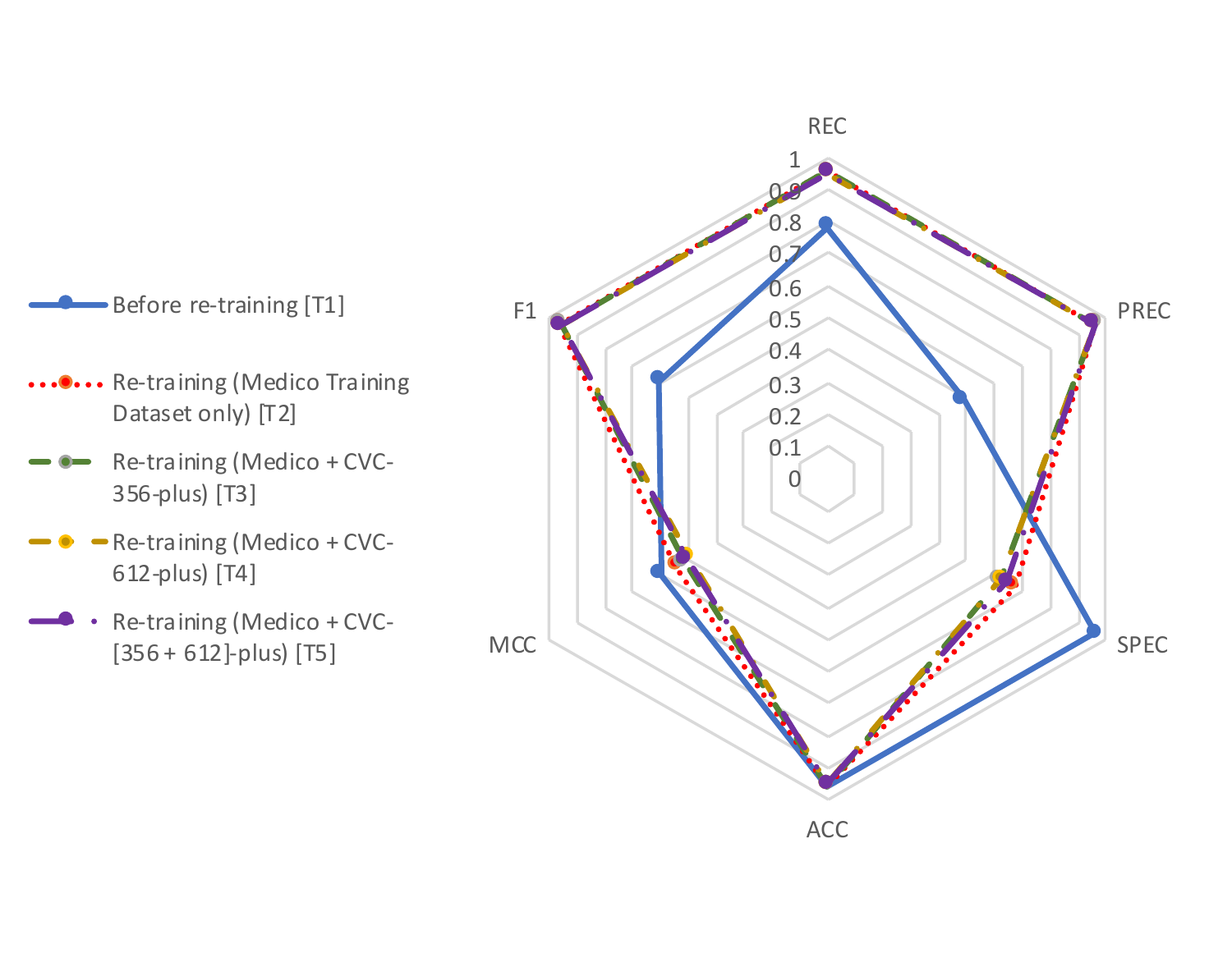}
        \label{plot:a}
    }
   % \hspace{0mm}
   \hfill
   \rotatebox[origin=c]{0}{(b)}\quad
    \subfloat{
        \includegraphics[width=.4\linewidth, trim=0cm 1.3cm 0cm 1.3cm, clip, valign=c]{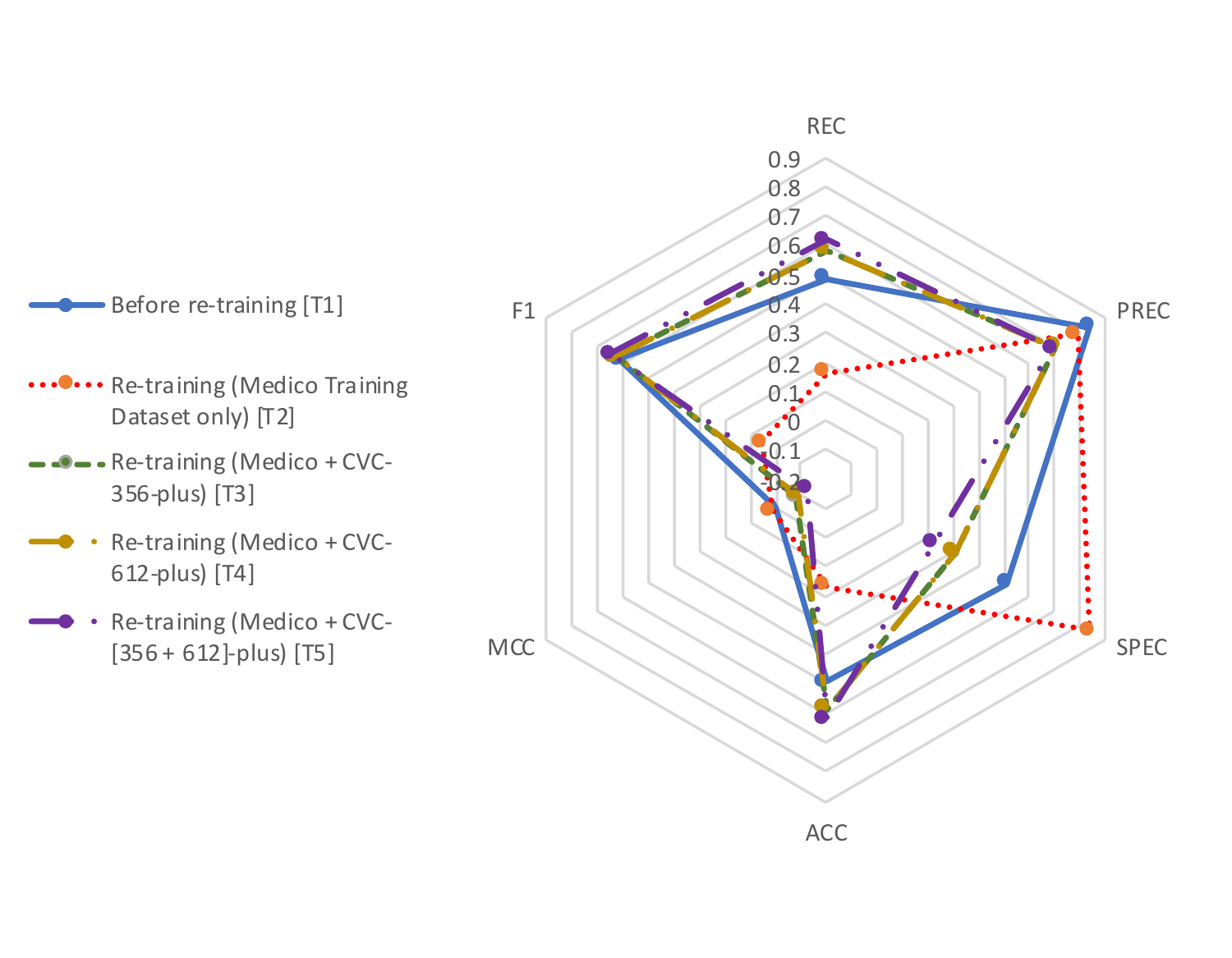}
        \label{plot:b}
    }
    
    \rotatebox[origin=c]{0}{(c)}\quad
    \subfloat{
        \includegraphics[width=.4\linewidth, trim=0cm 1.3cm 0cm 1.3cm, clip, valign=c]{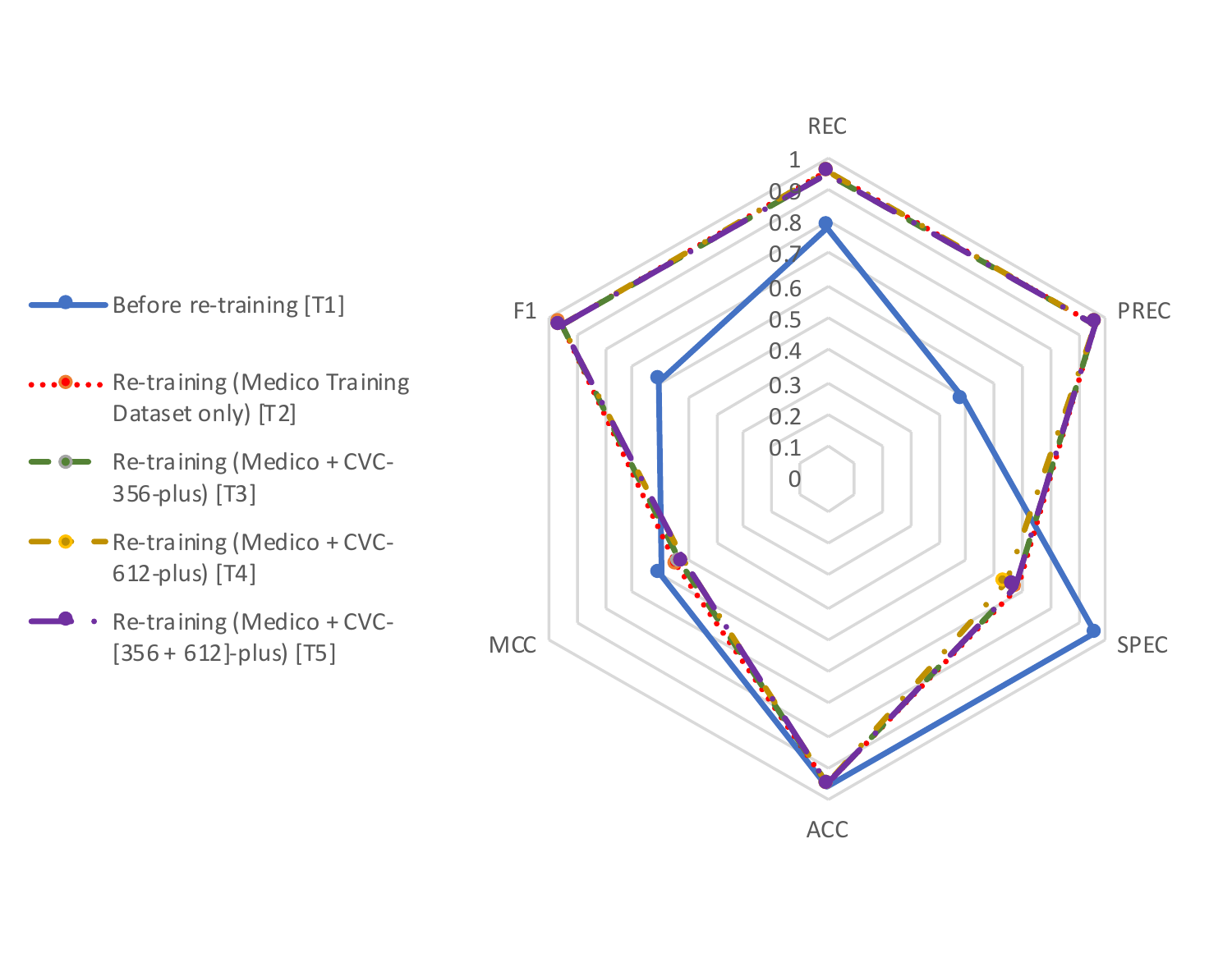}
        \label{plot:c}
    }
   % \hspace{0mm}
   \hfill
    \rotatebox[origin=c]{0}{(d)}\quad
    \subfloat{
        \includegraphics[width=.4\linewidth, trim=0cm 1.3cm 0cm 1.3cm, clip, valign=c]{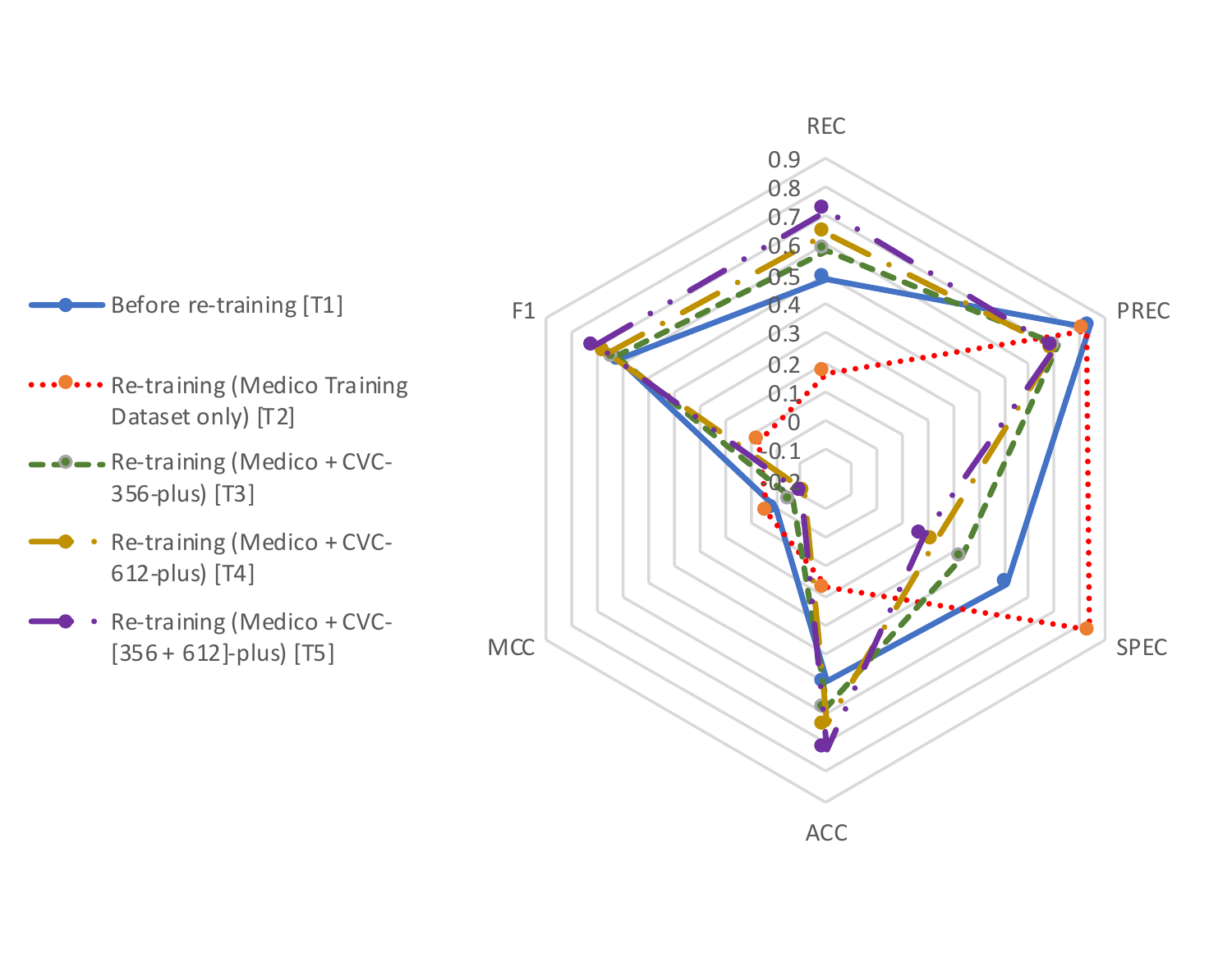}
        \label{plot:d}
    }
    
     \rotatebox[origin=c]{0}{(e)}\quad
    \subfloat{
        \includegraphics[width=.4\linewidth, trim=0cm 1.3cm 0cm 1.3cm, clip, valign=c]{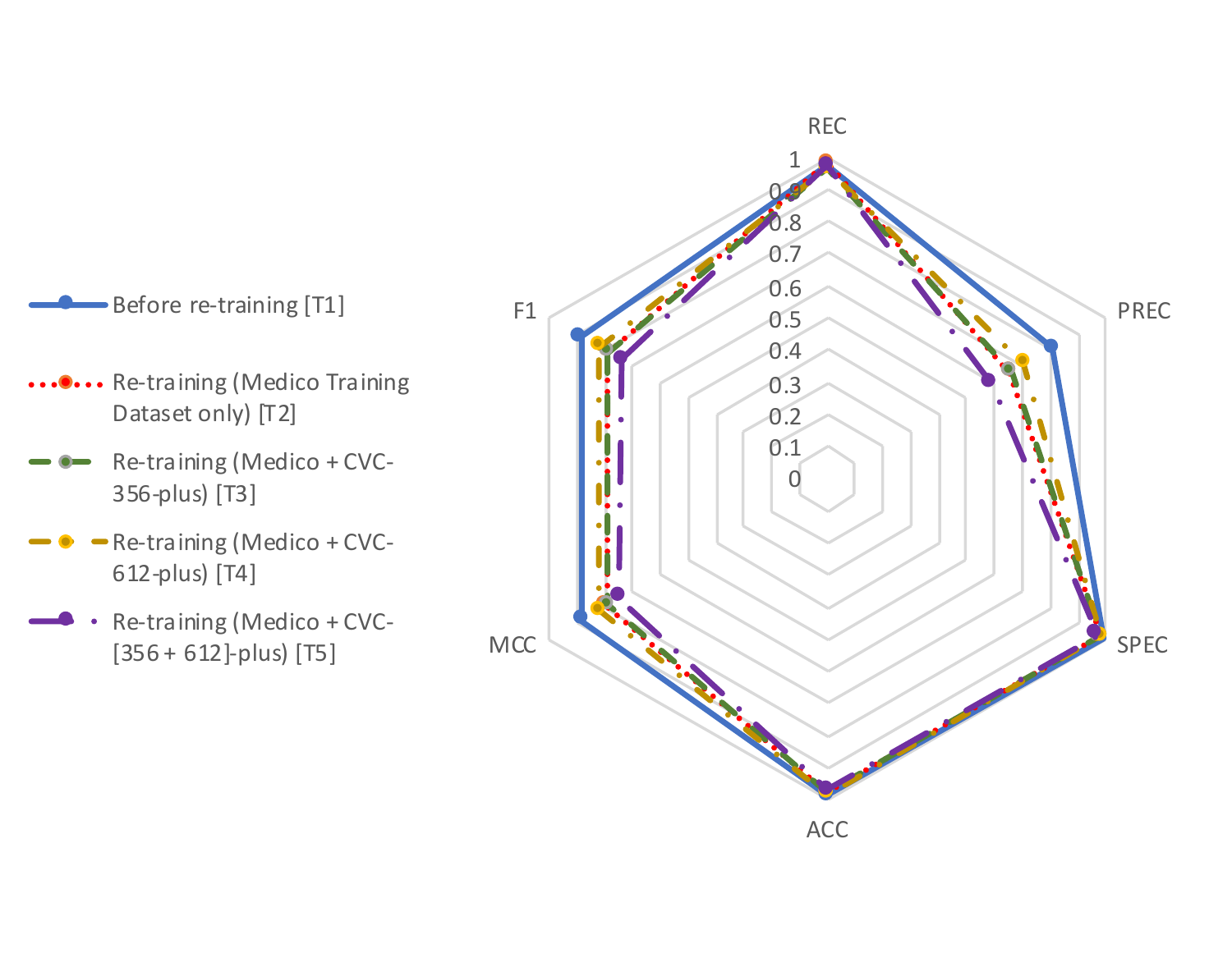}
        \label{plot:e}
    }
   % \hspace{0mm}
   \hfill
    \rotatebox[origin=c]{0}{(f)}\quad
    \subfloat{
        \includegraphics[width=.4\linewidth, trim=0cm 1.3cm 0cm 1.3cm, clip, valign=c]{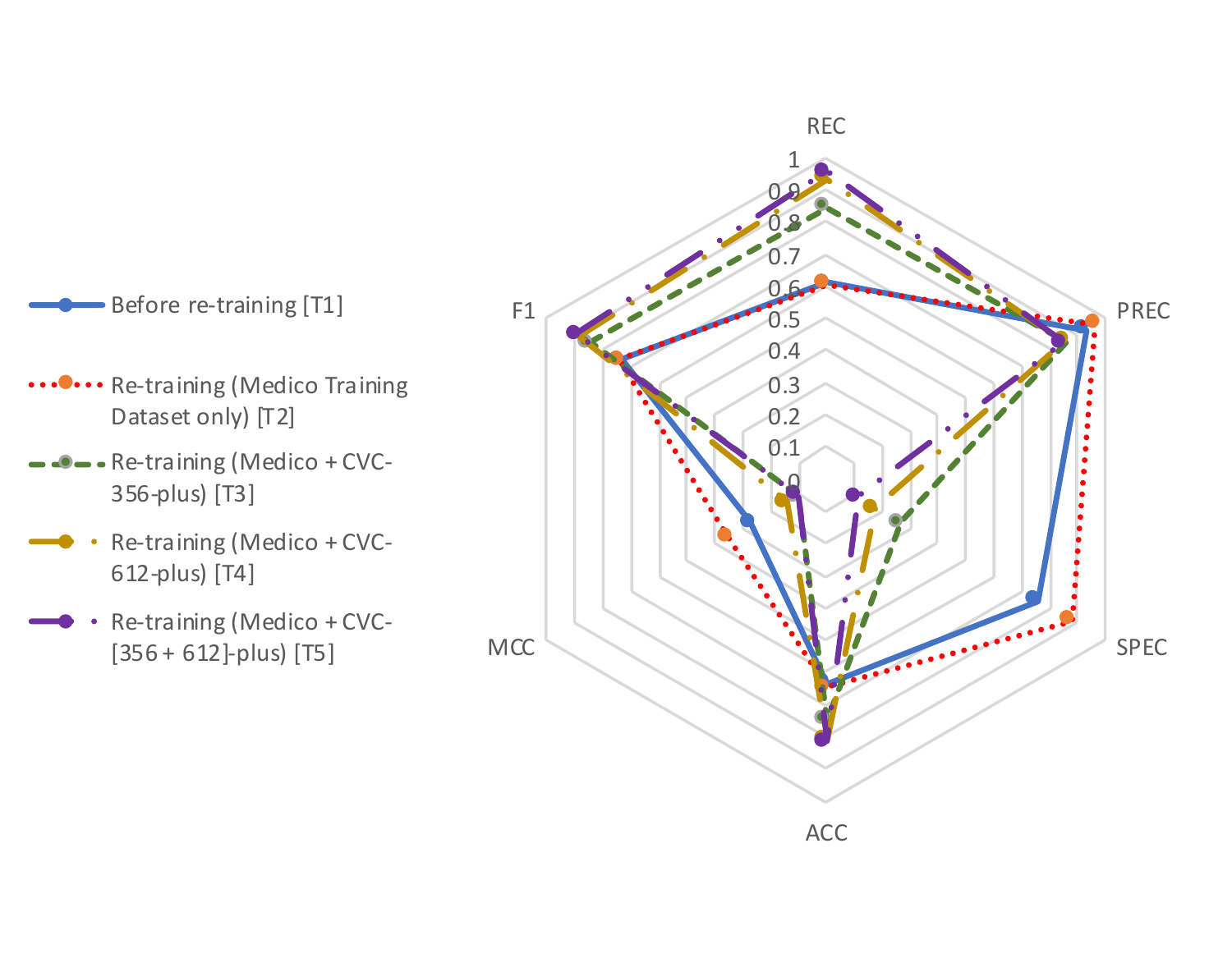}
        \label{plot:f}
    }
    
     \rotatebox[origin=c]{0}{(g)}\quad
    \subfloat{
        \includegraphics[width=.4\linewidth, trim=0cm 1.3cm 0cm 1.3cm, clip, valign=c]{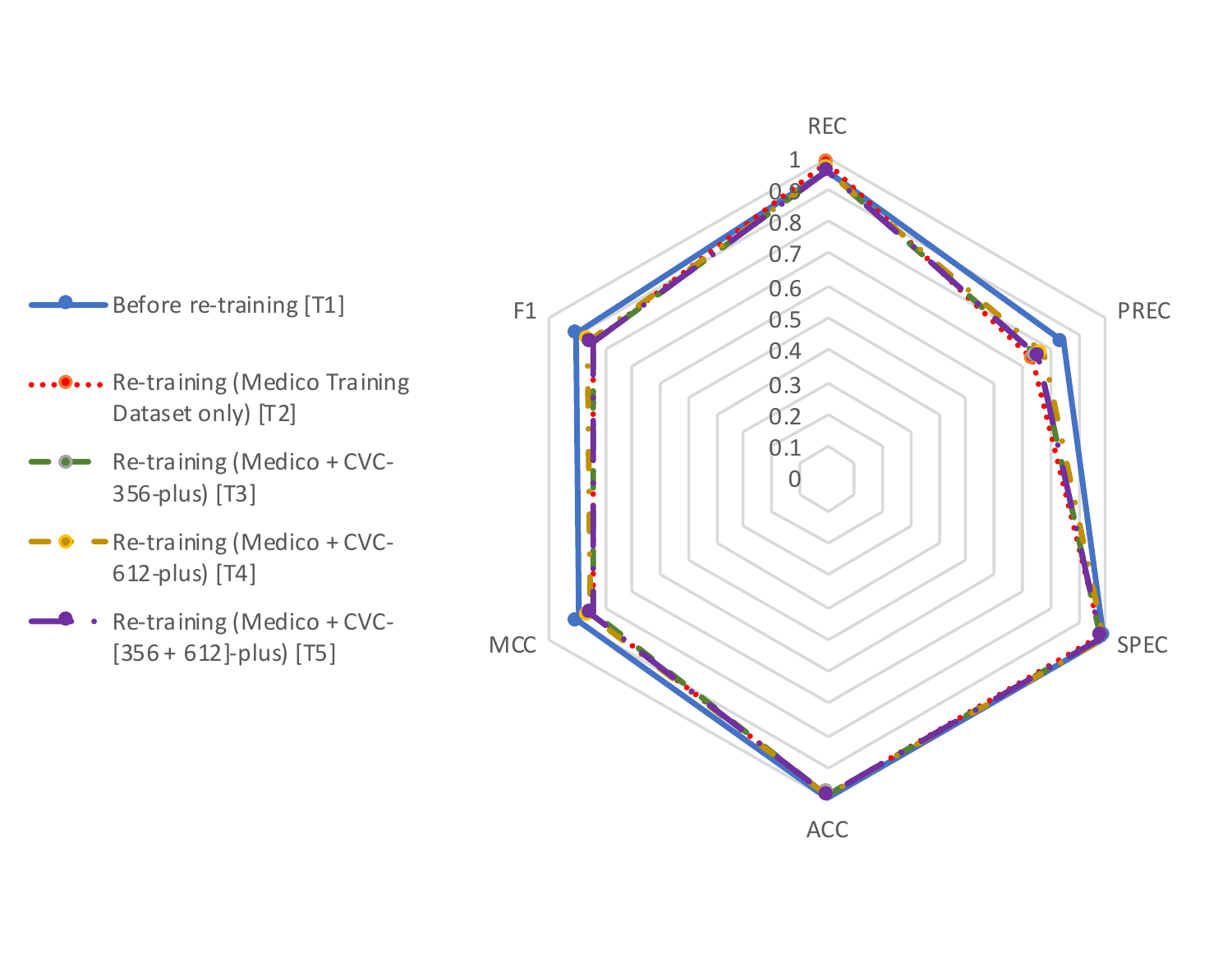}
        \label{plot:g}
    }
   % \hspace{0mm}
   \hfill
    \rotatebox[origin=c]{0}{(h)}\quad
    \subfloat{
        \includegraphics[width=.4\linewidth, trim=0cm 1.3cm 0cm 1.3cm, clip, valign=c]{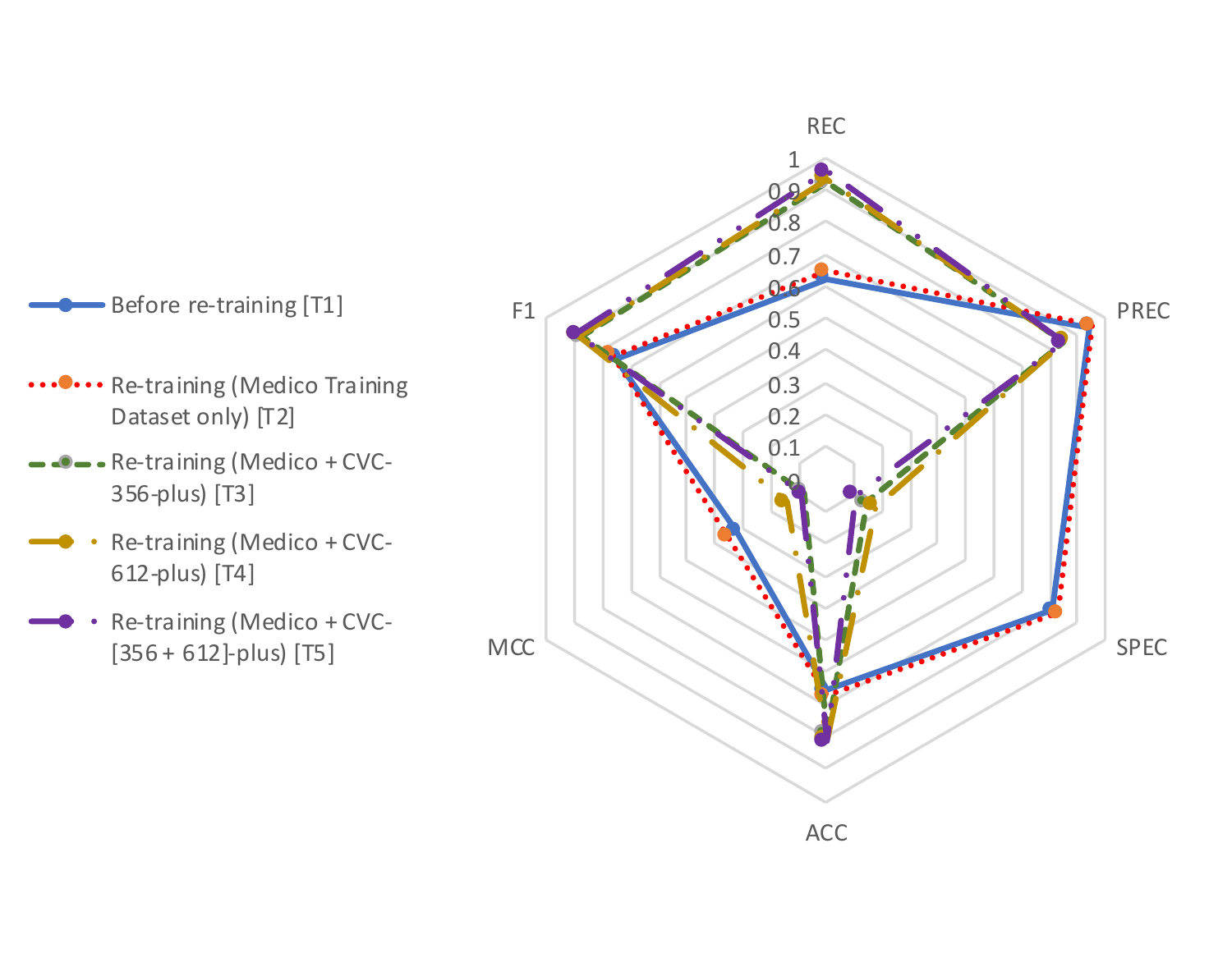}
        \label{plot:h}
    }

     \rotatebox[origin=c]{0}{(i)}\quad
    \subfloat{
        \includegraphics[width=.4\linewidth, trim=0cm 1.3cm 0cm 1.3cm, clip, valign=c]{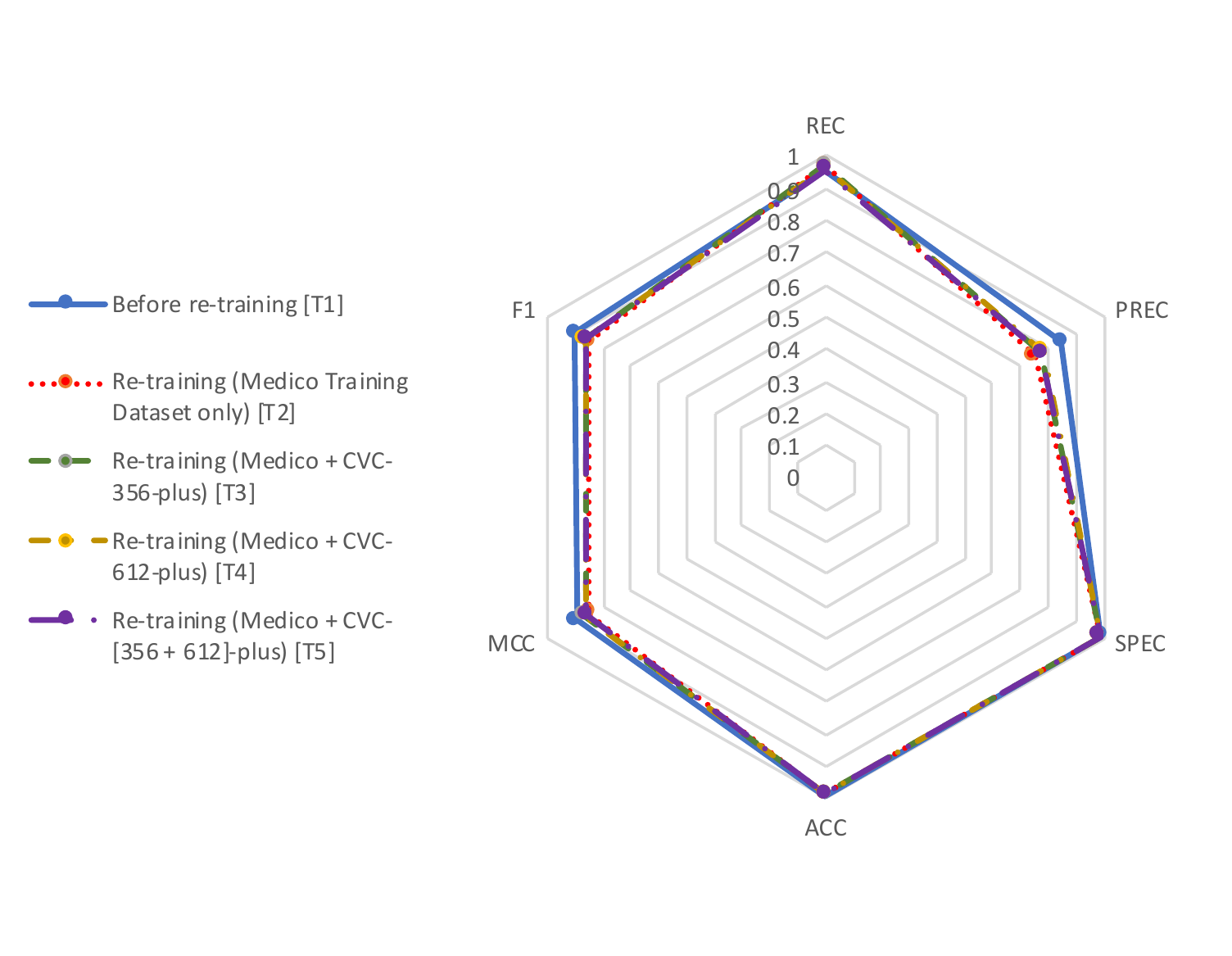}
        \label{plot:i}
    }
   % \hspace{0mm}
   \hfill
    \rotatebox[origin=c]{0}{(j)}\quad
    \subfloat{
        \includegraphics[width=.4\linewidth, trim=0cm 1.3cm 0cm 1.3cm, clip, valign=c]{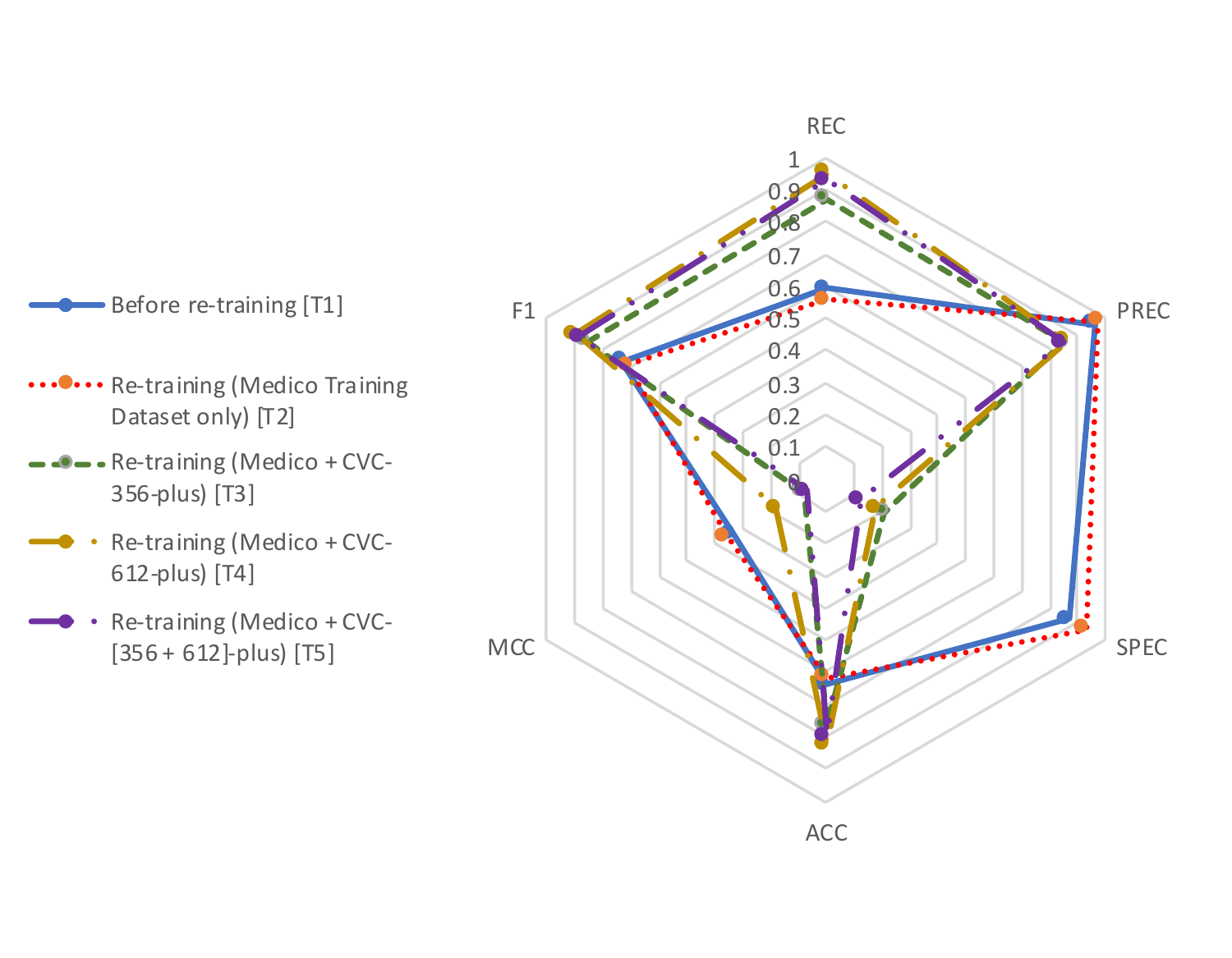}
        \label{plot:j}
    }

    \caption{Polyps and non-polyps classification using the proposed \ac{ML} methods: 1, 2, 3, 4 and 5. The first column (subfigures (a, c, e, g, i)) shows the results of the Medico test dataset and the second column(subfigures {(b, d, f, h, j))} shows the results of the CVC-12k dataset. The methods are represented as following: a-b is for method 1, c-d is for method 2, e-f is for method 3, g-h is for method 4 and i-j is for method 5, respectively.}
    \label{plot:overall_performance_all_methods}
\end{figure}

%The first column shows the results obtained from the Medico Task test dataset and the second column shows the results obtained from the CVC-12k dataset as the test dataset.

%%%%%%%%%%%%%%%%%%%%%%%%%%%%%%%%%%%%%%%%%%%%%%%%%%%%%%%%%%%%%%%%%%%%%

In the first comparison, we plotted performance changes for the retraining with the different training datasets and tested them on the Medico test dataset. The changes in the \ac{REC}, \ac{PREC}, \ac{SPEC}, \ac{ACC}, \ac{MCC} and \ac{F1} values can be seen as hexagon plots in Figures \ref{plot:a}, \ref{plot:c}, \ref{plot:e}, \ref{plot:g}, and \ref{plot:i} which are corresponding to methods 1, 2 ,3, 4 and 5, respectively. In these plots, T1 is used to present performance values before retraining the \ac{ML} models into two-class classification (binary classification). In this case, 15 classes except for the polyp class of the 16 classes were considered as the non-polyp class, and the polyp class is counted as the same polyp class. Furthermore, from T2 to T5 lines are used to present models with only two outputs. The T2 plot represents models' performance for the retraining using the Medico training dataset. Similarly, T3, T4, and T5 represent the retraining process using the Medico dataset and the CVC-356-plus dataset, the Medico dataset and the CVC-612-plus dataset, and the Medico dataset, the CVC-356-plus dataset, and the CVC-612 dataset respectively.

In the second series of experiments in this session, the same experiments were performed and tested on the CVC-12k dataset. The results obtained from these experiments are tabulated in Tables~\ref{tab:cvc_resutls} and \ref{tab:cvc_retraining}. Then, relevant results from these tables are plotted in Figures \ref{plot:b}, \ref{plot:d}, \ref{plot:f}, \ref{plot:h}, and \ref{plot:j}. These plots use same line notations similar to the above experiments.

Using the plot series in Figure \ref{plot:overall_performance_all_methods}, we can examine the re-usability of \ac{ML} models to classify polyps and non-polyps, which are sub-classes of the primary classes on the task. For example, if we compare plots in Figures \ref{plot:a} and \ref{plot:b}, then we can know that how method 1 performs to classify polyps and non-polyps within the test dataset same as the training dataset and within an entirely new dataset. While investigating these plots, the proportion of the number of polyps and non-polyps is an important factor in explaining the shape of these hexagon plots.

If we compare the \ac{GF} methods (Figures \ref{plot:a} $-$ \ref{plot:d} ) and the \ac{DL} methods (Figures \ref{plot:e} $-$ \ref{plot:j}), it is clear that \ac{DL} methods outperforms the \ac{GF} methods in both Medico Task and polyps classification task introduced in this paper. This implies that the \ac{DL} methods are capable of extracting deep features that cannot be extracted by manual feature extraction methods used by the \ac{GF} methods.  With the retraining process in the \ac{GF} methods, we can see performance differences between the Medico dataset and the CVC-12k dataset. The main conclusion that we make is that \ac{GF}-based methods are not able to capture the underlying patterns that would allow for efficient classification; thus their performance is low.

%%%%%%%%%%%%%%%%%%%%%%%%%%
% New ROC and PRC courves
%%%%%%%%%%%%%%%%%%%%%%%%%

\begin{figure}
    \centering
    \subfloat[][\centering ROC curves for the Medico test dataset.]{
        \includegraphics[width=.48\linewidth, trim=0cm 0cm 0cm 0cm, clip]{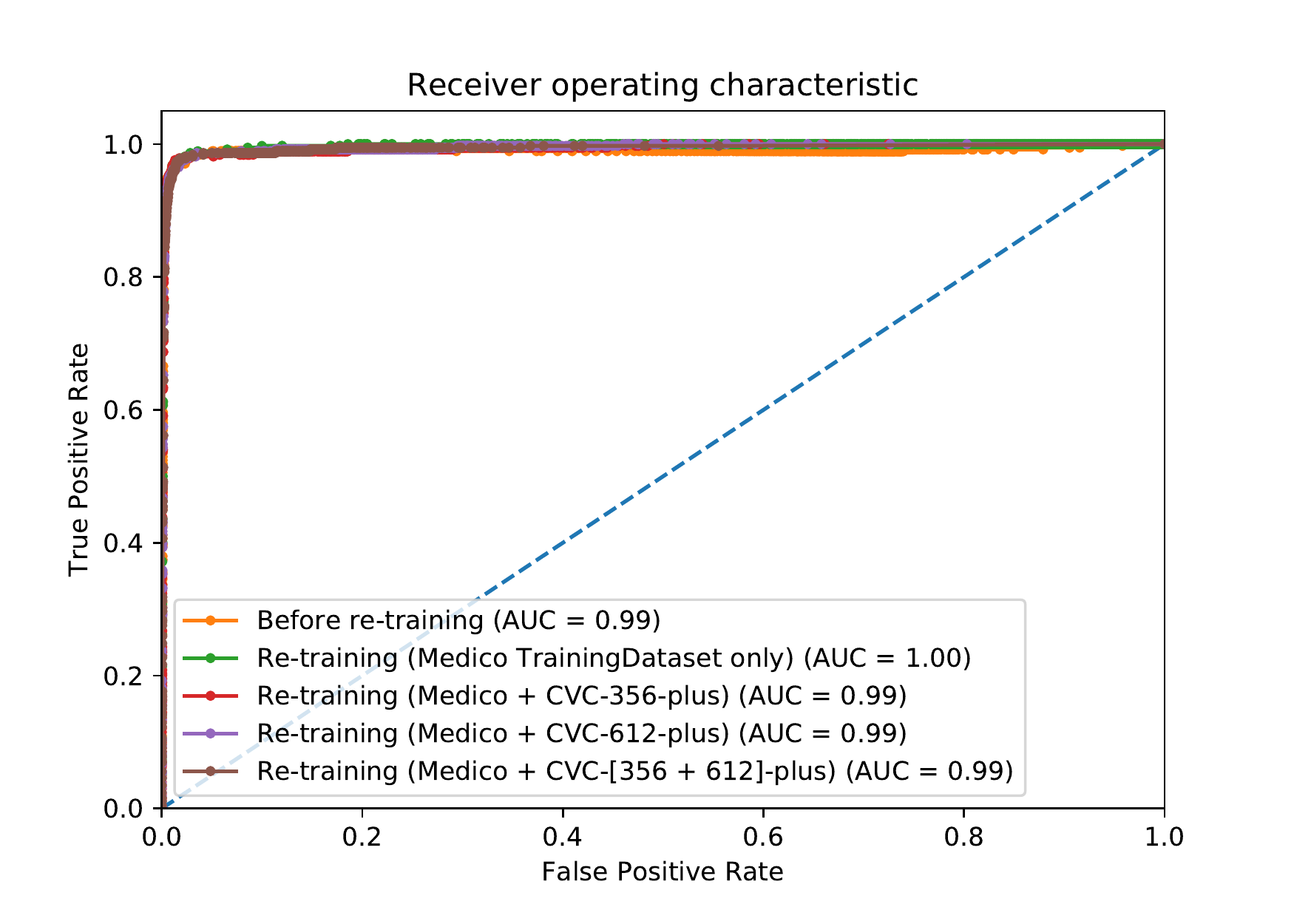}
        \label{plot:roc_medico_test_dataset}
    }\hfill
   % \hspace{0mm}
    \subfloat[][\centering ROC curves for the CVC-12k dataset.]{
        \includegraphics[width=.48\linewidth, trim=0cm 0cm 0cm 0cm, clip]{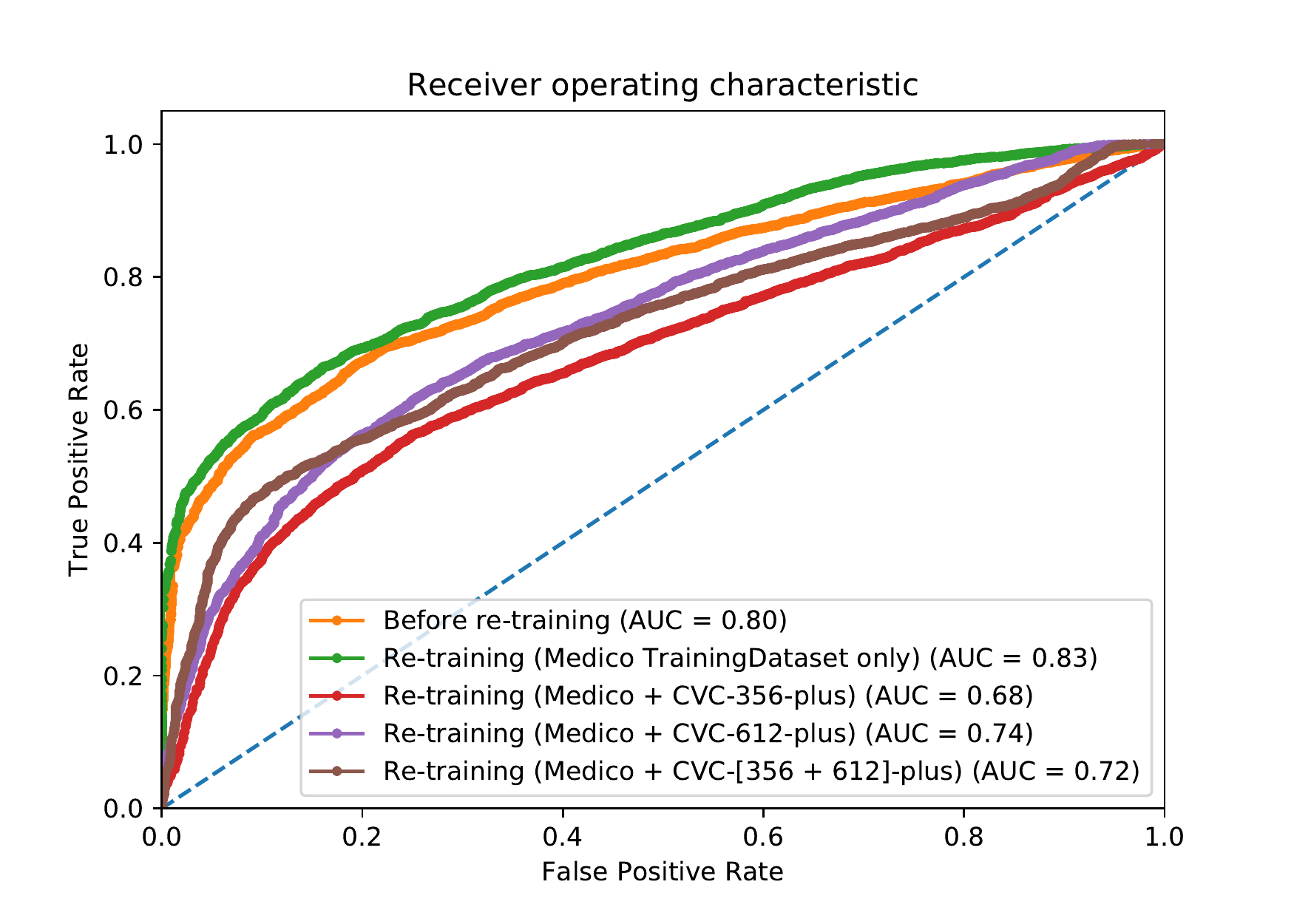}
        \label{plot:roc_cvc12k_dataset}
    }
    \vspace{-10pt}
    
    \subfloat[][\centering PRC curves for the Medico test dataset.]{
        \includegraphics[width=.48\linewidth, trim=0cm 0cm 0cm 0cm, clip]{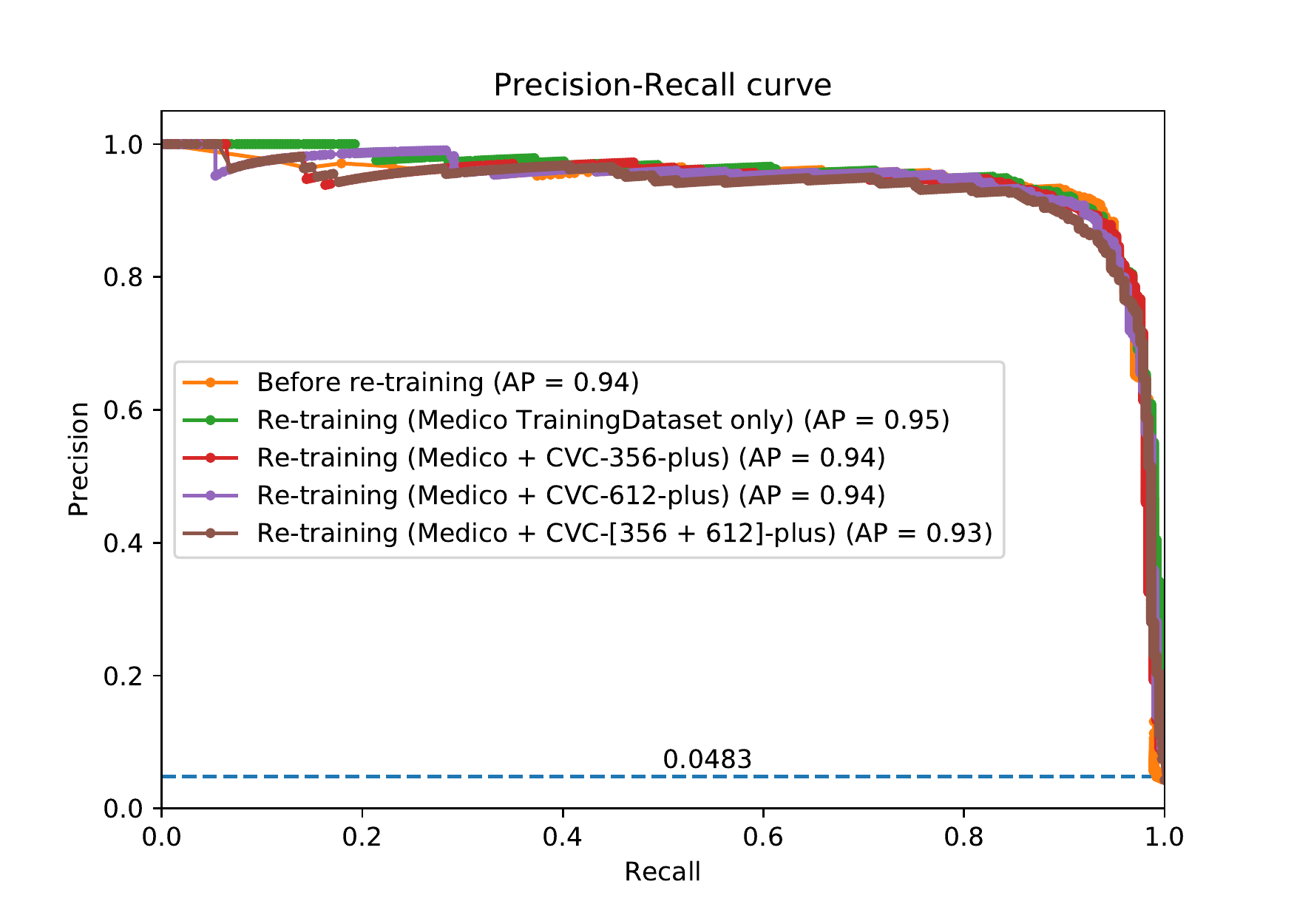}
        \label{plot:prc_medico_test_dataset}
    }\hfill
   % \hspace{0mm}
    \subfloat[][\centering PRC curves for the CVC-12k dataset.]{
        \includegraphics[width=.48\linewidth, trim=0cm 0cm 0cm 0cm, clip]{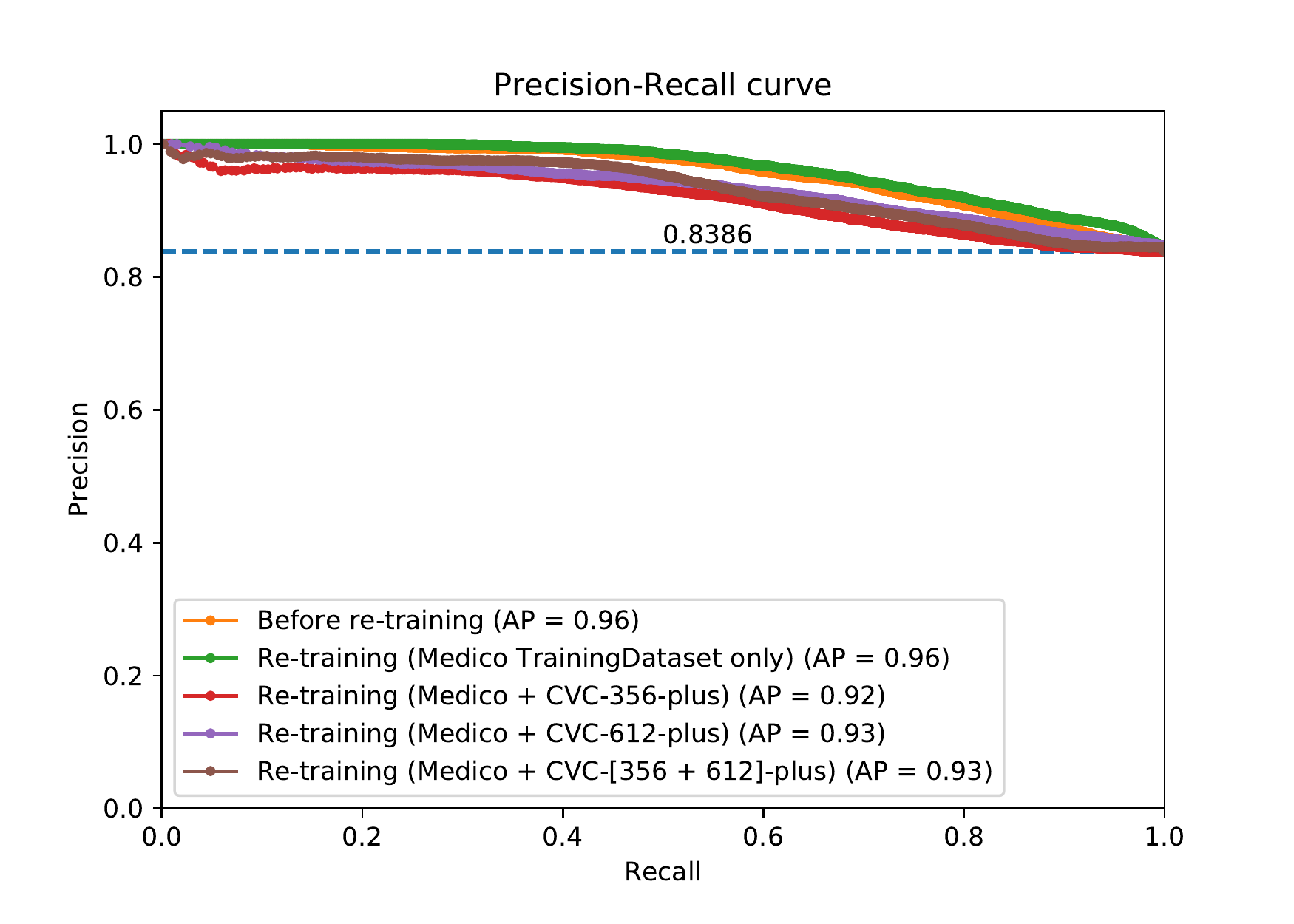}
        \label{plot:prc_cvc12k_dataset}
    }
    \caption{\ac{ROC} and \ac{PRC} curves for method 5 trained on the CVC-356-plus and CVC-612-plus dataset as mentioned in legends. Testing datasets are CVC-12k and the Medico test dataset. Overall good performance can be observed in both ROC and PRC. For CVC-12k the PRC curve shows the interesting case of a high random baseline for a biased dataset.}
    \label{plot:ROC and PRC curves}
    \vspace{-10pt}
\end{figure}

%%%%%%%%%%%%%%%%%%%%%%%%

Plots in the first column and the second column in Figure \ref{plot:overall_performance_all_methods} show completely different behaviors for the same retraining process when we use different test datasets. The test dataset for the first column comes from the same domain as the training data, and the test dataset for the second column comes from the completely new domain like CVC-12k dataset. To investigate these unusual performance changes, we generated and examined \ac{ROC} curves and \ac{PRC} curves for the best \ac{DNN} model (method 5). The \ac{ROC} and \ac{PRC} curves for method 5 with the Medico test data (for the plot in Figure \ref{plot:overall_performance_all_methods}) are depicted in Figures \ref{plot:roc_medico_test_dataset} and \ref{plot:prc_medico_test_dataset}. Similarly, the \ac{ROC} curve and the \ac{PRC} curve for the method 5 with CVC-12k data (for the plot in Figure \ref{plot:overall_performance_all_methods}) are plotted in the Figures \ref{plot:roc_cvc12k_dataset} and~\ref{plot:prc_cvc12k_dataset}.

Analysis of \ac{ROC} curves is more robust for \ac{ML} models that are used with balanced datasets, whereas \ac{PRC} curves are more valuable for \ac{ML} methods when the methods engage with imbalanced datasets. However, we have used both curves in this paper to investigate the behavior of these curves while we are using highly imbalanced datasets. Consequently, the \ac{PRC} curves show completely different baseline values of $0.0483$ for the Medico test dataset and $0.8386$ for the CVC-12k dataset. The small baseline value arises in the plot in Figure \ref{plot:prc_medico_test_dataset} as a result of small polyps to the non-polyp proportion in the Medico test dataset. Conversely, the high baseline value in Figure~\ref{plot:prc_cvc12k_dataset} appears there as an effect on a high ratio of polyps to non-polyps.

To get a better understanding of the above plots, we selected the plots in Figures \ref{plot:i} and \ref{plot:j}, and \ac{ROC} and \ac{PRC} curves in Figure \ref{plot:ROC and PRC curves}. With this selection, first, we analyze T1 and T2 from the hexagon plots and the corresponding \ac{ROC} and \ac{PRC} curves. While T2 shows a performance loss compared to T1 in Figure \ref{plot:i}, Figure \ref{plot:j} shows that T2 achieves a performance improvement over T1. Next, we look for the reasons for these performance changes.

In method 5, the model with the 16 outputs corresponding to T1 has 15 choices to classify non-polyp images. Similarly, the Medico test dataset has more non-polyp images compared to polyp images. On the other hand, the model corresponding to T2 has a $50\%$ chance to classify polyps as well as non-polyps. As a result, the model of T1 shows better performance than the model of T2 in Figure~\ref{plot:i}. Because this shows a slight performance change, we cannot see the same difference in \ac{ROC} and \ac{PRC} curves in Figures~\ref{plot:roc_medico_test_dataset} and \ref{plot:prc_medico_test_dataset}. In contrast, T2 in the plot in Figure~\ref{plot:j} shows performance improvement when the model has a 50:50 chance for classifying polyps and non-polyps. This improvement occurred as a result of a large number of polyps in the CVC-12k dataset. The \ac{ROC} and \ac{PRC} curves in plots in Figures~\ref{plot:roc_cvc12k_dataset} and \ref{plot:prc_cvc12k_dataset} shows this performance difference precisely. In other words, the model of T2 has a better chance of classifying polyps compared to the $1/16$ chance in the model of T1.

The retrained models corresponding to T3, T4 and T5 do not show considerable performance changes for the Medico test dataset as we can see from plots in Figures~\ref{plot:i}, \ref{plot:roc_medico_test_dataset}, and \ref{plot:prc_medico_test_dataset}. Conversely, the retraining method used in T3, T4 and T5 for the CVC-12k dataset shows large performance changes in the plots in Figures~\ref{plot:j}, \ref{plot:roc_cvc12k_dataset}, and \ref{plot:prc_cvc12k_dataset}. However, these methods show an overall performance loss. More comparisons about these plots are discussed in Section~\ref{sec:discuss}.

\begin{figure}
\centering
    \subfloat[][\centering Medico  test data - (CVC-356-plus).]{
        \includegraphics[width=.3\linewidth, trim=3cm 0cm 3cm 0cm, clip]{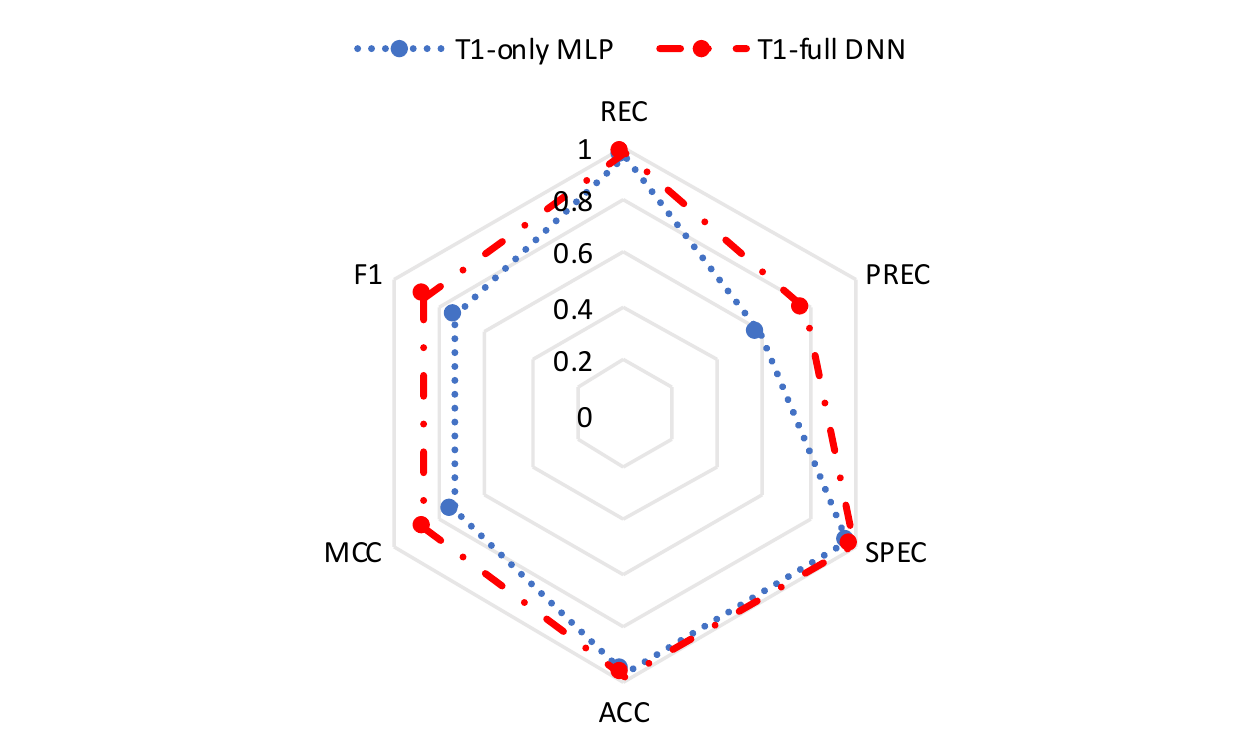}
        \label{plot:last_1}
    }\hfill
    \subfloat[][\centering Medico  test data - (CVC-612-plus).]{
        \includegraphics[width=.3\linewidth, trim=3cm 0cm 3cm 0cm, clip]{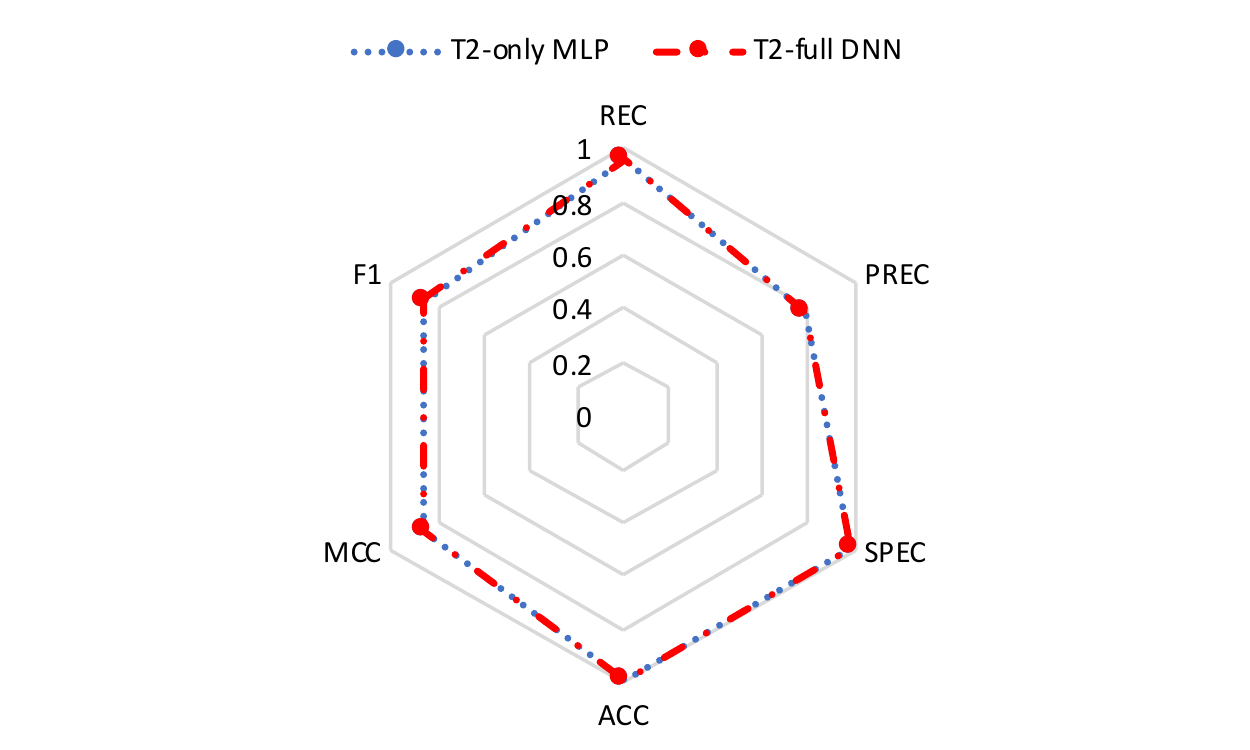}
        \label{plot:last_2}
    }\hfill
    \subfloat[][\centering Medico test data - (CVC-356-plus + CVC-612-plus).]{
        \includegraphics[width=.3\linewidth, trim=3cm 0cm 3cm 0cm, clip]{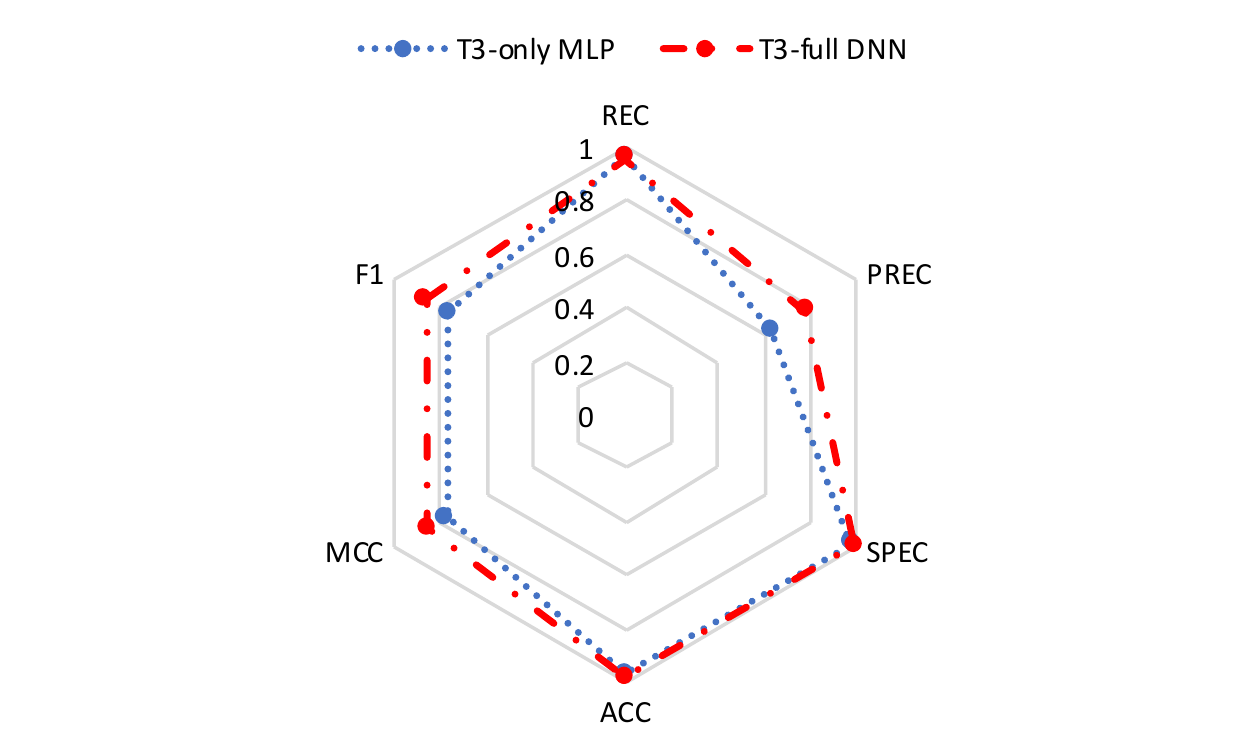}
        \label{plot:last_3}
    }
    \hspace{0mm}
    \subfloat[][\centering CVC-12k - (CVC-356-plus.)]{
        \includegraphics[width=.3\linewidth, trim=3cm 0cm 3cm 0cm, clip]{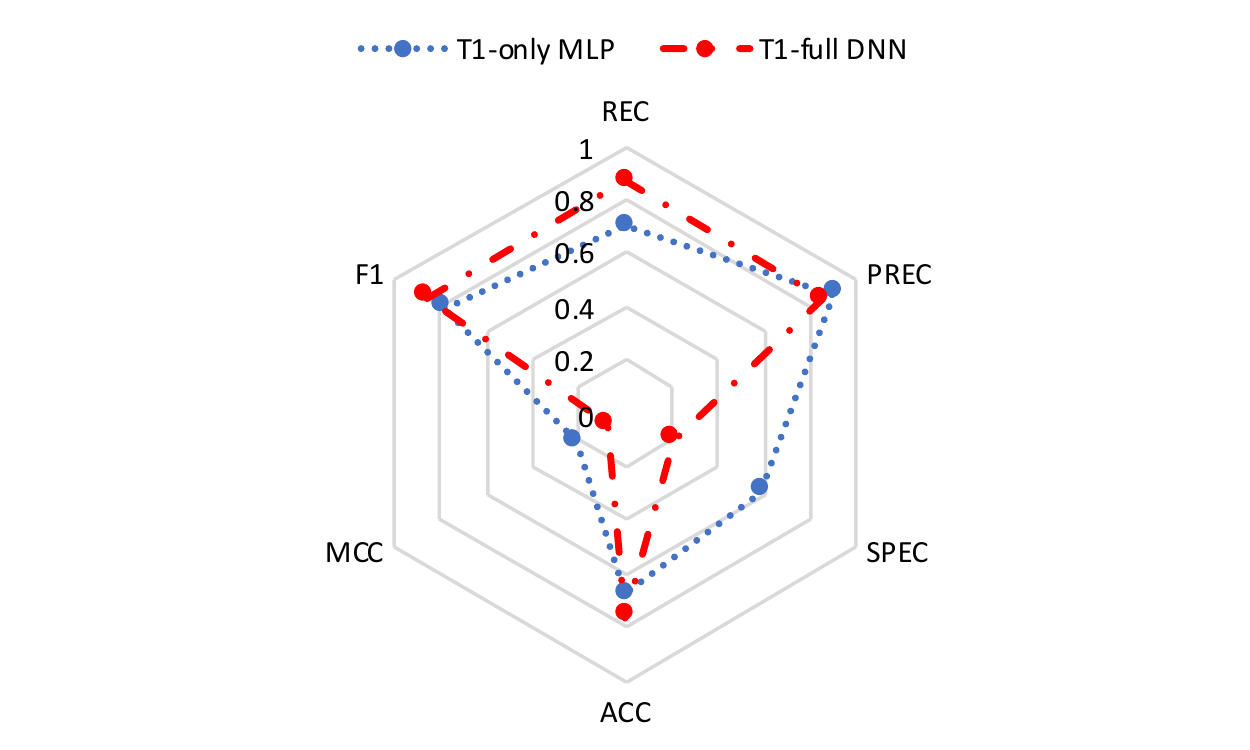}
        \label{plot:last_4}
    }\hfill
    \subfloat[][\centering CVC-12k - (CVC-612-plus).]{
        \includegraphics[width=.3\linewidth, trim=3cm 0cm 3cm 0cm, clip]{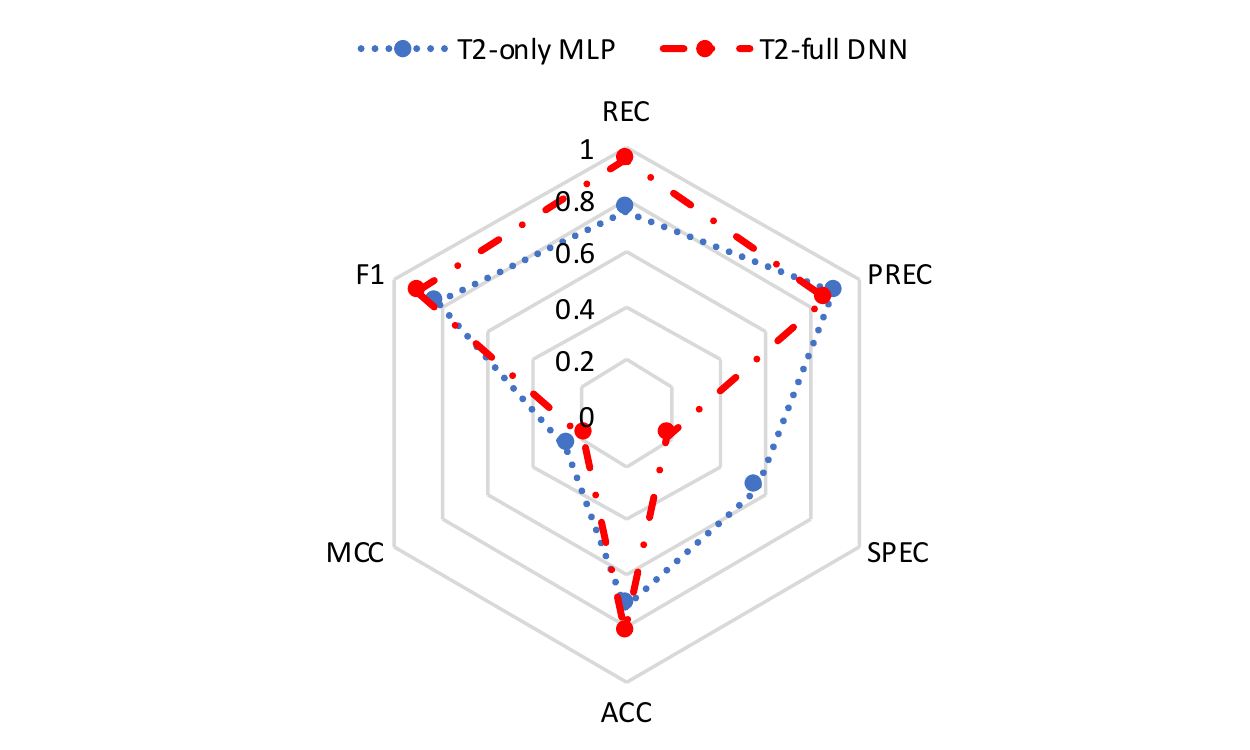}
        \label{plot:last_5}
    }\hfill
    \subfloat[][\centering CVC-12k - (CVC-356-plus + CVC-612-plus).]{
        \includegraphics[width=.3\linewidth, trim=3cm 0cm 3cm 0cm, clip]{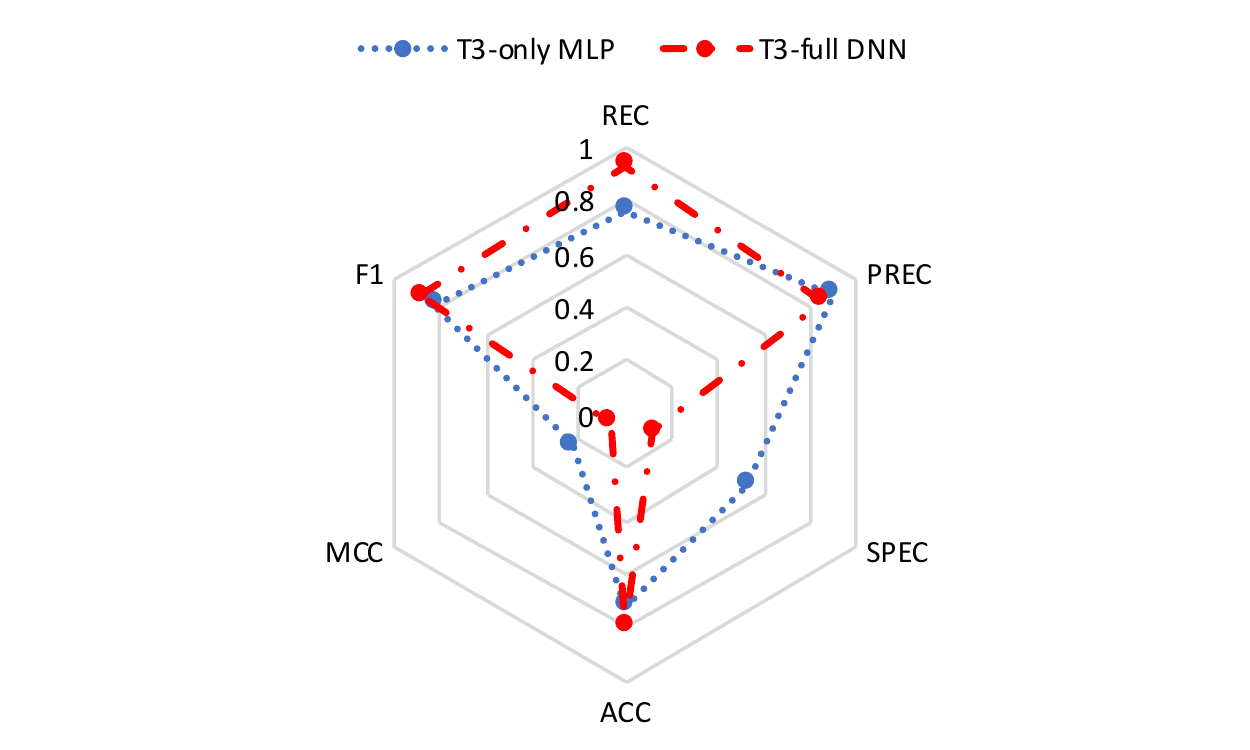}
        \label{plot:last_6}
   }
   \caption{Behavior of the complex \ac{DNN} method (method 5) while training only \ac{MLP} compared to training the whole \ac{DNN}. The first row shows the effects for these both cases when the test dataset is the Medico test data, and the second row shows the result when the test dataset is CVC-12k dataset. T1, T2, and T3 represent the training dataset used for the model. T1 - Medico training dataset + CVC356, T2 - Medico training dataset + CVC612, T3 - Medico training dataset + CVC356 + CVC612.}
%\caption*{\footnotesize T - Additional training dataset in addition to the MedicoTask training dataset; T1 - MedicoTask training dataset + CVC356, T2 - MedicoTask training dataset + CVC612, T3 - MedicoTask training dataset + CVC356 + CVC612}
\vspace{-13pt}
\label{plot:mlp_vs_whole_dnn}
\end{figure}

%%%%%%%%%%%%%%%%%%%%%%%%%%%%%%%%%%%%%%%%%%%%%%%%%%%%%%%%%%%%%%%%%%%%%%%%%%
For the following experiments, we analyzed method 5  even further. The main focus of this analysis is to understand the behavior of the best model for training only the \ac{MLP} versus training the whole \ac{DNN}. In this experiment, we collected results for two main test datasets: the Medico test dataset and CVC-12k dataset. Then, we collected performance measures from the two training mechanisms:training only the \ac{MLP} and training the whole \ac{DNN}. Furthermore, results were tabulated in Table \ref{tab:mlp_trainig_vs_dnn_training} and corresponding graphs were depicted in Figure \ref{plot:mlp_vs_whole_dnn} to analyze them.

% Validation accuracy for polyp detection after re-training
\begin{table} %[htbp]
\footnotesize
%\vspace{-12pt}
%\tiny 
  \caption{Method 5 - training only the MLP vs. complete-DNN. T - Additional training dataset which was added to the Medico dataset; T1 - Medico Dataset + CVC-356-plus, T2 - Medico Dataset + CVC-612-plus, T3 - Medico Dataset + CVC-356-plus + CVC-612-plus.}
%  \caption*{\footnotesize T - Additional training dataset in addition to the MedicoTask dataset; T1 - Medico Dataset + CVC-356-plus, T2 - Medico Dataset + CVC-612-plus, T3 - Medico Dataset + CVC-356-plus + CVC-612-plus}
%\vspace{-10pt}
  \label{tab:mlp_trainig_vs_dnn_training}
\begin{tabular}{P{6.5mm}|P{2mm}|P{6mm}P{6mm}P{6mm}P{6mm}P{6mm}P{7mm}|P{6mm}P{6mm}P{6mm}P{6mm}P{6mm}P{6mm}}
  %\begin{tabular}{llllllll|llllll}
    \toprule
   Test  &  & \multicolumn{6}{c}{Training only the MLP} & \multicolumn{6}{c}{Training the whole DNN}   \\
     Data&T&REC&PREC&SPEC&ACC&MCC&F1&REC&PREC&SPEC&ACC&MCC&F1\\
    \midrule
     
    \multirow{3}{*}{\rotatebox[origin=c]{90}{\parbox{1.2cm}{\centering\tiny Medico\\ Test\\Data }}}  
    & T1 & 0.9572 & 0.5859 & 0.9698 & 0.9692 & 0.7357 & 0.7269 & 0.9706 & 0.7773 & 0.9876 & 0.9868 & 0.8623 & 0.8633\\%[7pt]
    & T2 & 0.9599 & 0.7804 & 0.9879 & 0.9867 & 0.8591 & 0.8609 & 0.9626 & 0.7809 & 0.9879 & 0.9868 & 0.8606 & 0.8623\\%[7pt]
    & T3 & 0.9626 & 0.6316 & 0.9749 & 0.9744 & 0.7684 & 0.7627 & 0.9599 & 0.7771 & 0.9877 & 0.9865 & 0.8571 & 0.8589\\%[7pt]
    
     \midrule
    \multirow{3}{*}{\rotatebox[origin=c]{90}{\parbox{1.2cm}{\centering\tiny CVC-12k}}} 
    & T1 & 0.6984 & 0.8972 & 0.5842 & 0.6799 & 0.2184 & 0.7854 & 0.8694 & 0.8507 & 0.2068 & 0.7625  & 0.0802 & 0.8599\\%[7pt]
   & T2 & 0.7588 & 0.8993 & 0.5583 & 0.7265 & 0.2565 & 0.8231 & 0.9486 & 0.8571 & 0.1778 & 0.8242  & 0.1802 & 0.9005\\%[7pt]
   & T3 & 0.7614 & 0.8933 & 0.5272 & 0.7236 & 0.2352 & 0.8221 & 0.9278 & 0.8462 & 0.1239 & 0.7981  & 0.0699 & 0.8851\\%[7pt]
 
  \bottomrule
\end{tabular}
\vspace{-10pt}
\end{table}

%%%%%%%%%%%%%%%%%%%%%%%%%%%%%%%%%%%%%%%%%%%%%%%%%%%%%%%%%%%%%%%%%%%%%%%%%%%%%%%%%%%%%%%%%
% comprison of retrainig effect on medico test data and cvc-12k dataset based on the best method 5. 

%%%%%%%%%%%%%%%%%%%%%%%%%%%%%%%%%%%%%%%%%%%%%%%%%%%%%%%%%%%
%%%%%%%%%%%%%%%%%%%%%%%%%%%%%%

The first row of Figure~\ref{plot:mlp_vs_whole_dnn} shows the differences in the performance of testing with the Medico test data. In the second row, it presents the performance changes for the CVC-12k dataset. The dotted lines in plots in Figure~\ref{plot:mlp_vs_whole_dnn} represent the training only MLP.  Similarly, the dash lines represent the training of the whole \ac{DNN}. The three plots of each row represent results of retraining the model with the Medico training data and  CVC-356-plus dataset, CVC-612-plus dataset and both CVC-356-plus, and CVC-612-plus datasets, respectively. 

According to the plots in Figures~\ref{plot:last_1}, \ref{plot:last_2}, and \ref{plot:last_3}, it is clear that retraining the whole \ac{DNN} can be used to improve the overall performance of the \ac{DNN} model because we can see performance improvement in these plots except in Figure~\ref{plot:last_2}, which shows closely equal performance metrics. However, in test cases with the CVC-12k dataset, it shows a completely new behavior for retraining the whole \ac{DNN} as depicted in Figures~\ref{plot:last_4}, \ref{plot:last_5}, and \ref{plot:last_6}. These plots show large changes in the performance hexagons with considerable positive improvements for the recall and considerable performance loss for the specificity values. This experiment also shows that researchers could be misled by the performance monitoring process of \ac{DNN} methods using a single dataset. In other words, according to the first row of the figure, researchers may conclude that retraining the whole \ac{DNN} is a positive factor. However, the results of the second row prove that it is not always true by showing performance losses for the same technique.

The results presented in plots in Figures~\ref{plot:mlp_vs_whole_dnn} show difficulties in adapting \ac{ML} models for cross-dataset generalization with a different perspective. In that experiment, the performance loss in specificity, which is a parameter of reflecting \ac{TN} detection, shows that method 5 is affected by imbalanced data in the CVC-12k dataset. The main reason for the effect is that the CVC-12k dataset contains a lower percentage of negative images compared to positive ones. This reflects an important factor to take into account when developing generalizable \ac{ML} models which is that the ratio of negative and positive findings needs to be taken into account when looking at metrics. Metrics such as \ac{MCC} are better suited to interpret results. In terms of ROC compared to PRC, the results show that PRC reflects the performance of the model more realistically than ROC.

%=================================================================================

\section{Discussion}
\label{sec:discuss}

In this section, we discuss our findings and point out several important considerations for future research. Our discussion follows the same sequence as our contributions in this paper.

In our experiments, combinations of ResNet-152, Densenet-161, and an additional \ac{MLP} produced the best result for the Medico 2018 dataset. The reported results from this model for Medico Task led us to hold second place based on the \ac{MCC} values calculated by organizers, and there was only a tiny gap of around 0.0029. Furthermore, the winning team of this competition has used additional data items that were made by photo editing tools for the imbalanced classes such as the out-of-patients class. In contrast to this, our method 5 works well without using manually annotated data items because of the procedure we followed to implement and train that model. The procedure of implementing such a complex model is described step by step in Section~\ref{sec:method_5} and anyone can follow these steps to get a well-performing \ac{DL} model in a classification task.

In addition to the implementation and the procedure used in method~5, the data filling mechanism used to fill the out-of-patient imbalance class shows impressive performance gain.  This method is preferred when one class has a small number of data items in a multi-class classification task. In our work, without annotating more data ourself which also requires the help of medical experts, we prefer to use random images from the Internet, as described in Section~\ref{subsec:dataset}. This is an efficient way to add more data items without spending more time on manual annotation or creating synthetic data items. The above method works because the random images influence the \ac{ML} models to make a wider range of possibilities to classify images into a particular class.

Dyed-lifted-polyps, dyed-resection-margins, esophagitis, and normal-z-line arose classification conflicts in our best method (method 5). If we could overcome these conflicts, then the model would perform better than the current recorded performance in the 2018 Medico Task. To identify the reasons for this classification conflicts, we manually investigated the images of these classes. If we compare sample images of dyed-lifted-polyps (Figure~\ref{fig:dyed_lifted}), dyed-resection-margins (Figure~\ref{fig:dyed_resection}), esophagitis (Figure~\ref{fig:esophagitis}), and normal-z-line (Figure~\ref{fig:normal_z_line}), then we can identify that this conflict caused as a result of similar texture and shapes of these images. To overcome this problem, researchers can select only the images that made conflict and train a new \ac{DL} model to classify them into the correct classes. Then, this model can be added to the model introduced in method 5 using the property of its expandability.

% add example papers here

Can we use our best \ac{DL} model for real systems in hospitals to classify \ac{GI} findings? Or can we use the state of the art \ac{ML} classification models introduced by researchers in real applications? 
Towards answering this question,  this paper focuses on deep evaluations of the proposed methods as one of the main contributions. Regularly, researchers present the performance of their classification models using only a test dataset, which was reserved from the dataset used to produce the training data.  Also, they measure the performance selecting only a few measurements out of the \ac{REC}, \ac{PREC}, \ac{SPEC}, \ac{ACC}, \ac{MCC}, and \ac{F1}. However, we emphasize the requirements of an in-depth analysis of all of these six parameters at once to identify the real performance of \ac{ML} models. Several of the works listed in Table~\ref{tab:Overview_of_the_related_work} do not use this methodology as part of their evaluations. This makes it difficult to reason about the real-world performance of the proposed methods and how they compare with other methods. In this paper we also consider  the importance of evaluating \ac{ML} models with cross datasets.

Why do we need cross-dataset evaluations? To explain this requirement, we consider the research work done by Wang et al.~\cite{wang2018development}. They presented an area under the receiver operating characteristic curve of 0.984  and a per-image sensitivity of $94.38$ for polyp detection. In our first look, these results show a good \ac{DL} model. Similarly, our results in Figures \ref{plot:i}, \ref{plot:roc_medico_test_dataset} and \ref{plot:prc_medico_test_dataset} reflect the same impression in the first look because it shows excellent performance as a \ac{DL} model.  However, after analyzing cross-dataset performance for polyp detection with a completely new dataset like CVC-12k, we recognized that performance gain is not enough for applying it in real applications. Therefore, from this paper, we emphasize that researchers want to consider cross-dataset evaluations very thoroughly before applying their solutions in real-world applications. Otherwise, the selection bias, the capture bias, and the category bias (label bias) problems may appear in the results. Then, we may end up with the wrong conclusion about research works. All of these facts imply that more research works must be performed to improve the generalizability along with the performance improvement on a single dataset or single data source.

%===========================

\section{Conclusions}
\label{sec:conclusion}

We have studied cross-dataset bias and evaluation metrics interpretation in \acf{ML} using five methods and four different datasets within the field of \acf{GI} endoscopy as respective use case. In particular,  we performed an extensive study of \ac{ML} models in the context medical applications based on a use-case of \ac{GI} tract abnormality classification across different datasets. The main conclusion and resulting recommendation is that a multi-center or cross-dataset evaluation is important if not essential for \ac{ML} models in the medical field to obtain a realistic understanding of the performance of such models in real world settings.  

We found that the combination of the \acf{DNN} ResNet-152 and Densenet-161 with an additional \acf{MLP} performed best on both the validation and test datasets. This model shows that a combination of multiple pre-trained \ac{DNN} models can have better capabilities to classify images into the correct classes because of their cumulative decision-making capabilities. We also proposed an evaluation method using six measures: \acf{REC}, \acf{PREC}, \acf{SPEC}, \acf{ACC}, \acf{MCC}, and \acf{F1}. Moreover, we suggest that these measures should be presented all at once using hexagon plots that convey a complete view of real performance. We hope that these tools can enable a more realistic evaluation and comparison of \ac{ML} methods.

Furthermore, we presented cross-dataset evaluations to identify the generalizability of our \ac{ML} models emphasizing on the fact that achieving high scores for evaluation metrics does not always represent the real performance of \ac{ML} models and should be interpreted with care. By evaluating the \ac{ML} models with cross-datasets experiments, we showed the complexity of understanding the real functional performance of the models.  The state of the art research works which perform classification cannot be used in practical applications because of their lack of generalizability. Based on the experimental results, we conclude that researchers should focus on implementing and researching generalizable \ac{ML} models with cross-dataset evaluations. Rather than presenting metrics calculated from a simple training and testing split of the data, we suggest to always rely on cross-dataset evaluation to obtain a real-world representative indication of the model performance. This is especially important in a medical context since one has to make sure that the obtained models are reliable and not just perform well on a specific dataset.

Finally, we want point out that the lack of generalization, as evidenced by the poor result for cross-dataset evaluation presented in this article, rises the very important question:
In the context of cross-dataset or multi-center studies, is it really possible to have generalizable \ac{ML} models?
This is something that we ourselves plan to investigate further in future work, and we hope that other researchers in computer science and medicine will do the same or at least have the question in their mind when performing similar studies.

%==========================

\begin{acks}
We would like the reviewers for their contributions to the article. This work is funded in part by the Research Council of Norway project number 263248 (Privaton).
\end{acks}

% Bibliography
\bibliographystyle{ACM-Reference-Format}
\bibliography{mainreferences}

%%%%%%%%%%%%%%%%%%%%%%%%%%%%%%
\end{document}